\documentclass[10pt,letterpaper]{article}
\usepackage[top=0.85in,left=2.75in,footskip=0.75in]{geometry}

\usepackage{amsmath,amssymb}

\usepackage{changepage}

\usepackage[utf8x]{inputenc}

\usepackage{textcomp,marvosym}

\usepackage{cite}

\usepackage{nameref,hyperref}

\usepackage[right]{lineno}

\usepackage{array}

\newcolumntype{+}{!{\vrule width 2pt}}

\newlength\savedwidth

\raggedright
\usepackage[aboveskip=1pt,labelfont=bf,labelsep=period,justification=raggedright,singlelinecheck=off]{caption}

\makeatletter
\renewcommand{\@biblabel}[1]{\quad#1.}
\makeatother

\usepackage{lastpage,fancyhdr}
\usepackage{graphicx}
\usepackage{color}

\usepackage{lastpage,fancyhdr}
\fancyheadoffset[L]{2.25in}
\fancyfootoffset[L]{2.25in}
\usepackage{bm,url,booktabs,algorithm,algorithmic}
\newcommand{\argmin}{\mathop{\rm argmin}\limits}

\begin{document}
\vspace*{0.2in}

\begin{flushleft}
{\Large
\textbf\newline{Prediction of hierarchical time series using structured regularization and its application to artificial neural networks
} % Please use "sentence case" for title and headings (capitalize only the first word in a title (or heading), the first word in a subtitle (or subheading), and any proper nouns).
}
\newline
\\
Tomokaze Shiratori\textsuperscript{1},
Ken Kobayashi\textsuperscript{2*},
Yuichi Takano\textsuperscript{3}
\\
\bigskip
\textbf{1} Graduate School of Systems and Information Engineering, University of Tsukuba, Tsukuba, Ibaraki, Japan
\\
\textbf{2} Artificial Intelligence Laboratory, Fujitsu Laboratories Ltd., Kawasaki, Kanagawa, Japan
\\
\textbf{3} Faculty of Engineering, Information and Systems, University of Tsukuba, Tsukuba, Ibaraki, Japan
\\
\bigskip

% Insert additional author notes using the symbols described below. Insert symbol callouts after author names as necessary.
% 
% Remove or comment out the author notes below if they aren't used.
%
% Primary Equal Contribution Note
%\Yinyang These authors contributed equally to this work.

% Additional Equal Contribution Note
% Also use this double-dagger symbol for special authorship notes, such as senior authorship.
%\ddag These authors also contributed equally to this work.

%\textcurrency Current Address: Dept/Program/Center, Institution Name, City, State, Country % change symbol to "\textcurrency a" if more than one current address note
% \textcurrency b Insert second current address 
% \textcurrency c Insert third current address


% Group/Consortium Author Note
%\textpilcrow Membership list can be found in the Acknowledgments section.

% Use the asterisk to denote corresponding authorship and provide email address in note below.
* ken-kobayashi@fujitsu.com %ytakano@sk.tsukuba.ac.jp

\end{flushleft}
\section*{Abstract}
This paper discusses the prediction of hierarchical time series, where each upper-level time series is calculated by summing appropriate lower-level time series. 
Forecasts for such hierarchical time series should be coherent, meaning that the forecast for an upper-level time series equals the sum of forecasts for corresponding lower-level time series. 
Previous methods for making coherent forecasts consist of two phases: first computing base (incoherent) forecasts and then reconciling those forecasts based on their inherent hierarchical structure. 
With the aim of improving time series predictions, we propose a structured regularization method for completing both phases simultaneously. 
The proposed method is based on a prediction model for bottom-level time series and uses a structured regularization term to incorporate upper-level forecasts into the prediction model. 
We also develop a backpropagation algorithm specialized for application of our method to artificial neural networks for time series prediction. 
Experimental results using synthetic and real-world datasets demonstrate the superiority of our method in terms of prediction accuracy and computational efficiency.

% Please keep the Author Summary between 150 and 200 words
% Use first person. PLOS ONE authors please skip this step. 
% Author Summary not valid for PLOS ONE submissions.   
%\section*{Author summary}
%Lorem ipsum dolor sit amet, consectetur adipiscing elit. Curabitur eget porta erat. Morbi consectetur est vel gravida pretium. Suspendisse ut dui eu ante cursus gravida non sed sem. Nullam sapien tellus, commodo id velit id, eleifend volutpat quam. Phasellus mauris velit, dapibus finibus elementum vel, pulvinar non tellus. Nunc pellentesque pretium diam, quis maximus dolor faucibus id. Nunc convallis sodales ante, ut ullamcorper est egestas vitae. Nam sit amet enim ultrices, ultrices elit pulvinar, volutpat risus.

%\linenumbers
\nolinenumbers
% Use "Eq" instead of "Equation" for equation citations.
\section*{Introduction}

Multivariate time series data often have a hierarchical (tree) structure in which each upper-level time series is calculated by summing appropriate lower-level time series. 
For instance, numbers of tourists are usually counted on a regional basis, such as sites, cities, regions, or countries~\cite{AtAh09}. 
Similarly, many companies require regionally aggregated forecasts to support resource allocation decisions~\cite{KrSi16}.
Product demand is often analyzed by category to reduce the overall forecasting burden~\cite{Fl99}. 

Forecasts for such hierarchical time series should be \emph{coherent}, meaning that the forecast for an upper-level time series equals the sum of forecasts for corresponding lower-level time series~\cite{BeTa17,WiAt19}. 
Smoothing methods such as the moving average and exponential smoothing are widely used in both academia and industry for time series predictions~\cite{BrDa91,HyAt18}. 
Although these methods provide coherent forecasts for hierarchical time series, they have low accuracy, especially for rapidly changing time series. 

Another common approach for making coherent forecasts is the use of bottom-up and top-down methods~\cite{Fl99,KaMa19,Lu11,WiVi09}. 
These methods first develop base forecasts by separately predicting each time series and then reconcile those base forecasts based on their inherent hierarchical structure. 
The bottom-up method calculates base forecasts for bottom-level time series and then aggregates them for upper-level time series. 
In contrast, the top-down method calculates base forecasts only for a root (total) time series and then disaggregates them according to historical proportions of lower-level time series. 
Park and Nassar~\cite{PaNa14} considered a hierarchical Bayesian dynamic proportions model for the top-down method to sequentially disaggregate upper-level forecasts. 
The middle--out method calculates base forecasts for intermediate-level time series and then applies the bottom-up and top-down methods to make upper- and lower-level forecasts, respectively. 
However, the bottom-up method often accumulates prediction errors as the time series level rises, and the top-down method cannot exploit detailed information about lower-level time series.
Notably, when base forecasts are unbiased, only the bottom-up method gives unbiased forecasts~\cite{HyAh11}. 

Hyndman et~al.~\cite{HyAh11} proposed a linear regression approach to optimal base forecasts by the bottom-up method. 
This forecast reconciliation method worked well for predicting tourism demand~\cite{AtAh09} and monthly inflation~\cite{CaCo10}, and this approach can be extended to hierarchical and grouped time series~\cite{HyLe16}. 
van Erven and Cugliari~\cite{VaCu15} devised a game-theoretically optimal reconciliation method. 
Regularized regression models have also been employed to deal with high-dimensional time series~\cite{BeKo19,BeYu17}. 
Wickramasuriya et~al.~\cite{WiAt19} devised a sophisticated method for optimal forecast reconciliation through trace minimization, 
and their experimental results showed that this trace minimization method performed very well with synthetic and real-world datasets. 
Note, however, that all of these forecast reconciliation methods consist of two phases: first computing base forecasts and then reconciling those forecasts based on a hierarchical structure. 
The aim of this study was to produce better time series predictions by simultaneously completing these two phases. 

Structured regularization uses inherent structural relations among explanatory variables to construct a statistical model~\cite{HaTi15,JeAu11,ZhRo09}.
Various regularization methods have been proposed for multivariate time series~\cite{NiMa17,ScMo17}, hierarchical explanatory variables~\cite{BiTa13,KiXi12,LiHa15,SaTa19}, and artificial neural networks~\cite{WeWu16}.
Prediction of multivariate time series is related to multitask learning, which shares useful information among related tasks to enhance the prediction performance for all tasks~\cite{Ca97,ZhYa17}. 
Tailored regularization methods have been developed for multitask learning~\cite{EvPo04,JaVe09} and applied to artificial neural networks~\cite{Ru17}. 
To the best of our knowledge, however, no prior studies have applied structured regularization methods to predictions of hierarchical time series. 

In this study, we aimed to develop a structured regularization method that takes full advantage of hierarchical structure for better time series predictions. 
Our method is based on a prediction model for bottom-level time series and uses a structured regularization term to incorporate upper-level forecasts into the prediction model. 
This study particularly focused on application of our method to artificial neural networks, which have been effectively used in time series prediction~\cite{GaMu18,Hs04,KhBi10,LaCa18,Zh03,ZhQi05}. 
We developed a backpropagation algorithm specialized for our structured regularization model based on artificial neural networks. 
Experiments involving application of our method to synthetic and real-world datasets demonstrated the superiority of our method in terms of prediction accuracy and computational efficiency. 

\section*{Methods}
This section briefly reviews forecasts for hierarchical time series and forecast reconciliation methods. 
It then presents our structured regularization model and its application to artificial neural networks. 
This section also describes a backpropagation algorithm for artificial neural networks with structured regularization. 

\subsection*{Forecasts for hierarchical time series}
\begin{figure}[t]
\includegraphics[keepaspectratio, scale=0.25]{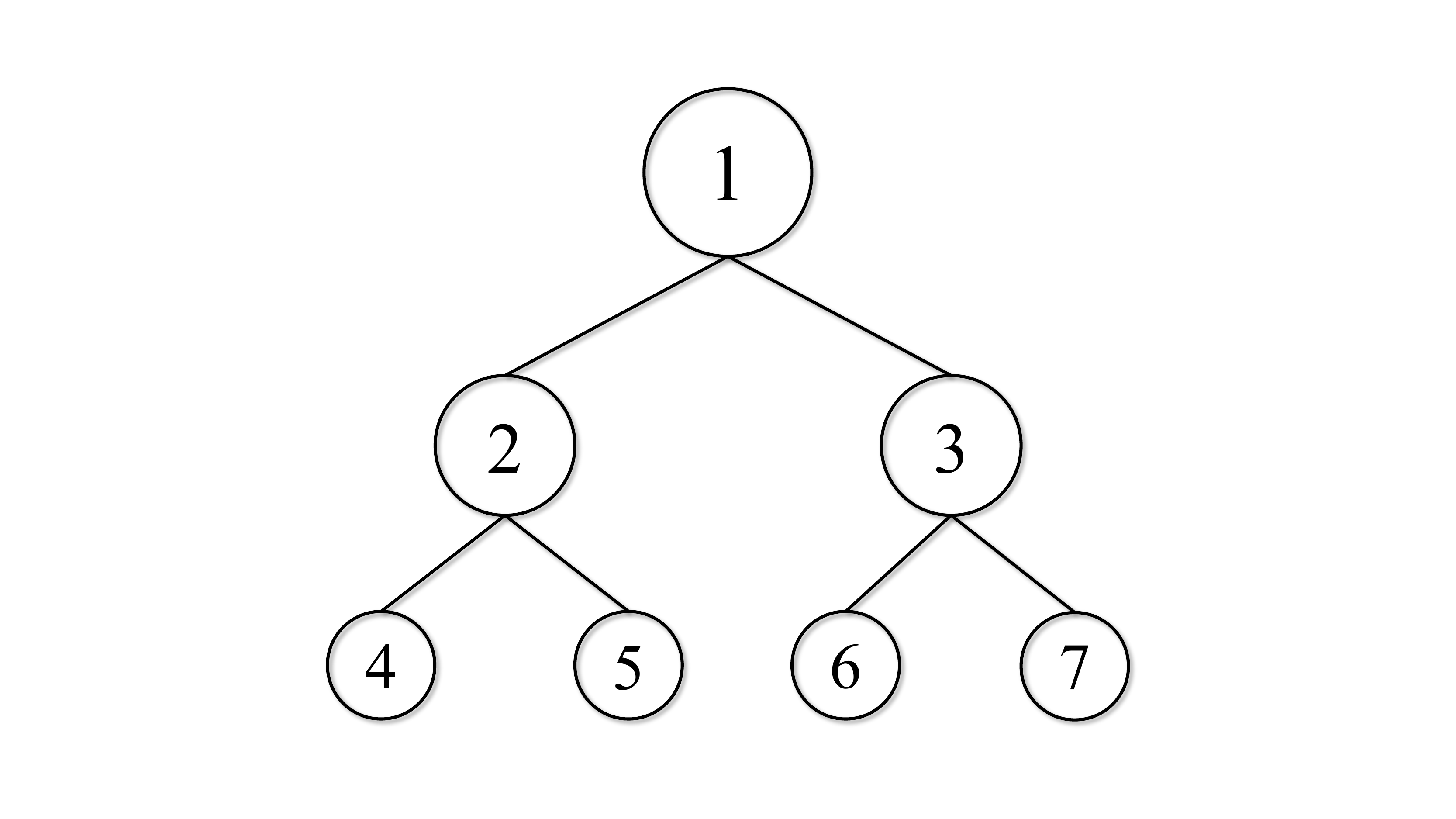}
\caption{Two-level hierarchical structure with $|N|=7$.}
\label{fig:hier_mini}
\end{figure}

We address prediction of multivariate time series where each series is represented as a node in a hierarchical (tree) structure. 
Let $y_{it}$ be an observation of node $i \in N$ at time $t \in T$, where $N$ is the set of nodes and $T$ is the set of timepoints. 
For simplicity, we focus on two-level hierarchical structures. 
Fig.~\ref{fig:hier_mini} shows the example of a two-level hierarchical structure with $|N|=13$ nodes, where $|\cdot|$ denotes the number of set elements. 
The nodes are classified as 
\[
N = \{1\} \cup M \cup B, \quad M = \{2,3\}, \quad B = \{4,5,6,7\},
\]
where node~1 is the root (level-zero) node, and $M$ and $B$ are sets of mid-level (level-one) and bottom-level (level-two) nodes, respectively. 
The associated time series is characterized by the \emph{aggregation constraint} 
\begin{align}\label{eq:agcon1}
\begin{cases}
~y_{1t} = y_{4t} + y_{5t} + y_{6t} + y_{7t}, \\
~y_{2t} = y_{4t} + y_{5t}, \\
~y_{3t} = y_{6t} + y_{7t}, 
\end{cases}
\quad (t \in T). 
\end{align}
Each upper-level time series is thus calculated by summing the corresponding lower-level time series. 

A hierarchical structure is represented by the \emph{structure matrix} $\bm{H} := (h_{ki})_{(k,i) \in (N \setminus B) \times B}$ as 
\begin{align}\notag
h_{ki} =
\begin{cases}
1 & \mbox{if node $k$ is an ascendant of node $i$}, \\
0 & \mbox{otherwise}, \\
\end{cases}
\quad (k \in N \setminus B,~i \in B).
\end{align}
We define the \emph{summing matrix} as
\[
\bm{S} := (s_{ki})_{(k,i) \in N \times B} := 
\begin{bmatrix}
\bm{H} \\
\bm{I}_{|B|}
\end{bmatrix},
\]
where $\bm{I}_n$ is the identity matrix of size $n$. 
In Fig.~\ref{fig:hier_mini}, we have 
\[
\bm{H} = 
\begin{bmatrix}
1 & 1 & 1 & 1 \\
1 & 1 & 0 & 0 \\
0 & 0 & 1 & 1
\end{bmatrix}, \quad 
\bm{S} = 
\begin{bmatrix}
1 & 1 & 1 & 1 \\
1 & 1 & 0 & 0 \\
0 & 0 & 1 & 1 \\
1 & 0 & 0 & 0 \\
0 & 1 & 0 & 0 \\
0 & 0 & 1 & 0 \\
0 & 0 & 0 & 1
\end{bmatrix}.
\]

Let $\bm{y}_t := (y_{it})_{i \in N}$ be a column vector comprising observations of all nodes at time $t \in T$. 
Similarly, for a node subset $A \subseteq N$ we define $\bm{y}^A_t := (y_{it})_{i \in A}$ as the observation vector of nodes $i \in A$ at time $t \in T$. 
In Fig.~\ref{fig:hier_mini}, we have
\begin{align}
& \bm{y}_t = (y_{1t},y_{2t},y_{3t},y_{4t},y_{5t},y_{6t},y_{7t})^{\top}, \notag \\
& \bm{y}^M_t = (y_{2t},y_{3t})^{\top}, \quad \bm{y}^B_t = (y_{4t},y_{5t},y_{6t},y_{7t})^{\top} \quad (t \in T). \notag
\end{align}

The aggregation constraint~\eqref{eq:agcon1} is then expressed as 
\begin{align} \label{eq:agg1}
\bm{y}^{N \setminus B}_t = \bm{H} \bm{y}^B_t \quad (t \in T)
\end{align}
or, equivalently, 
\begin{align} \label{eq:agg2}
\bm{y}_t = \bm{S} \bm{y}^B_t \quad (t \in T).
\end{align}

Let $\hat{\bm{y}}_t := (\hat{y}_{it})_{i \in N}$ be a column vector comprising \emph{base forecasts} at time $t \in T$. 
Note that the base forecasts are calculated separately for each node $i \in N$, so they do not satisfy the aggregation constraint~\eqref{eq:agg1}. 
For a node subset $A \subseteq N$, we define $\hat{\bm{y}}^A_t := (\hat{y}_{it})_{i \in A}$ at time $t \in T$. 
Such base forecasts can be converted into \emph{coherent forecasts} satisfying the aggregation constraint~\eqref{eq:agg1} by using the \emph{reconciliation matrix} $\bm{P} :=(p_{ij})_{(i,j) \in B \times N}$. 
Specifically, we develop bottom-level forecasts $\tilde{\bm{y}}^B_t = \bm{P}\hat{\bm{y}}_t$ and use the aggregation constraint~\eqref{eq:agg2} to obtain coherent forecasts, as
\begin{align}\label{eq:cohe}
\tilde{\bm{y}}_t = \bm{S}\bm{P}\hat{\bm{y}}_t \quad (t \in T).  
\end{align}

A typical example of a reconciliation matrix is  
\[
\bm{P} = [\bm{O}_{|B| \times |N \setminus B|},~\bm{I}_{|B|}], 
\]
where $\bm{O}_{m \times n}$ is a $m \times n$ zero matrix. 
This leads to the bottom-up method 
\begin{align}\label{eq:bu}
\tilde{\bm{y}}_t = \bm{S} \hat{\bm{y}}^B_t \quad (t \in T). 
\end{align}
Another example is
\[
\bm{P} = [\bm{p},~\bm{O}_{|B| \times |N \setminus \{1\}|}],
\]
where $\bm{p} = (p_i)_{i \in B}$ is a column vector comprising historical proportions of bottom-level time series. 
This results in the top-down method 
\[
\tilde{\bm{y}}_t = \bm{S} (\hat{y}_{1t}\bm{p}) \quad (t \in T).
\]

In this manner, we can make coherent forecasts from various reconciliation matrices. 
The condition $\bm{S}\bm{P}\bm{S} = \bm{S}$ is proven to ensure that when base forecasts are unbiased, the resultant coherent forecasts~\eqref{eq:cohe} are also unbiased~\cite{HyAh11}. 
This condition is also known to be fulfilled only by the bottom-up method~\cite{HyAh11}. 

\subsection*{Forecast reconciliation methods}

Hyndman et~al.~\cite{HyAh11} introduced the following linear regression model based on the aggregation constraint~\eqref{eq:agg2}:  
\[
\hat{\bm{y}}_t = \bm{S} \bm{\beta}_t + \bm{\varepsilon}_t \quad (t \in T), 
\]
where $\bm{\beta}_t := (\beta_{it})_{i \in B}$ is a column vector of bottom-level estimates, and $\bm{\varepsilon}_t := (\varepsilon_{it})_{i \in B}$ is a column vector of errors having zero mean and covariance matrix $\mbox{var}(\bm{\varepsilon}_t) := \bm{\Sigma}_t$. 
The bottom-up method~\eqref{eq:bu} with $\hat{\bm{y}}^B_t = \bm{\beta}_t$ is then used to makes coherent forecasts. 

If the base forecasts are unbiased and the covariance matrix $\bm{\Sigma}_t$ is known, the generalized least-squares estimation yields the minimum variance unbiased estimate of $\bm{\beta}_t$. 
However, the covariance matrix $\bm{\Sigma}_t$ is nonidentifiable and therefore impossible to estimate~\cite{WiAt19}. 

In contrast, Wickramasuriya et~al.~\cite{WiAt19} focused on differences between observations and coherent forecasts~\eqref{eq:cohe},
\[
\bm{e}_t := \bm{y}_t - \tilde{\bm{y}}_t = \bm{y}_t - \bm{S}\bm{P}\hat{\bm{y}}_t  \quad (t \in T). 
\]
The associated covariance matrix is derived as
\begin{align}\label{eq:cov}
\mbox{var}(\bm{e}_t) = \bm{S} \bm{P} \bm{W}_t \bm{P}^{\top} \bm{S}^{\top}  \quad (t \in T), 
\end{align}
where $\bm{W}_t := \mathbb{E}[(\bm{y}_t - \hat{\bm{y}}_t) (\bm{y}_t - \hat{\bm{y}}_t)^{\top}]$ is the covariance matrix of base forecasts. 
The trace of the covariance matrix~\eqref{eq:cov} is minimized subject to the unbiasedness condition $\bm{S}\bm{P}\bm{S} = \bm{S}$. 
This yields the optimal reconciliation matrix 
\[
\bm{P} = (\bm{S}^{\top}\bm{W}_t^{-1}\bm{S})^{-1} \bm{S}^{\top} \bm{W}_t^{-1}, 
\]
and coherent forecasts~\eqref{eq:cohe} are given by 
\begin{align}\label{eq:minT}
\tilde{\bm{y}}_t = \bm{S} (\bm{S}^{\top}\bm{W}_t^{-1}\bm{S})^{-1} \bm{S}^{\top} \bm{W}_t^{-1} \hat{\bm{y}}_t \quad (t \in T).
\end{align}
See Wickramasuriya et~al.~\cite{WiAt19} for the full details. 

Note, however, that in these forecast reconciliation methods, base forecasts are first determined regardless of the underlying hierarchical structure, then those forecasts are corrected based on the hierarchical structure. 
In contrast, our proposal is a structured regularization model that directly computes high-quality forecasts based on the hierarchical structure. 

\subsection*{Structured regularization model}

We consider a prediction model for bottom-level time series. 
Its predictive value is denoted by the column vector $\hat{\bm{y}}_t^B(\bm{\Theta}) := (\hat{y}_{it}(\bm{\Theta}))_{i \in B}$, where $\bm{\Theta}$ is a set of model parameters. 
As an example, the first-order vector autoregressive model is represented as
\[
\hat{y}_{it}(\bm{\Theta}) = \sum_{j \in B} \theta_{ij} y_{j,t-1} \quad (i \in B,~t \in T),
\]
where $\bm{\Theta} = (\theta_{ij})_{(i,j) \in B \times B}$. 

The residual sum of squares for bottom-level time series is given by 
\begin{align} \label{eq:RSS}
\sum_{t \in T} \|\bm{y}_t^B - \hat{\bm{y}}_t^B(\bm{\Theta})\|_2^2
= \sum_{t \in T} \sum_{i \in B} (y_{it} - \hat{y}_{it}(\bm{\Theta}))^2.
\end{align}
We also introduce a structured regularization term that quantifies the error for upper-level time series based on the hierarchical structure. 
Let $\bm{\Lambda} := \mathrm{Diag}(\bm{\lambda})$ be a diagonal matrix of regularization parameters, where $\bm{\lambda} := (\lambda_i)_{i \in N \setminus B}$ is a vector of its diagonal entries. 
Then, we construct a structured regularization term based on the aggregation constraint~\eqref{eq:agg1} as
\begin{align} \label{eq:RegTerm}
\sum_{t \in T} \|\bm{\Lambda}(\bm{y}_t^{N \setminus B} - \bm{H}\hat{\bm{y}}_t^B(\bm{\Theta}))\|_2^2.
\end{align}
Minimizing this term aids in correcting bottom-level forecasts, thus improving the upper-level forecasts. 

Adding the regularization term~\eqref{eq:RegTerm} to the residual sum of squares~\eqref{eq:RSS} yields the objective function $E(\bm{\Theta})$ to be minimized. 
Consequently, our structured regularization model is posed as 
\begin{align}\label{eq:RegModel}
\bm{\Theta}^* \in \argmin_{\bm{\Theta}} \left\{E(\bm{\Theta}) := \frac{1}{2} \sum_{t \in T} \|\bm{y}_t^B - \hat{\bm{y}}_t^B(\bm{\Theta})\|_2^2 + \frac{1}{2} \sum_{t \in T} \|\bm{\Lambda}(\bm{y}_t^{N \setminus B} - \bm{H}\hat{\bm{y}}_t^B(\bm{\Theta}))\|_2^2 \right\}. 
\end{align}
Here, matrix $\bm{\Lambda}$ adjusts the tradeoff between minimizing the error term~\eqref{eq:RSS} for bottom-level times series and minimizing the error term~\eqref{eq:RegTerm} for upper-level time series. 
In the experiments section, we set its diagonal entries as
\begin{align}\label{eq:RegPara}
\lambda_i = 
\begin{cases}
\lambda_1 & (i=1), \\
\lambda_M & (i \in M), 
\end{cases}
\end{align}
where $\lambda_1$ and $\lambda_{M}$ are regularization parameters for root and mid-level time series, respectively. 

After solving the structured regularization model~\eqref{eq:RegModel}, we use the bottom-up method~\eqref{eq:bu} to obtain coherent forecasts
\[
\tilde{\bm{y}}_t = \bm{S} \hat{\bm{y}}_t^B(\bm{\Theta^*}). 
\]

\subsection*{Application to artificial neural networks}

\begin{figure}[t]
\includegraphics[keepaspectratio, scale=0.25]{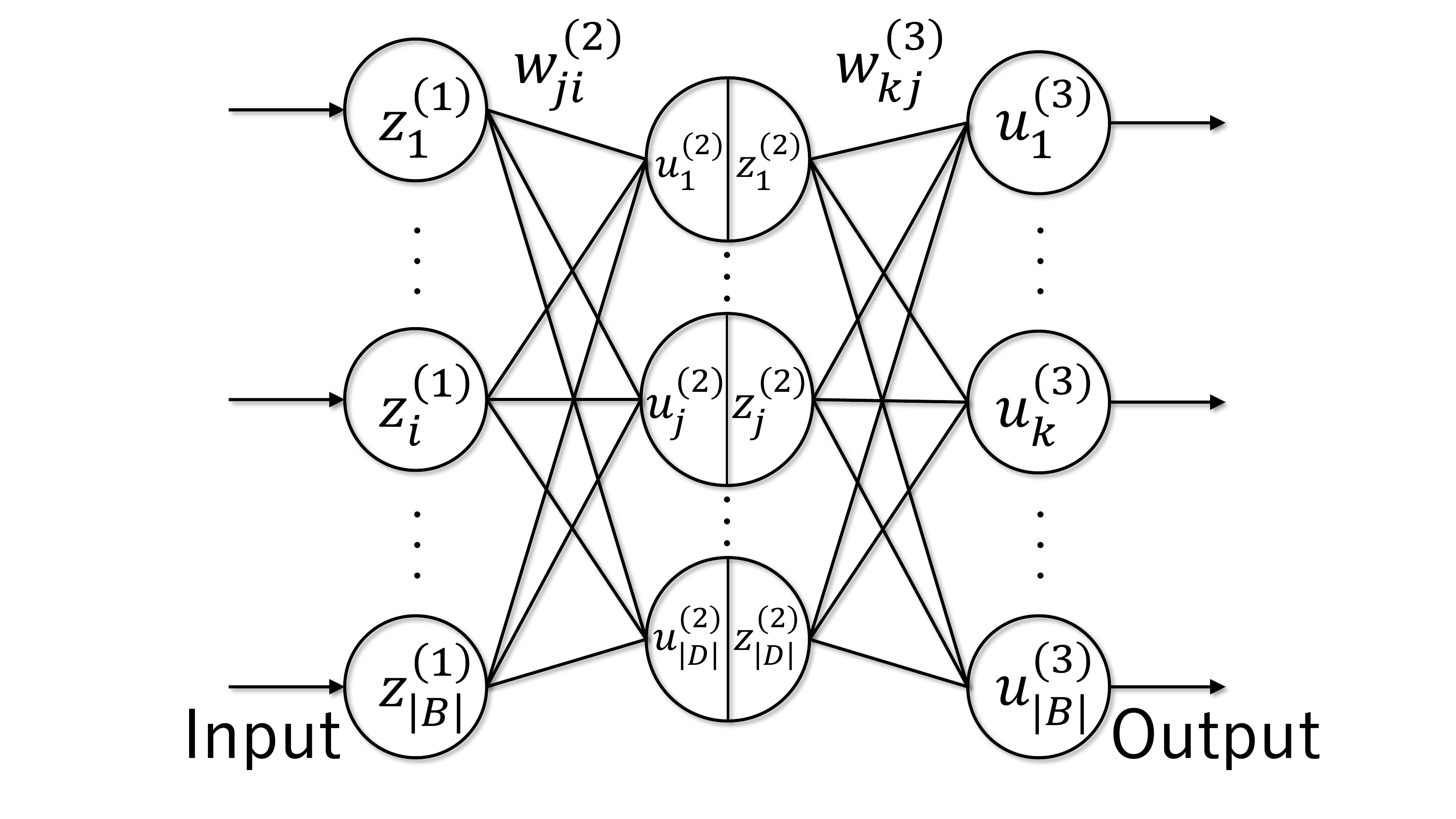}
\caption{Network diagram for a two-layer neural network.}
\label{fig:nn}
\end{figure}

This study focused on application of our structured regularization model~\eqref{eq:RegModel} to artificial neural networks for time series prediction; see Bishop~\cite{Bi06} and Goodfellow et~al.~\cite{GoBe16} for general descriptions of artificial neural networks. 
For simplicity, we consider a two-layer neural network like the one shown in Fig.~\ref{fig:nn}, where the input vector $\bm{z}^{(1)} := (z^{(1)}_i)_{i \in B}$ is defined as
\[
z^{(1)}_i = y_{i,t-1} \quad (i \in B). 
\]

First, we calculate the vector $\bm{u}^{(2)} := (u^{(2)}_j)_{j \in D}$ as the weighted sum of the input entries 
\begin{align} \label{eq:u2}
u_j^{(2)} = \sum_{i \in B} w^{(2)}_{ji} z^{(1)}_i \quad (j \in D),
\end{align}
where $\bm{W}^{(2)} := (w^{(2)}_{ji})_{(j,i) \in D \times B}$ is a weight matrix to be estimated. 
This vector $\bm{u}^{(2)}$ is transferred from the input units to \emph{hidden units}, as shown in Fig.~\ref{fig:nn}. 

Next, we generate the vector $\bm{z}^{(2)} := (z^{(2)}_j)_{j \in D}$ by nonlinear \emph{activation functions} as
\[
z^{(2)}_j = f(u_j^{(2)}) \quad (j \in D). 
\]
Typical examples of activation functions include the logistic sigmoid function
\begin{align}\label{eq:sig}
f(u) = \frac{1}{1 + \exp(- u)} 
\end{align}
and the rectified linear function
\[
f(u) = \max\{u,0\}. 
\]
The vector $\bm{z}^{(2)}$ is transferred from the hidden units to the output units as shown in Fig.~\ref{fig:nn}. 

Finally, we calculate the vector $\bm{u}^{(3)} := (u^{(3)}_k)_{k \in B}$ as the weighted sum of the output entries from the hidden units as 
\begin{align} \label{eq:u3}
u^{(3)}_k = \sum_{j \in D} w^{(3)}_{kj} z^{(2)}_j = \sum_{j \in D} w^{(3)}_{kj} f(u_j^{(2)}) \quad (k \in B), 
\end{align}
where $\bm{W}^{(3)} := (w^{(3)}_{kj})_{(k,j) \in B \times D}$ is a weight matrix to be estimated. 

This process is summarized as 
\begin{align}\label{eq:fwd}
\bm{z}^{(1)} = \bm{y}_{t-1}^B, \quad
\bm{u}^{(2)} = \bm{W}^{(2)} \bm{z}^{(1)}, \quad 
\bm{z}^{(2)} = \bm{f}(\bm{u}^{(2)}), \quad 
\bm{u}^{(3)} = \bm{W}^{(3)} \bm{z}^{(2)}, 
\end{align}
where the set of model parameters is 
\[
\bm{\Theta} = \{\bm{W}^{(2)},\bm{W}^{(3)}\}. 
\]
This neural network outputs $\hat{\bm{y}}_t^B(\bm{\Theta}) = \bm{u}^{(3)}$ as a vector of predictive values. 

\subsection*{Backpropagation algorithm}

We develop a backpropagation algorithm specialized for training artificial neural networks in our structured regularization model~\eqref{eq:RegModel}; see Bishop~\cite{Bi06} and Goodfellow et~al.~\cite{GoBe16} for overviews of backpropagation algorithms. 
Our algorithm sequentially minimizes the following error function for time $t \in T$: 
\begin{align} \label{eq:errfun}
& E_t(\bm{\Theta}) := \frac{1}{2} \|\bm{y}_t^B - \bm{u}^{(3)}\|_2^2 + \frac{1}{2} \|\bm{\Lambda}(\bm{y}_t^{N \setminus B} - \bm{H}\bm{u}^{(3)})\|_2^2 \quad (t \in T).
\end{align}

We first define vectors $\bm{\delta}^{(2)} := (\delta_j^{(2)})_{j \in D}$ and $\bm{\delta}^{(3)} := (\delta_k^{(3)})_{k \in B}$, which consist of partial derivatives of the error function~\eqref{eq:errfun} with respect to intermediate variables~\eqref{eq:u2} and~\eqref{eq:u3} as follows:
\begin{align}
\delta_j^{(2)} := \frac{\partial E_t(\bm{\Theta})}{\partial u_j^{(2)}} \quad (j \in D), \qquad
\delta_k^{(3)} := \frac{\partial E_t(\bm{\Theta})}{\partial u_k^{(3)}} \quad (k \in B). \notag 
\end{align}
From Eqs.~\eqref{eq:u2} and~\eqref{eq:u3}, the partial derivatives of the error function~\eqref{eq:errfun} can be calculated as 
\begin{align}
\frac{\partial E_t(\bm{\Theta})}{\partial w^{(2)}_{ji}} 
= \frac{\partial E_t(\bm{\Theta})}{\partial u_j^{(2)}} \frac{\partial u_j^{(2)}}{\partial w^{(2)}_{ji}}
= \delta_j^{(2)} z_i^{(1)} \quad (i \in B,~j \in D), \label{eq:grad2} \\
\frac{\partial E_t(\bm{\Theta})}{\partial w^{(3)}_{kj}} 
= \frac{\partial E_t(\bm{\Theta})}{\partial u_k^{(3)}} \frac{\partial u_k^{(3)}}{\partial w^{(3)}_{kj}}
= \delta_k^{(3)} z_j^{(2)} \quad (j \in D,~ k \in B). \label{eq:grad3}
\end{align}

From Eq.~\eqref{eq:u3}, we have 
\[
\frac{\partial u^{(3)}_k}{\partial u^{(2)}_j} = w^{(3)}_{kj} f'(u^{(2)}_j) \quad (j \in D,~ k \in B). 
\]
Therefore, 
\begin{align}
\delta_j^{(2)} 
= \frac{\partial E_t(\bm{\Theta})}{\partial u_j^{(2)}} 
= \sum_{k \in B} \frac{\partial E_t(\bm{\Theta})}{\partial u_k^{(3)}} \frac{\partial u_k^{(3)}}{\partial u_j^{(2)}} 
= \sum_{k \in B} \delta^{(3)}_k w^{(3)}_{kj} f'(u^{(2)}_j) \quad (j \in D). \label{eq:delta12}
\end{align}
It follows from Eq.~\eqref{eq:errfun} that 
\begin{align}
\bm{\delta}^{(3)} := \frac{\partial E_t(\bm{\Theta})}{\partial \bm{u}^{(3)}} 
& = -(\bm{y}_t^B - \bm{u}^{(3)}) - (\bm{\Lambda} \bm{H})^{\top} \bm{\Lambda}(\bm{y}_t^{N \setminus B} - \bm{H}\bm{u}^{(3)}) \notag \\
& = -(\bm{y}_t^B - \bm{u}^{(3)}) - \bm{H}^{\top} \bm{\Lambda}^2 (\bm{y}_t^{N \setminus B} - \bm{H}\bm{u}^{(3)}) \notag \\
& = - [\bm{H}^{\top} \bm{\Lambda}^2,~\bm{I}_{|B|}] \bm{y}_t + (\bm{I}_{|B|} + \bm{H}^{\top} \bm{\Lambda}^2 \bm{H}) \bm{u}^{(3)} \label{eq:delta3}.
\end{align}

Algorithm~\ref{alg:bp} summarizes our backpropagation algorithm. 
\begin{algorithm}
\caption{Backpropagation algorithm.}
\label{alg:bp}
\begin{algorithmic}
\STATE
\begin{description}
\item[Step~0~(Initialization):]
Let $\eta > 0$ be a step size and $\varepsilon > 0$ be a threshold for convergence. 
Set $E \leftarrow \infty$ as an incumbent value of the objective function $E(\bm{\Theta}) = \sum_{t \in T} E_t(\bm{\Theta})$. 
\item[Step~1~(Backpropagation):]
Repeat the following steps for all $t \in T$: 
\begin{description}
\item[Step 1.1:] Compute $\bm{z}^{(1)}$, $\bm{u}^{(2)}$, $\bm{z}^{(2)}$, and $\bm{u}^{(3)}$ from Eq.~\eqref{eq:fwd}. 
\item[Step 1.2:] Compute $\bm{\delta}^{(3)}$ from Eq.~\eqref{eq:delta3} and then $\bm{\delta}^{(2)}$ from Eq.~\eqref{eq:delta12}. 
\item[Step 1.3:] Compute the partial derivatives~\eqref{eq:grad2} and~\eqref{eq:grad3}. 
\end{description}
\item[Step~2~(Gradient Descent):]
Update the weight parameter values as 
\begin{align}\notag
\begin{cases}
\displaystyle ~w^{(2)}_{ji} \leftarrow w^{(2)}_{ji} - \eta \sum_{t \in T} \frac{\partial E_t(\bm{\Theta})}{\partial w^{(2)}_{ji}} & (i \in B,~j \in D), \\
\displaystyle ~w^{(3)}_{kj} \leftarrow w^{(3)}_{kj} - \eta \sum_{t \in T} \frac{\partial E_t(\bm{\Theta})}{\partial w^{(3)}_{kj}} & (j \in D,~k \in B). 
\end{cases}
\end{align}
\item[Step~3~(Termination Condition):]
If $E(\bm{\Theta}) > (1 - \varepsilon) E$, terminate the algorithm with $\bm{\Theta }= \{\bm{W}^{(2)},\bm{W}^{(3)}\}$. 
Otherwise, set $E \leftarrow E(\bm{\Theta})$ and return to Step 1. 
\end{description}
\end{algorithmic}
\end{algorithm}

\section*{Experimental results and discussion}
\begin{figure}[t]
\includegraphics[keepaspectratio, scale=0.25]{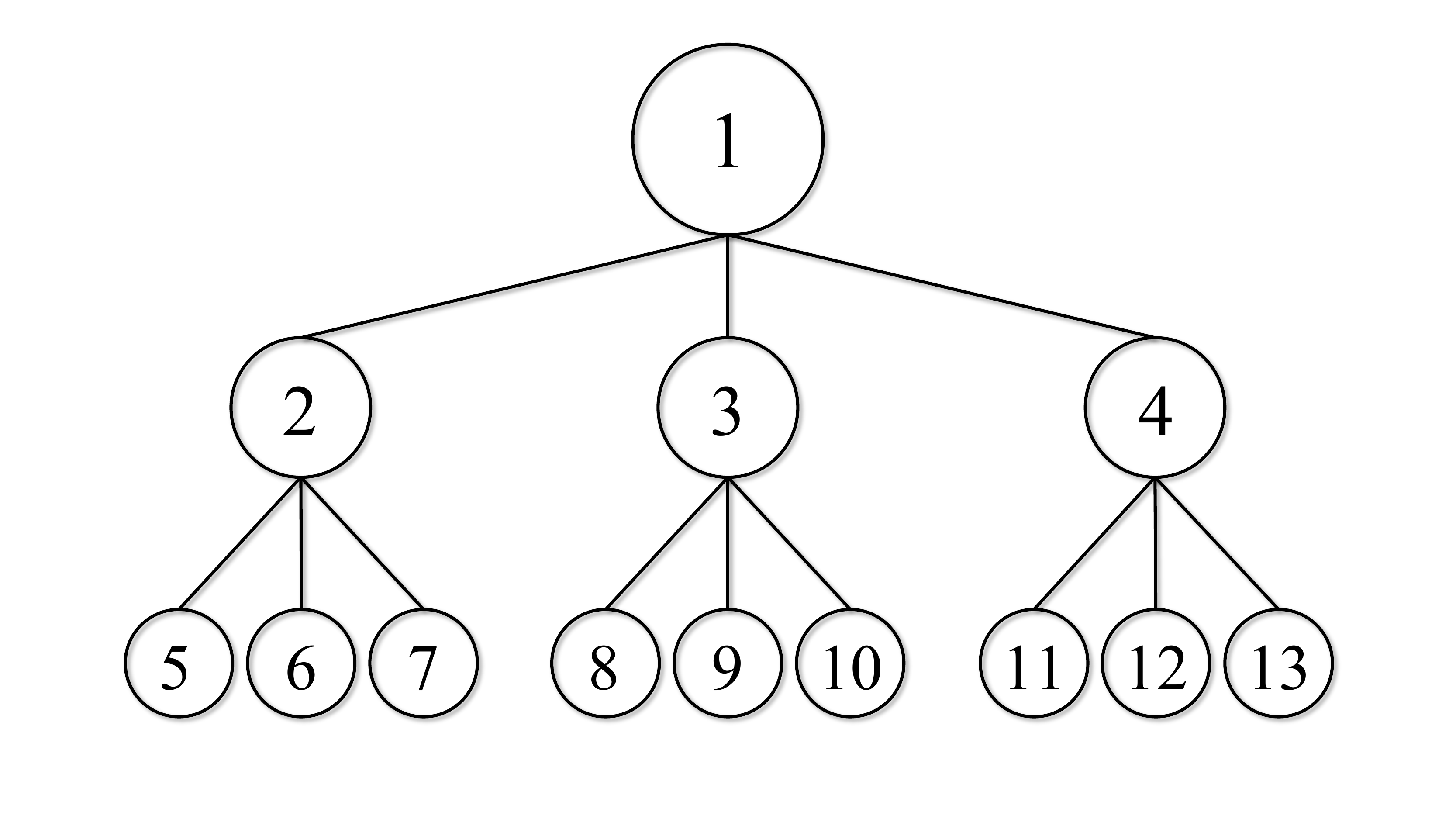}
\caption{Two-level hierarchical structure with $|N|=13$.}
\label{fig:hier_sim}
\end{figure}

The experimental results reported in this section evaluate the effectiveness of our structured regularization model when applied to artificial neural networks. 
These experiments focused on the two-level hierarchical structure shown in Fig.~\ref{fig:hier_sim}, where
\[
N = \{1\} \cup M \cup B, \quad M = \{2,3,4\}, \quad B = \{5,6,\ldots,13\}.
\]

\subsection*{Performance evaluation methodology}

To evaluate out-of-sample prediction performance, we considered training and test periods of time series data, where the training period was used to train prediction models, and the test period was used to compute prediction errors in the trained models. 
We calculated the root-mean-squared error (RMSE) for each node $i \in N$ during the test period $\hat{T}$ as
\[
    \mathrm{RMSE} := \sqrt{\frac{\sum_{t \in \hat{T}} (y_{it} - \tilde{y}_{it})^2}{|\hat{T}|}} \quad (i \in N). 
\]

We compared performance of the following methods for time series prediction. 
\begin{description}
\item[MA($n$):] moving average of the previous $n$ values, 
\[
\tilde{y}_{it} = \frac{\sum_{k=1}^n y_{i,t-k}}{n} \quad (i \in N,~t \in T)
\]
\item[ES($\alpha$):] exponential smoothing with smoothing parameter $\alpha \in [0,1]$, 
\[
\tilde{y}_{it} = \alpha y_{i,t-1} + (1 - \alpha) \tilde{y}_{i,t-1} \quad (i \in N,~t \in T)
\]
\item[NN+BU:] bottom-up method~\eqref{eq:bu} using artificial neural networks for base forecasts $\hat{\bm{y}}^B_t$ 
\item[NN+MinT:] forecast reconciliation method~\eqref{eq:minT} through trace minimization (i.e., MinT(Sample)~\cite{WiAt19}) using artificial neural networks for base forecasts $\hat{\bm{y}}_t$ 
\item[NN+SR($\lambda_1,\lambda_M$):] our structured regularization model~\eqref{eq:RegModel} applied to artificial neural networks with regularization parameters $\lambda_1$ and $\lambda_M$; see also Eq.~\eqref{eq:RegPara} 
\end{description}
Here, we determined parameter values for $n$ and $\alpha$ that minimized RMSE in the training period. 
During the training period, we tuned regularization parameters $\lambda_1$ and $\lambda_M$ through hold-out validation~\cite{BeHy18}. 

We adopted two-layer artificial neural networks (Fig.~\ref{fig:nn}), using the previous two values $y_{i,t-2}$ and $y_{i,t-1}$ to compute $\hat{y}_{it}(\bm{\Theta})$. 
Following prior studies~\cite{Ka12,Li87}, we set the number of hidden units to twice the number of input units (i.e., $|D| = 4 \cdot |B|$). 
Bias parameters were added to hidden and output units. 

We implemented the backpropagation algorithm (Algorithm~\ref{alg:bp}) in the R programming language, with the step size and convergence threshold set as $\eta = 1\cdot 10^{-5}$ and $\varepsilon = 5 \cdot 10^{-5}$, respectively. 
We employ the logistic sigmoid function~\eqref{eq:sig} as an activation function. 
The algorithm was repeated 30 times by randomly generating initial values for the parameter $\bm{\Theta}$ from a standard normal distribution. 
The following sections show average RMSE values with 95\% confidence intervals over the 30 trials. 

\subsection*{Synthetic datasets}

We generated common factors to express correlations among time series. 
Denote as $\mbox{N}(\mu,\sigma^2)$ a normal distribution with mean $\mu$ and standard deviation $\sigma$. 
For common factors, we used the first-order autoregressive models 
\[
\psi_{it} \sim \mbox{N}(\phi_i \psi_{i,t-1}, \sigma_i^2) \quad (i \in \{1\} \cup M,~t \in T),
\]
where $\phi_i$ is an autoregressive coefficient, and $\sigma_i$ is the standard deviation of white noise for the $i$th common factor. 
Note that $\psi_{it}$ reflects the overall trend for $i=1$ and mid-level trends for $i \in M = \{2,3,4\}$. 

Bottom-level time series were produced by combining the overall trend, mid-level trends, and autocorrelation. 
We denote the parent (mid-level) node of node $i$ as
\[
m(i) = 
\begin{cases}
~2 & (i \in \{5,6,7\}), \\
~3 & (i \in \{8,9,10\}), \\
~4 & (i \in \{11,12,13\}).  
\end{cases}
\]
For bottom-level time series, we used the first-order autoregressive models
\[
y_{it} \sim \mbox{N}(\rho_i \psi_{1t} + \theta_i \psi_{m(i),t} + \phi_i y_{i,t-1}, \sigma^2_i) \quad (i \in B,~t \in T), 
\]
where $\rho_i$ and $\theta_i$ respectively indicate effects of the common factors $\psi_{1t}$ and $\psi_{m(i),t}$ on the $i$th time series. 
After that, we generated upper-level time series ($y_{it}$ for $i \in N \setminus B$) according to the aggregation constraint~\eqref{eq:agg1}.

We prepared three synthetic datasets: NgtvC, WeakC, and PstvC. 
Table~\ref{tb:params} lists the parameter values used to generate these datasets. 
Time series are negatively correlated in the NgtvC dataset, weakly correlated in the WeakC dataset, and positively correlated in the PstvC dataset. 
Each dataset consists of time series data at 100 timepoints; the first 70 and latter 30 times were used as training and test periods, respectively. 
We standardized each time series according to the mean and variance over the training period. 

\renewcommand{\arraystretch}{0.75}
\begin{table}[t]
%\centering
\small
\caption{Parameter values for the synthetic datasets.}
\begin{tabular}{crrrrrrrr} \toprule
    &          &            & \multicolumn{2}{c}{NgtvC} & \multicolumn{2}{c}{WeakC} & \multicolumn{2}{c}{PstvC} \\ \cmidrule(lr){4-5}\cmidrule(lr){6-7}\cmidrule(lr){8-9}
Node $i$ & $\phi_i$ & $\sigma_i$ & $\rho_{i}$ & $\theta_{i}$ & $\rho_{i}$ & $\theta_{i}$ & $\rho_{i}$ & $\theta_{i}$\\ \midrule
1  & 0.3 & 0.3 & --- & --- & --- & --- & --- & --- \\[5pt]
2  & 0.3 & 0.3 & --- & --- & --- & --- & --- & --- \\
3  & 0.3 & 0.3 & --- & --- & --- & --- & --- & --- \\
4  & 0.3 & 0.3 & --- & --- & --- & --- & --- & --- \\[5pt]
5  & 0.3 & 0.3 & 0.1 & 1.0 & 0.1 & 0.1 &1.0 & 1.0 \\
6  & 0.3 & 0.3 & $-$0.1 & $-$1.0 & 0.1 & 0.1 & 1.0 & 1.0 \\
7  & 0.3 & 0.3 &1.0 & 0.1 & 0.1 & 0.1 & 1.0 & 1.0 \\
8  & 0.3 & 0.3 & 0.1 & 1.0 &0.1 & 0.1 & 1.0 & 1.0  \\
9  & 0.3 & 0.3 & $-$0.1 & $-$1.0 & 0.1 & 0.1 & 1.0 & 1.0 \\
10 & 0.3 & 0.3 &$-$1.0 & 0.1 & 0.1 & 0.1 & 1.0 & 1.0 \\ 
11 & 0.3 & 0.3 & 0.1 & 1.0 & 0.1 & 0.1 & 1.0 & 1.0  \\
12 & 0.3 & 0.3 & $-$0.1 & $-$1.0 & 0.1 & 0.1 & 1.0 & 1.0 \\
13 & 0.3 & 0.3 & 1.0 & 0.1 & 0.1 & 0.1 & 1.0 & 1.0 \\ \bottomrule
\end{tabular}
\label{tb:params}
\end{table}

\subsection*{Results for synthetic datasets}

Tables~\ref{tb:rmse_neg}--\ref{tb:rmse_pos} show the out-of-sample RMSE values provided by each method for each node in the NgtvC, WeakC, and PstvC datasets. 
In the tables, the rows labeled ``Mid-level'' and ``Bottom-level'' show the average RMSE values over the mid- and bottom-level nodes, respectively, with 
smallest RMSE values for each node indicated in bold. 

For the NgtvC dataset (Table~\ref{tb:rmse_neg}), our structured regularization method NN+SR clearly outperformed the other methods, except for the RMSE of the root node. 
For the WeakC dataset (Table~\ref{tb:rmse_ind}), our method was slightly inferior to the exponential smoothing method, but the differences were very small. 
For the PstvC dataset (Table~\ref{tb:rmse_pos}), the MinT method gave the best prediction performance, and our method attained the second-best value for average RMSE. 
These results show that our structured regularization method delivered good prediction performance for the three synthetic datasets. 
Our method was especially effective when the time series were strongly correlated, as in the NgtvC and PstvC datasets.

We next focus on the parameter values for our structured regularization.
Only for the PstvC dataset, our method NN+SR($\lambda_1,\lambda_M$) adopted $\lambda_1 > 0$ and performed significantly better than the bottom-up method in terms of the RMSE of the root node. 
Additionally, our method employed $\lambda_M > 0$ for all three datasets and outperformed the bottom-up method for mid-level RMSEs.
These results show an association between regularization weights and prediction accuracy at each time series level. 
Our method adjusts the regularization parameters to fit the data characteristic, thereby achieving better prediction performance. 

\begin{table}[t]
 \centering
 \small
 \caption{Prediction performance for the NgtvC dataset.}
 \begin{tabular}{crrrrr} \toprule
               &\multicolumn{5}{c}{RMSE} \\ \cmidrule(lr){2-6} 
  Node $i$     & MA(12) & ES(0.20) & NN+BU & NN+MinT & NN+SR(0.0, 2.1) \\ \midrule
  Root         & $\bm{1.09}$ & $1.10$ & $1.16 \pm{0.04}$ & $1.11 \pm{0.03}$ & $1.15 \pm{0.01}$ \\[5pt]
  2            & $0.63$ & $0.64$ & $0.66 \pm{0.03}$& $0.63 \pm{0.02}$ & $\bm{0.60} \pm{0.01}$  \\
  3            & $0.80$ & $0.76$ & $0.76 \pm{0.01}$& $\bm{0.72} \pm{0.01}$ & $0.73 \pm{0.01}$  \\
  4            & $0.71$ & $0.72$ & $0.70 \pm{0.02}$ & $0.69 \pm{0.01}$& $\bm{0.67} \pm{0.01}$    \\ \cmidrule(lr){2-6}
  Mid-level    & $0.71$ & $0.71$ & $0.71 \pm{0.01}$ & $0.68 \pm{0.00}$ & $\bm{0.67} \pm{0.00}$  \\[5pt]
  5            & $0.53$ & $0.48$ & $0.48\pm{0.02}$ & $0.48\pm{0.01}$ & $\bm{0.44}\pm{0.01}$  \\
  6            & $0.69$ & $\bm{0.64}$ & $0.65\pm{0.02}$ & $0.65\pm{0.02}$ & $\bm{0.64}\pm{0.02}$ \\
  7            & $0.39$ & $\bm{0.37}$ & $0.38\pm{0.00}$ & $0.43\pm{0.01}$ & $0.38\pm{0.00}$ \\
  8            & $0.42$ & $\bm{0.39}$ & $0.41\pm{0.01}$ & $0.41\pm{0.01}$ & $0.41\pm{0.01}$  \\
  9            & $\bm{0.35}$ & $\bm{0.35}$ & $0.38\pm{0.01}$ & $0.41\pm{0.01}$ & $0.38\pm{0.01}$ \\
  10           & $0.58$ & $0.56$ & $0.55\pm{0.02}$ & $0.54\pm{0.01}$ & $\bm{0.53}\pm{0.01}$ \\
  11           & $0.58$ & $0.49$ & $\bm{0.47}\pm{0.01}$ & $0.49\pm{0.01}$ & $\bm{0.47}\pm{0.01}$ \\
  12           & $0.50$ & $0.47$ & $0.47\pm{0.01}$ & $0.48\pm{0.01}$ & $\bm{0.46}\pm{0.00}$ \\ 
  13           & $0.48$ & $0.48$ & $0.47\pm{0.01}$ & $0.49\pm{0.02}$ & $\bm{0.46}\pm{0.01}$ \\ \cmidrule(lr){2-6}
  Bottom-level & $0.50$ & $0.47$ & $0.47 \pm{0.00}$ & $0.49 \pm{0.00}$ & $\bm{0.46} \pm{0.00}$ \\[5pt] 
  Average      & $0.60$ & $0.57$ & $0.58 \pm{0.00}$ & $0.58 \pm{0.00}$ & $\bm{0.56} \pm{0.00}$ \\ \bottomrule
 \end{tabular}
 \label{tb:rmse_neg}
\end{table}

\begin{table}[t]
 \centering
 \small
 \caption{Prediction performance for the WeakC dataset.}
 \begin{tabular}{crrrrr} \toprule
               &\multicolumn{5}{c}{RMSE} \\ \cmidrule(lr){2-6} 
  Node $i$     & MA(12) & ES(0.00) & NN+BU & NN+MinT & NN+SR(0.0, 1.2) \\ \midrule
  Root         & $1.06$ & $\bm{1.00}$ &  $1.06\pm{0.02}$ & $1.11\pm{0.04}$ & $1.06\pm{0.01}$ \\[5pt]
  2            & $0.46$ & $\bm{0.41}$ &  $0.45\pm{0.01}$ & $0.48\pm{0.02}$ & $0.44\pm{0.01}$ \\
  3            & $0.60$ & $\bm{0.56}$ &  $0.60\pm{0.01}$ &  $0.59\pm{0.02}$ & $0.58\pm{0.01}$ \\
  4            & $0.61$ & $0.60$ &  $0.59\pm{0.01}$ & $0.59\pm{0.01}$ & $\bm{0.57}\pm{0.01}$ \\ \cmidrule(lr){2-6}
  Mid-level    & $0.56$ & $\bm{0.52}$ & $0.55 \pm{0.00}$ & $0.55 \pm{0.01}$ & $0.53 \pm{0.00}$ \\[5pt]
  5            & $0.32$ & $\bm{0.30}$ & $0.31\pm{0.01}$ & $0.32\pm{0.01}$ & $\bm{0.30}\pm{0.00}$ \\
  6            & $0.39$ & $\bm{0.37}$ & $0.39\pm{0.01}$ & $0.41\pm{0.01}$ & $0.38\pm{0.01}$ \\
  7            & $\bm{0.24}$ & $\bm{0.24}$ & $0.25\pm{0.00}$ & $0.28\pm{0.01}$ & $\bm{0.24}\pm{0.00}$ \\
  8            & $0.30$ & $\bm{0.29}$ & $0.33\pm{0.01}$ & $0.33\pm{0.01}$ & $0.31\pm{0.01}$ \\
  9            & $0.27$ & $\bm{0.26}$ & $0.27\pm{0.01}$ & $0.29\pm{0.01}$ & $\bm{0.26}\pm{0.01}$ \\
  10           & $0.37$ & $\bm{0.34}$ & $0.35\pm{0.01}$ & $\bm{0.34}\pm{0.01}$ & $0.35\pm{0.01}$ \\
  11           & $0.39$ & $0.36$ & $0.34\pm{0.01}$ & $0.35\pm{0.01}$ & $\bm{0.33}\pm{0.01}$ \\
  12           & $0.37$ & $\bm{0.36}$ & $0.37\pm{0.01}$ & $0.37\pm{0.01}$ & $\bm{0.36}\pm{0.01}$ \\
  13           & $0.29$ & $0.29$ & $0.29\pm{0.00}$ & $0.30\pm{0.01}$ & $\bm{0.28}\pm{0.00}$ \\ \cmidrule(lr){2-6}
  Bottom-level & $0.33$ & $\bm{0.31}$ & $0.32 \pm{0.00}$ & $0.33 \pm{0.00}$ & $\bm{0.31} \pm{0.00}$ \\[5pt]
  Average      & $0.44$ & $\bm{0.41}$ & $0.43 \pm{0.00}$ & $0.44 \pm{0.00}$ & $0.42 \pm{0.00}$ \\ \bottomrule
 \end{tabular}
 \label{tb:rmse_ind}
\end{table}

\begin{table}[t]
 \centering
 \small
 \caption{Prediction performance for the PstvC dataset.}
 \begin{tabular}{crrrrr} \toprule
               &\multicolumn{5}{c}{RMSE} \\ \cmidrule(lr){2-6} 
  Node $i$     & MA(1) & ES(0.89) & NN+BU & NN+MinT & NN+SR(0.4, 2.4) \\ \midrule
  Root        & $2.69$ & $2.69$ & $2.90\pm{0.05}$ & $\bm{2.48}\pm{0.06}$ & $2.49\pm{0.03}$ \\[5pt]
  2            & $1.20$ & $1.20$ & $1.33\pm{0.03}$ & $\bm{1.06}\pm{0.02}$ & $1.12\pm{0.01}$ \\
  3            & $1.49$ & $1.42$ & $\bm{1.12}\pm{0.01}$ & $1.28\pm{0.03}$ & $1.27\pm{0.02}$ \\
  4            & $1.11$ & $1.11$ & $1.25\pm{0.03}$ & $1.11\pm{0.02}$ & $\bm{1.06}\pm{0.01}$ \\ \cmidrule(lr){2-6}
  Mid-level    & $1.27$ & $1.24$ & $1.23 \pm{0.01}$ & $\bm{1.15} \pm{0.01}$ & $\bm{1.15} \pm{0.01}$  \\[5pt]
  5            & $0.53$ & $0.53$ & $0.56\pm{0.01}$ & $\bm{0.46}\pm{0.01}$ & $0.53\pm{0.01}$ \\
  6            & $0.49$ & $0.49$ & $0.55\pm{0.02}$ & $\bm{0.44}\pm{0.01}$ & $0.51\pm{0.02}$ \\
  7            & $0.46$ & $0.46$ & $0.49\pm{0.01}$ & $0.45\pm{0.01}$ & $\bm{0.43}\pm{0.01}$ \\
  8            & $0.56$ & $0.55$ & $\bm{0.48}\pm{0.01}$ & $0.52\pm{0.01}$ & $0.54\pm{0.03}$ \\
  9            & $0.57$ & $0.54$ & $\bm{0.42}\pm{0.01}$ & $0.54\pm{0.02}$ & $0.53\pm{0.04}$ \\
  10           & $0.49$ & $0.47$ & $\bm{0.43}\pm{0.01}$ & $0.43\pm{0.01}$ & $0.45\pm{0.01}$ \\
  11           & $\bm{0.48}$ & $\bm{0.48}$ & $0.55\pm{0.02}$ & $\bm{0.48}\pm{0.01}$ & $0.49\pm{0.02}$ \\
  12           & $0.59$ & $0.57$ & $0.51\pm{0.01}$ & $\bm{0.44}\pm{0.01}$ & $0.55\pm{0.02}$ \\ 
  13           & $0.44$ & $0.44$ & $0.45\pm{0.02}$ & $0.44\pm{0.01}$ & $\bm{0.43}\pm{0.01}$ \\ \cmidrule(lr){2-6}
  Bottom-level & $0.51$ & $0.50$ & $0.49 \pm{0.00}$ & $\bm{0.47} \pm{0.00}$ & $0.50 \pm{0.00}$ \\[5pt]
  Average      & $0.85$ & $0.84$ & $0.85 \pm{0.00}$ & $\bm{0.78} \pm{0.01}$ & $0.80 \pm{0.00}$ \\ \bottomrule
 \end{tabular}
 \label{tb:rmse_pos}
\end{table}

\begin{figure}[t!]
\centering
\footnotesize
\tabcolsep = 2pt
\begin{tabular}{ccc}
\includegraphics[keepaspectratio, scale=0.2]{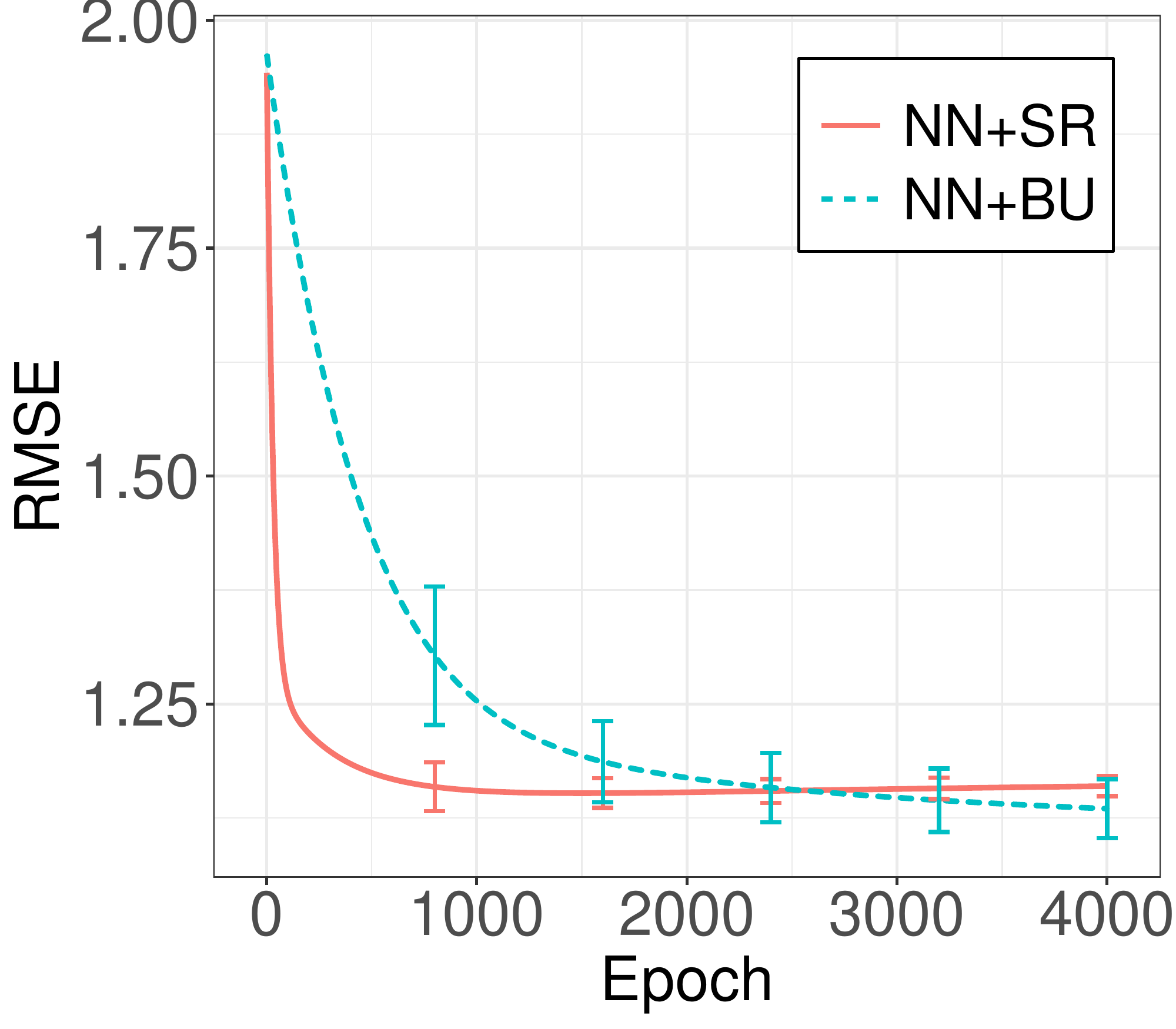} & 
\includegraphics[keepaspectratio, scale=0.2]{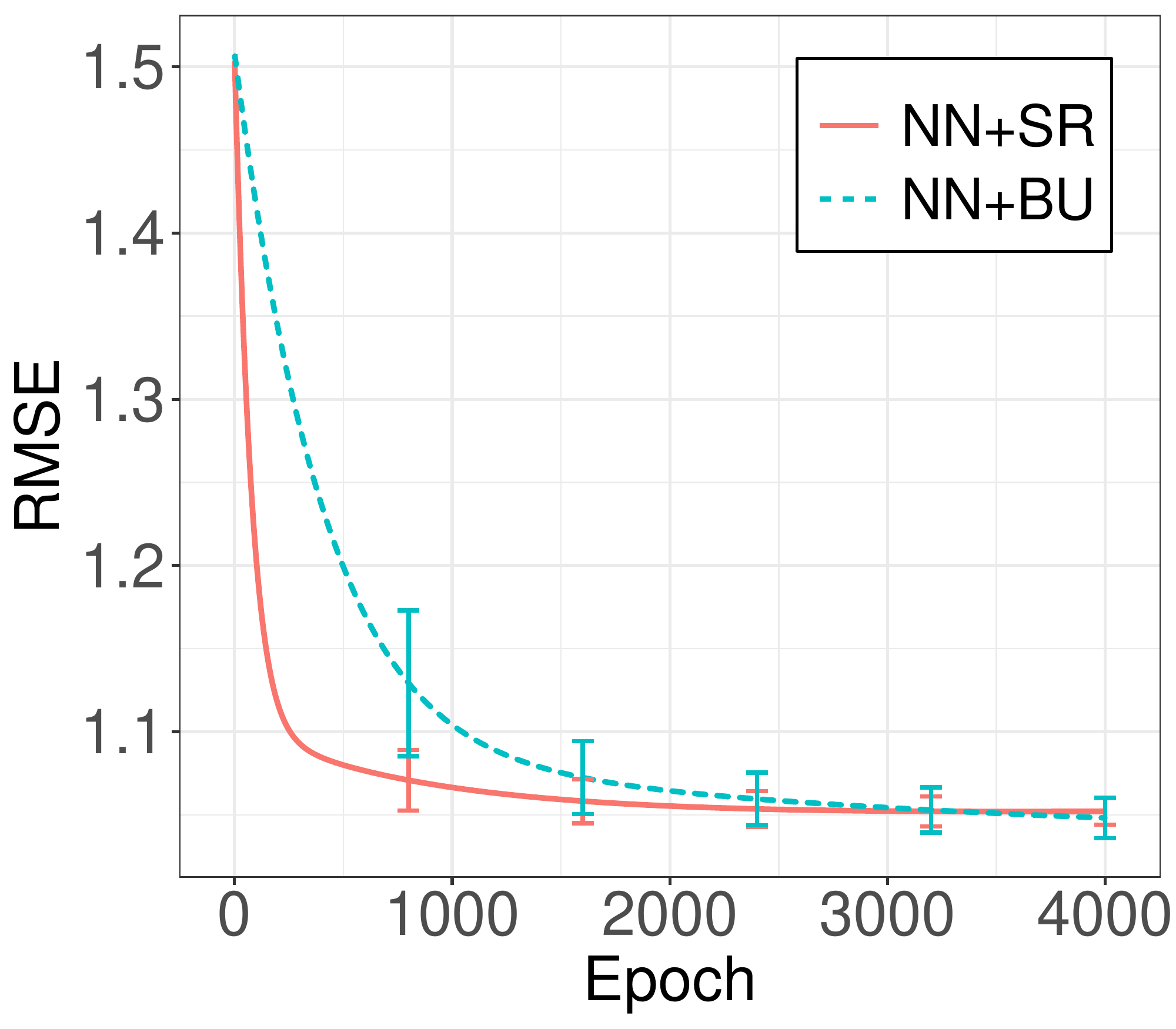} & 
\includegraphics[keepaspectratio, scale=0.2]{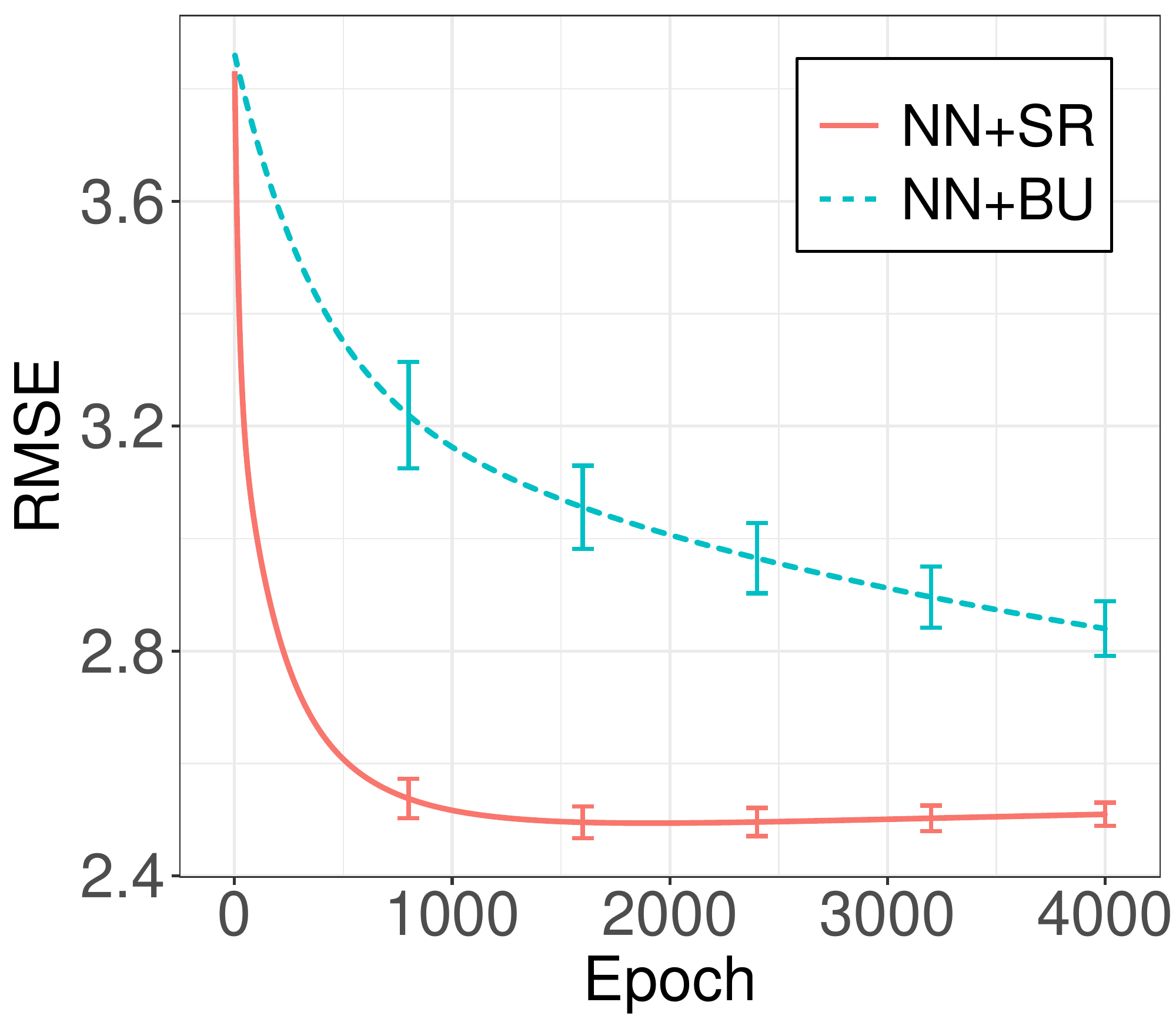} \\
(a) Root (NgtvC) & (b) Root (WeakC) & (c) Root (PstvC) \\[0.3cm]
\includegraphics[keepaspectratio, scale=0.2]{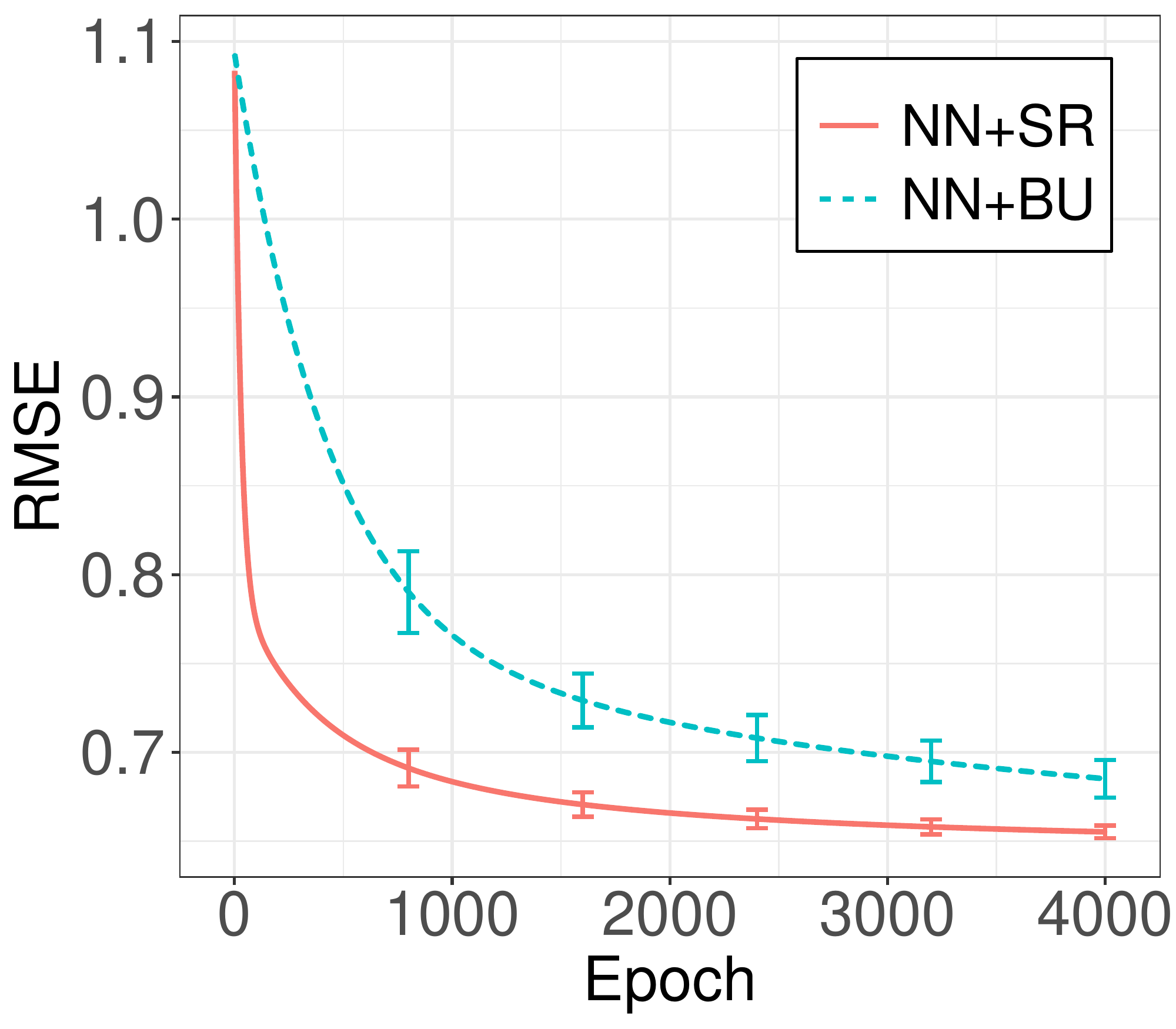} &
\includegraphics[keepaspectratio, scale=0.2]{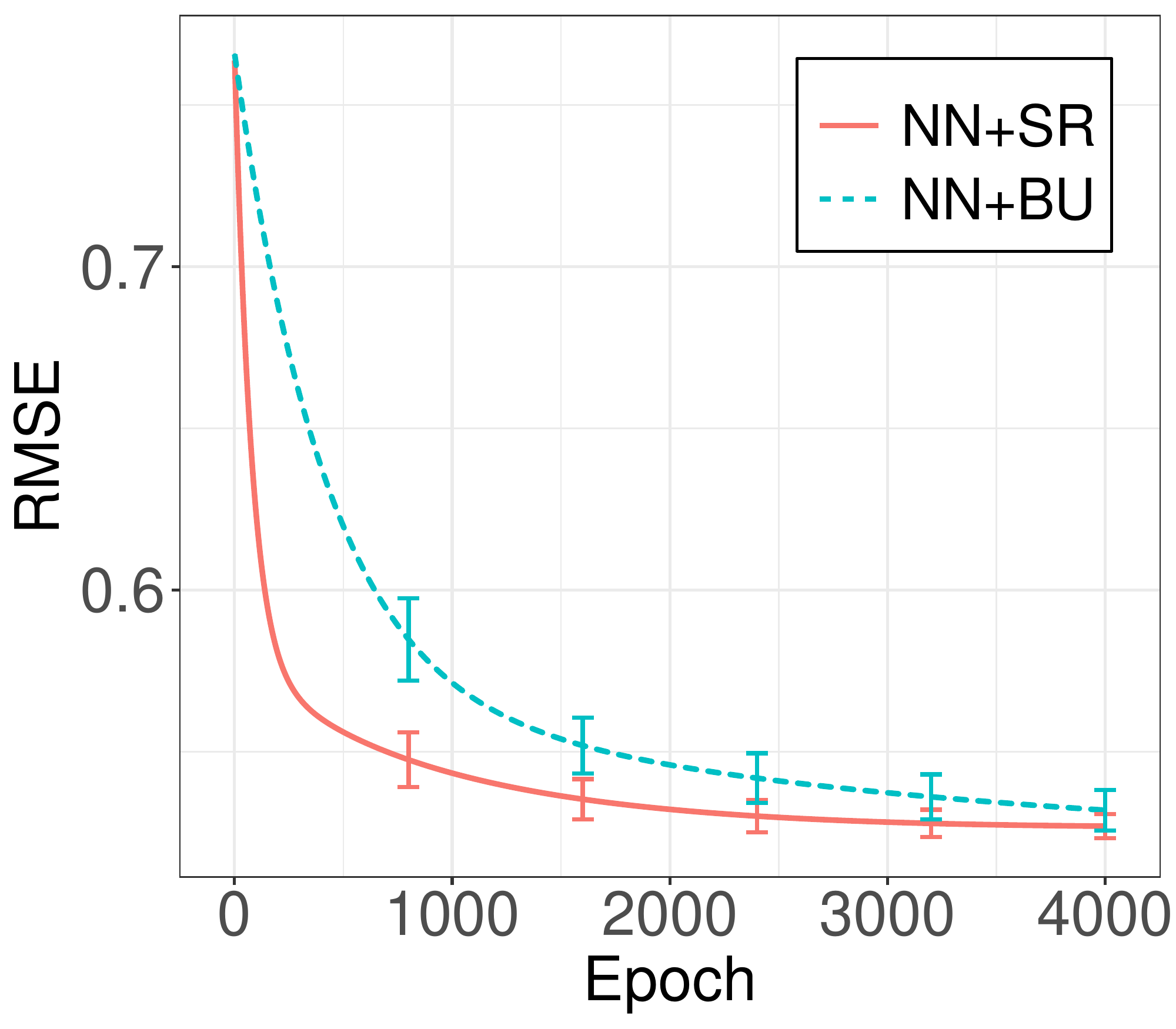} &
\includegraphics[keepaspectratio, scale=0.2]{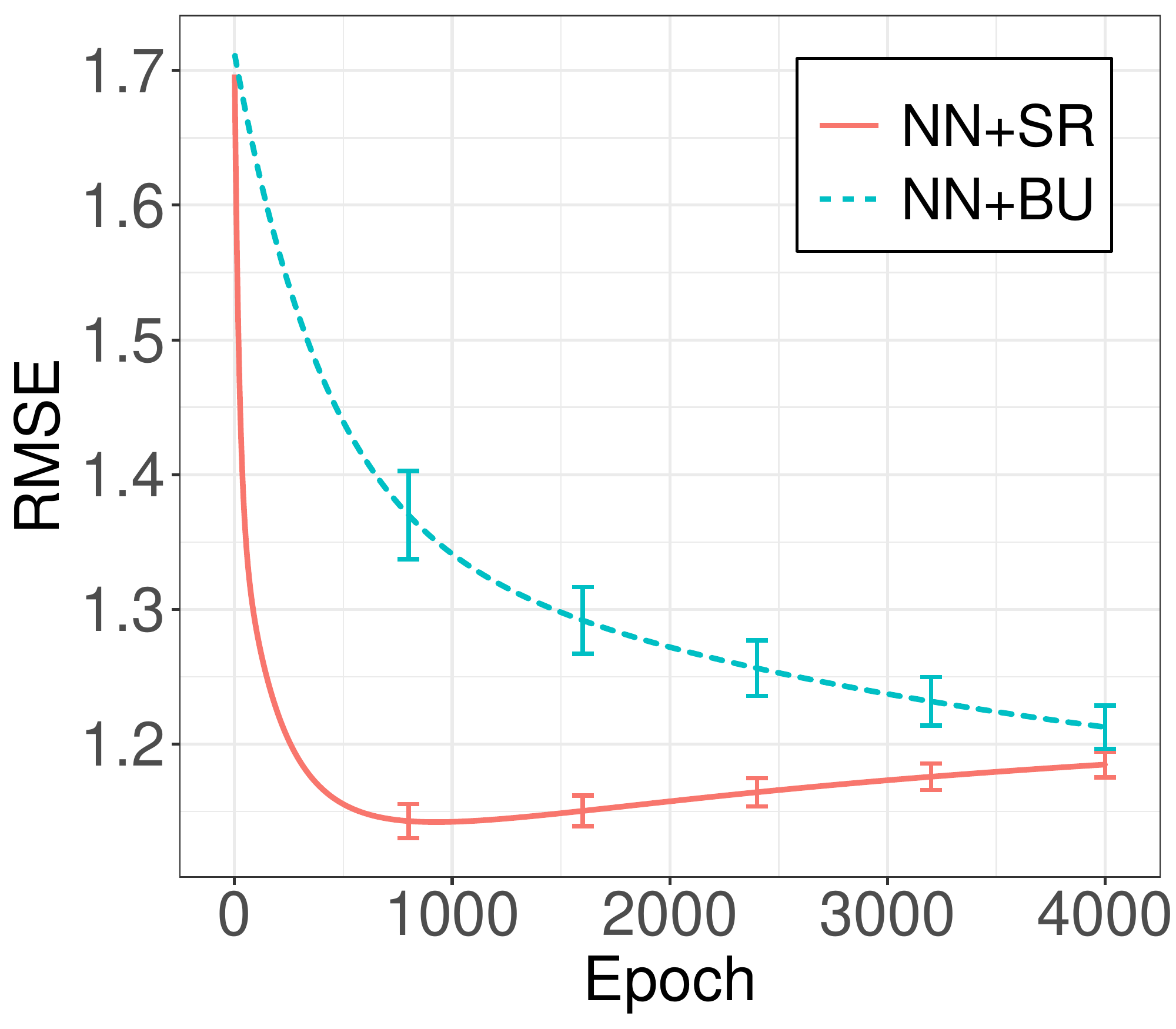} \\
(d) Mid-level (NgtvC) & (e) Mid-level (WeakC) & (f) Mid-level (PstvC) \\[0.3cm]
\includegraphics[keepaspectratio, scale=0.2]{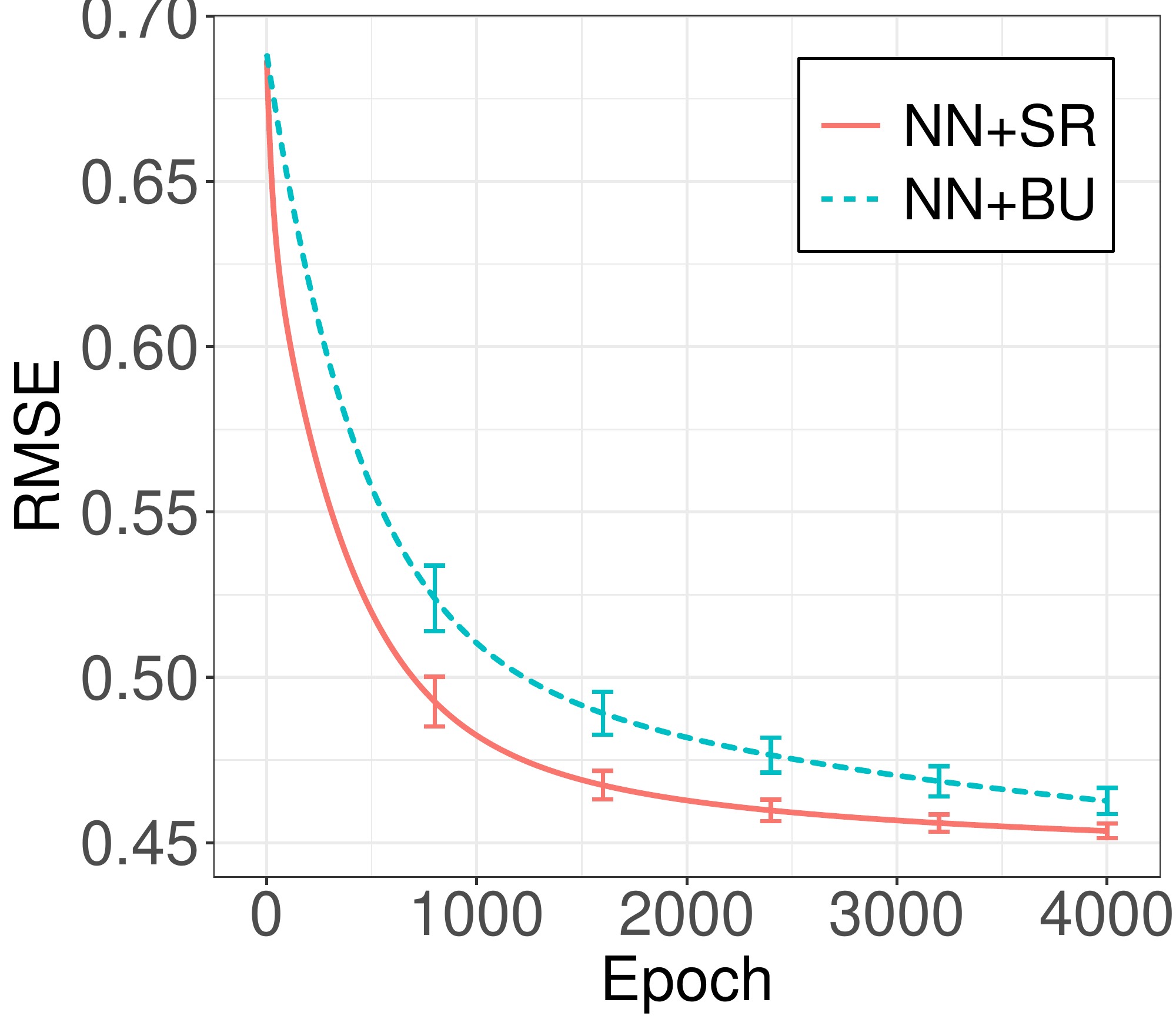} & 
\includegraphics[keepaspectratio, scale=0.2]{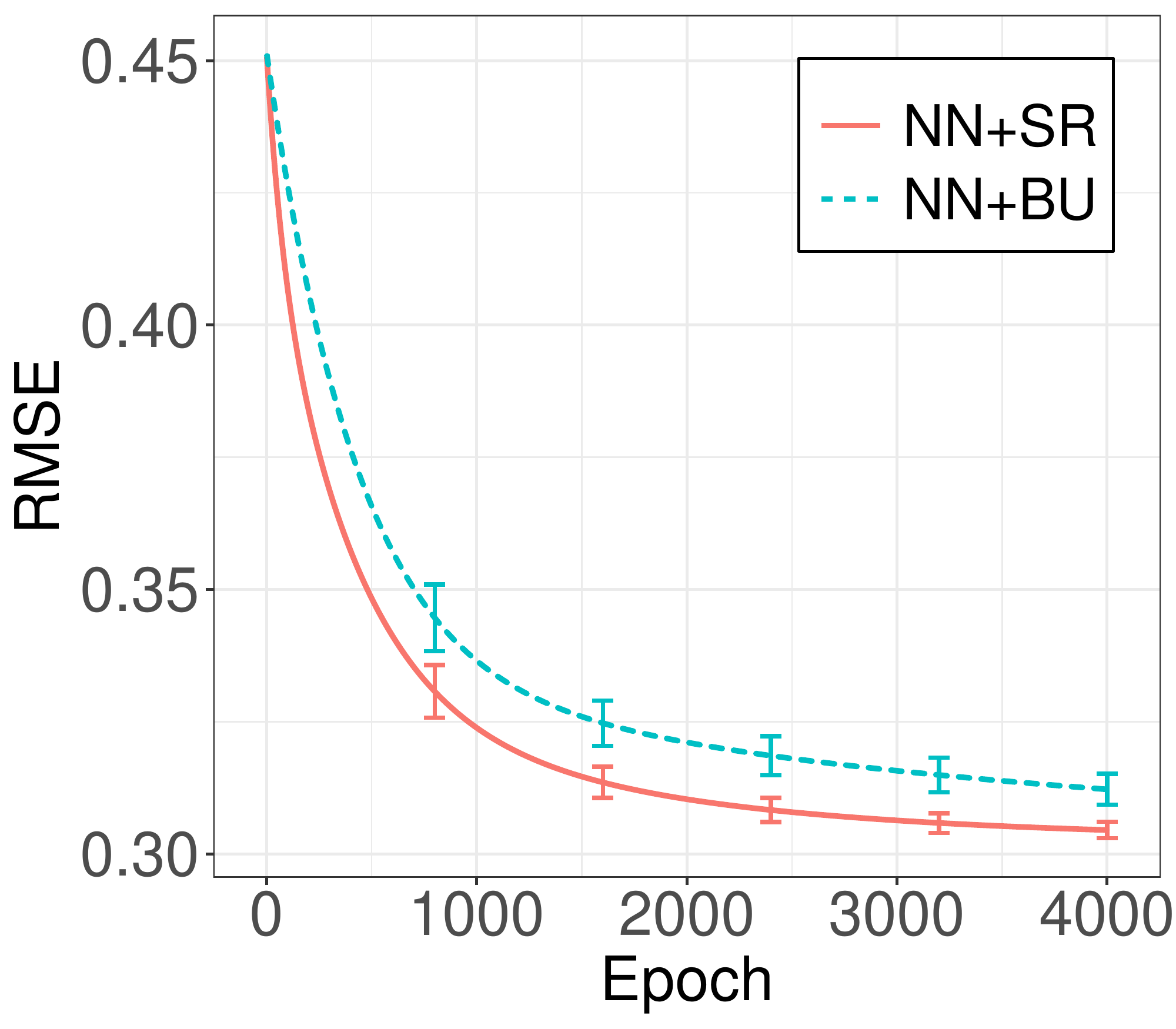} & 
\includegraphics[keepaspectratio, scale=0.2]{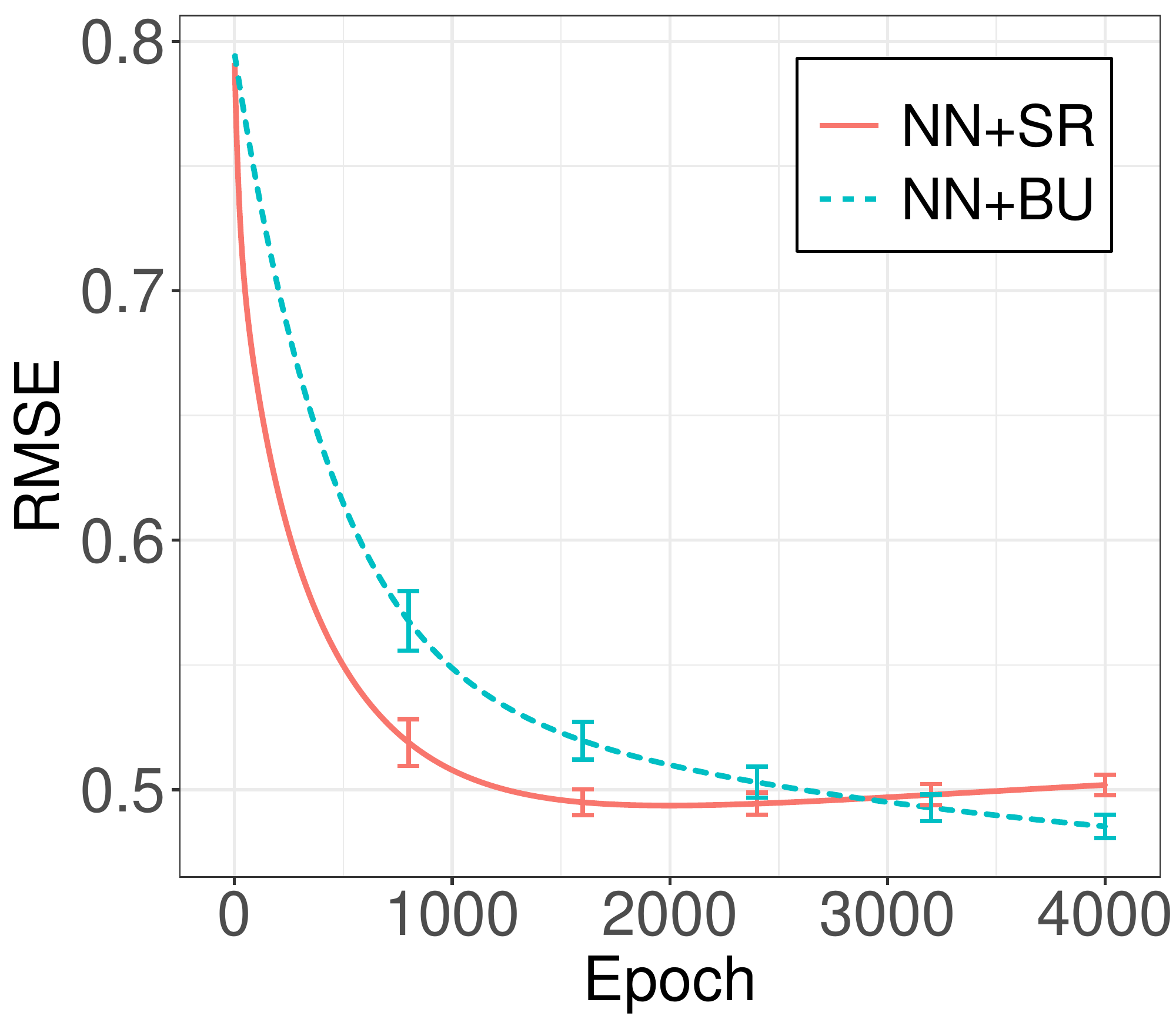} \\
(g) Bottom-level (NgtvC) & (h) Bottom-level (WeakC) & (i) Bottom-level (PstvC) \\[0.3cm]
\includegraphics[keepaspectratio, scale=0.2]{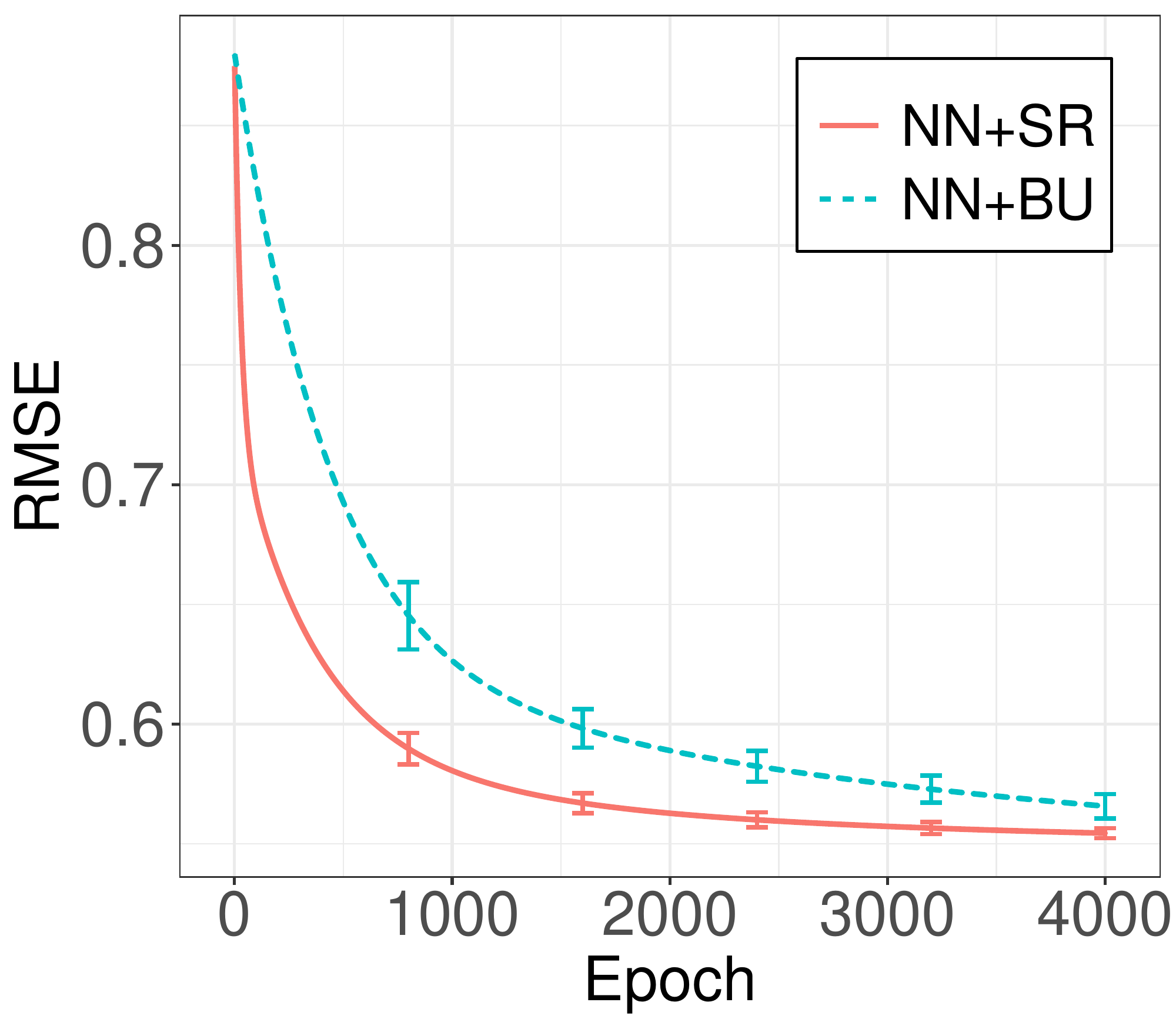} & 
\includegraphics[keepaspectratio, scale=0.2]{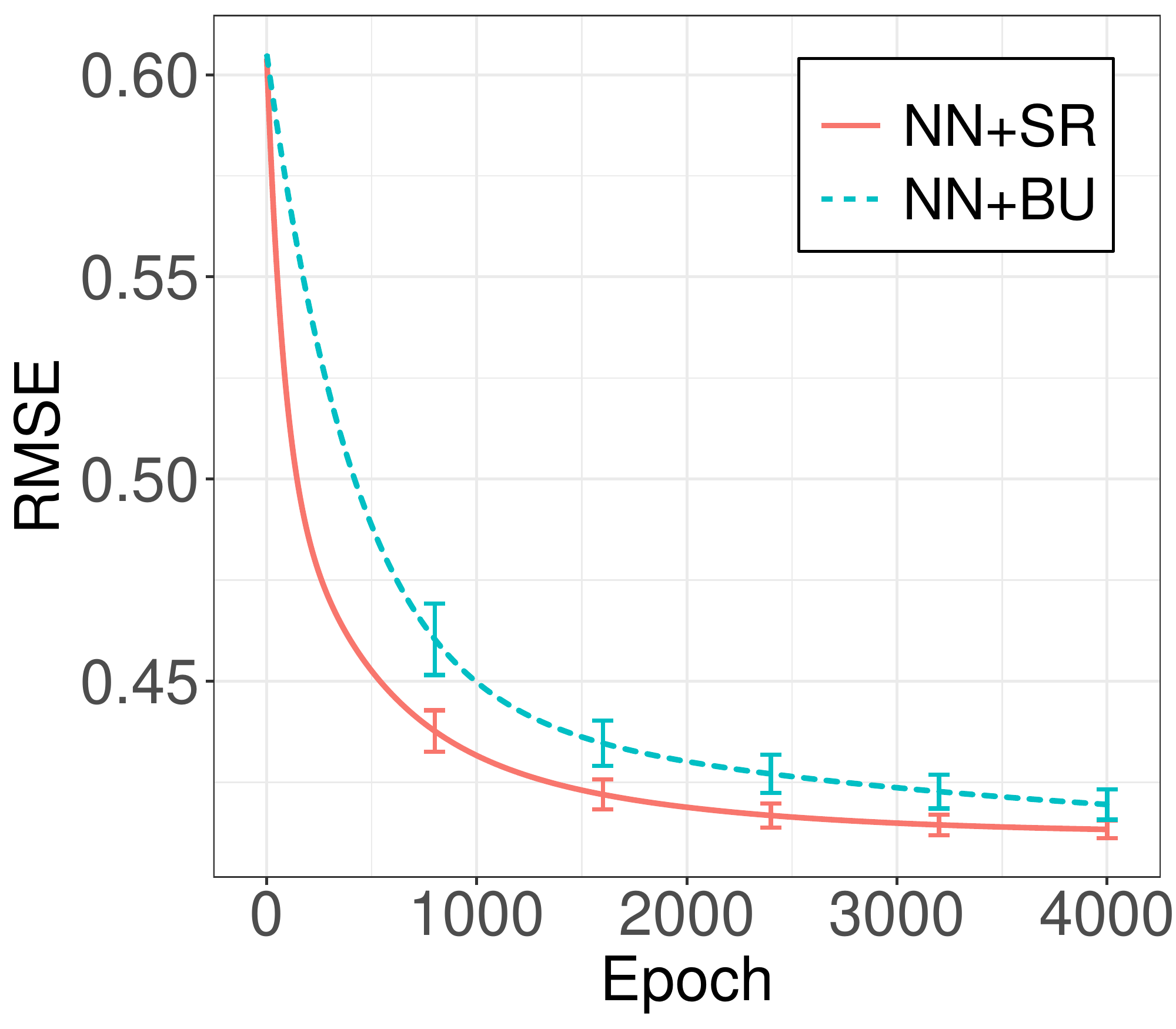} & 
\includegraphics[keepaspectratio, scale=0.2]{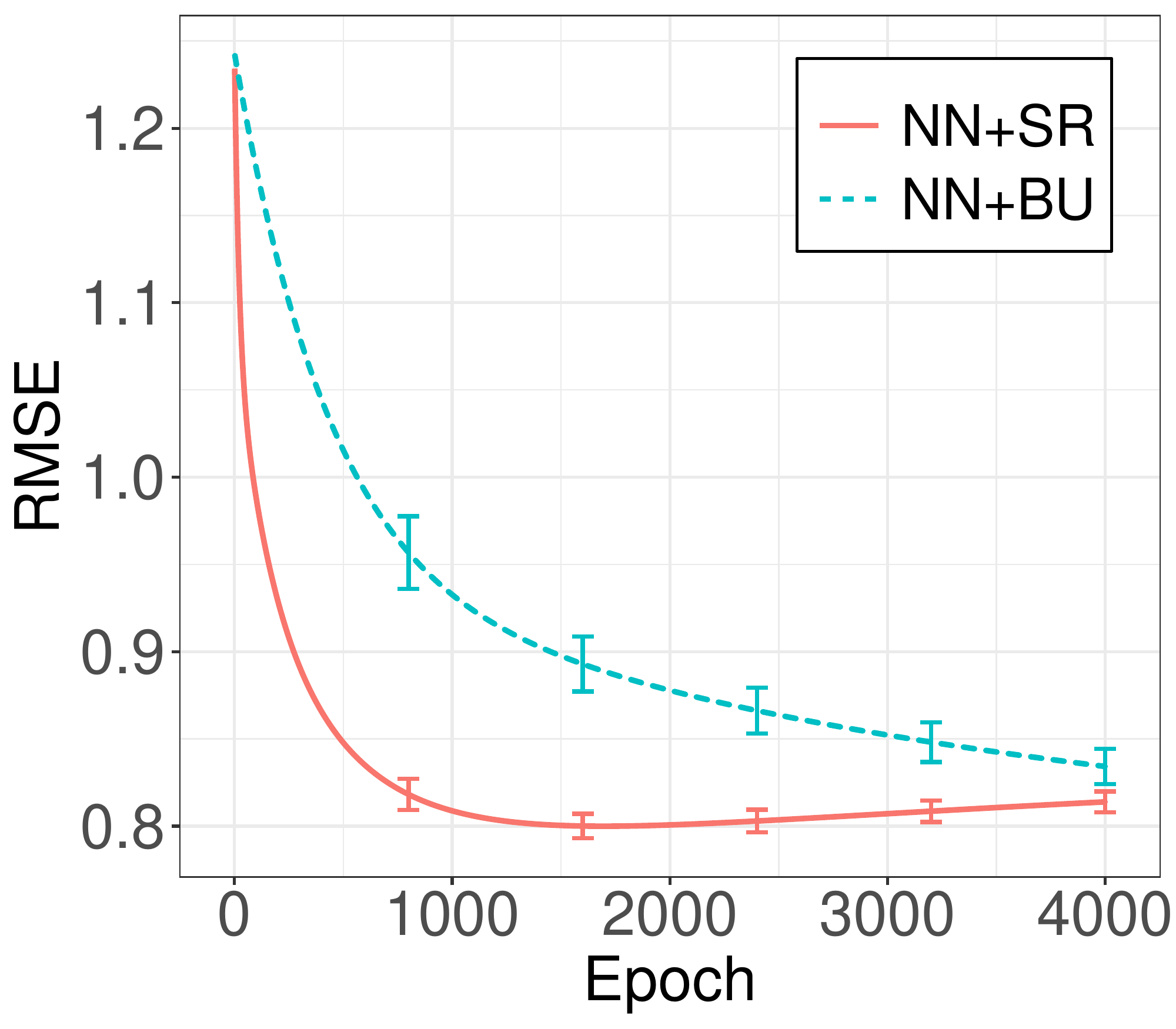} \\
(j) Average (NgtvC) & (k) Average (WeakC) & (l) Average (PstvC) \\[0.3cm]
\end{tabular}
\caption{Convergence performance of the backpropagation algorithm for the synthetic datasets.}
\label{fig:EpoSyn}
\end{figure}

Fig.~\ref{fig:EpoSyn} shows the out-of-sample RMSE values as a function of the epoch (number of iterations) in the backpropagation algorithm for the synthetic datasets. 
RMSEs decreased faster for our structured regularization method NN+SR than for the bottom-up method NN+BU. 
The convergence performance of the two methods greatly differed, especially for the PstvC dataset and upper-level time series. 
Consequently, our structured regularization method improved both prediction accuracy and convergence speed of the backpropagation algorithm. 
This suggests that our method will deliver good prediction performance even if the backpropagation algorithm is terminated in the middle of computation. 

\begin{figure}[t!]
\centering
\footnotesize
\tabcolsep = 2pt
\begin{tabular}{ccc}
\includegraphics[keepaspectratio, scale=0.2]{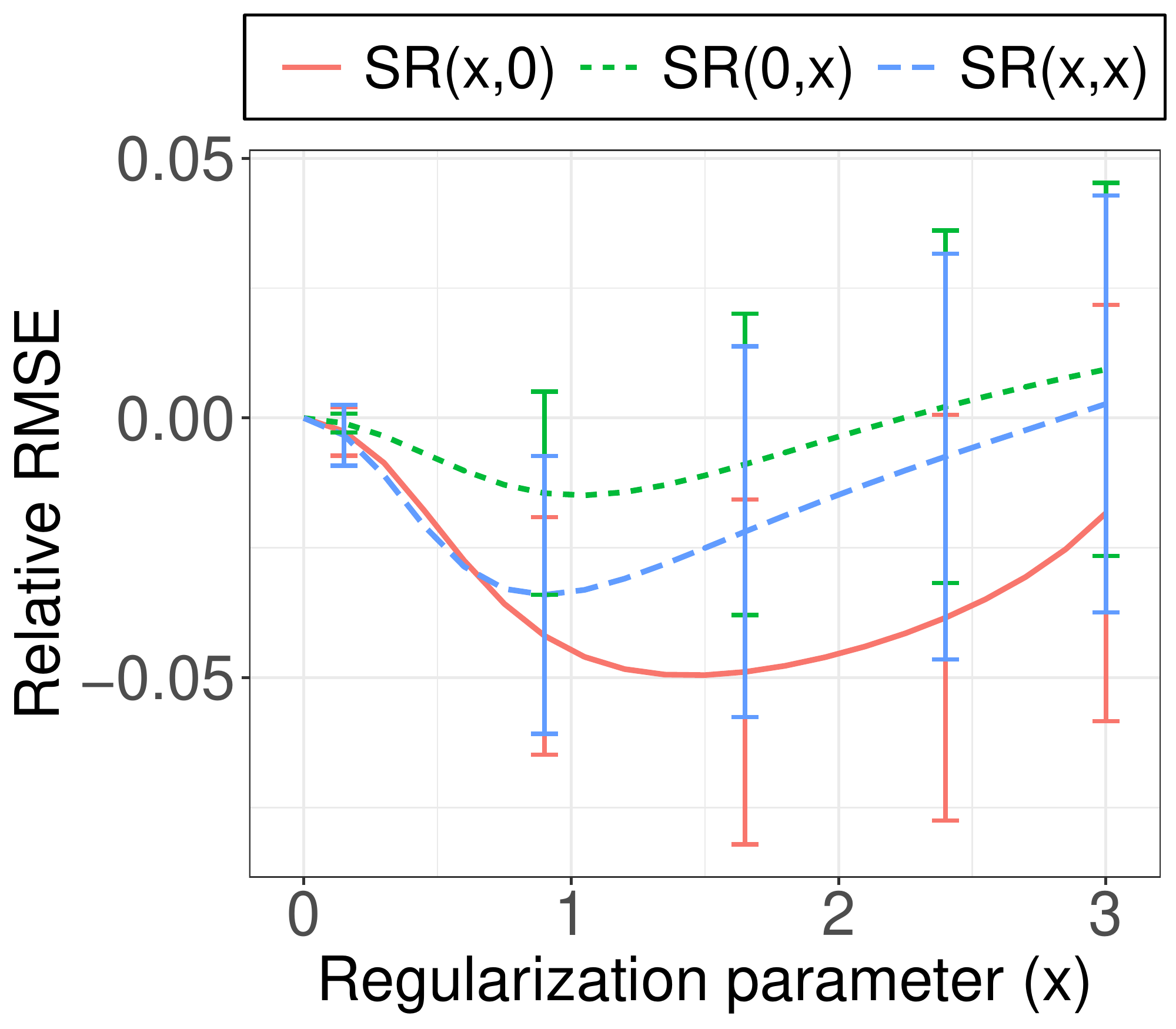} & 
\includegraphics[keepaspectratio, scale=0.2]{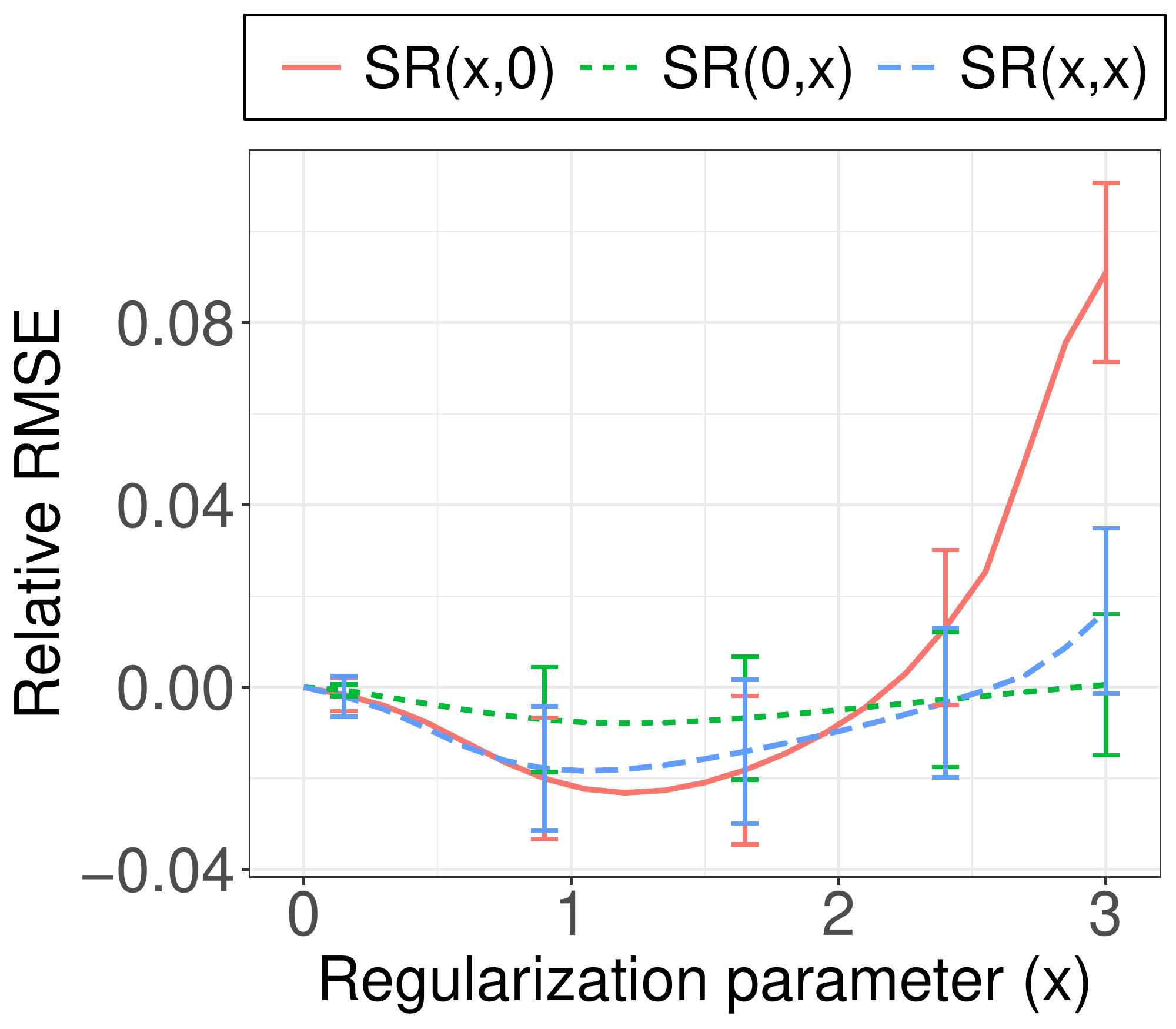} & 
\includegraphics[keepaspectratio, scale=0.2]{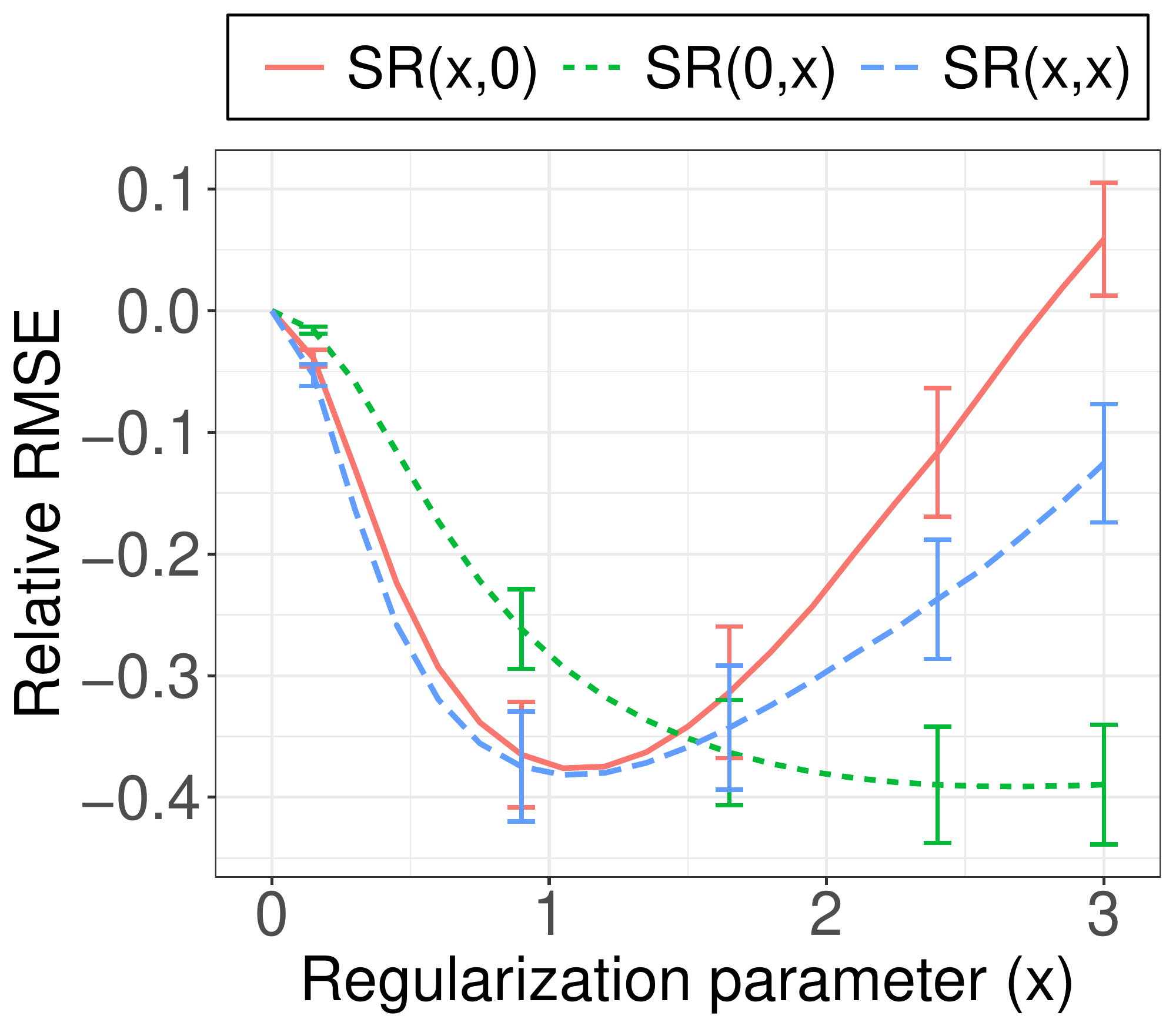} \\
(a) Root (NgtvC) & (b) Root (WeakC) & (c) Root (PstvC) \\[0.3cm]
\includegraphics[keepaspectratio, scale=0.2]{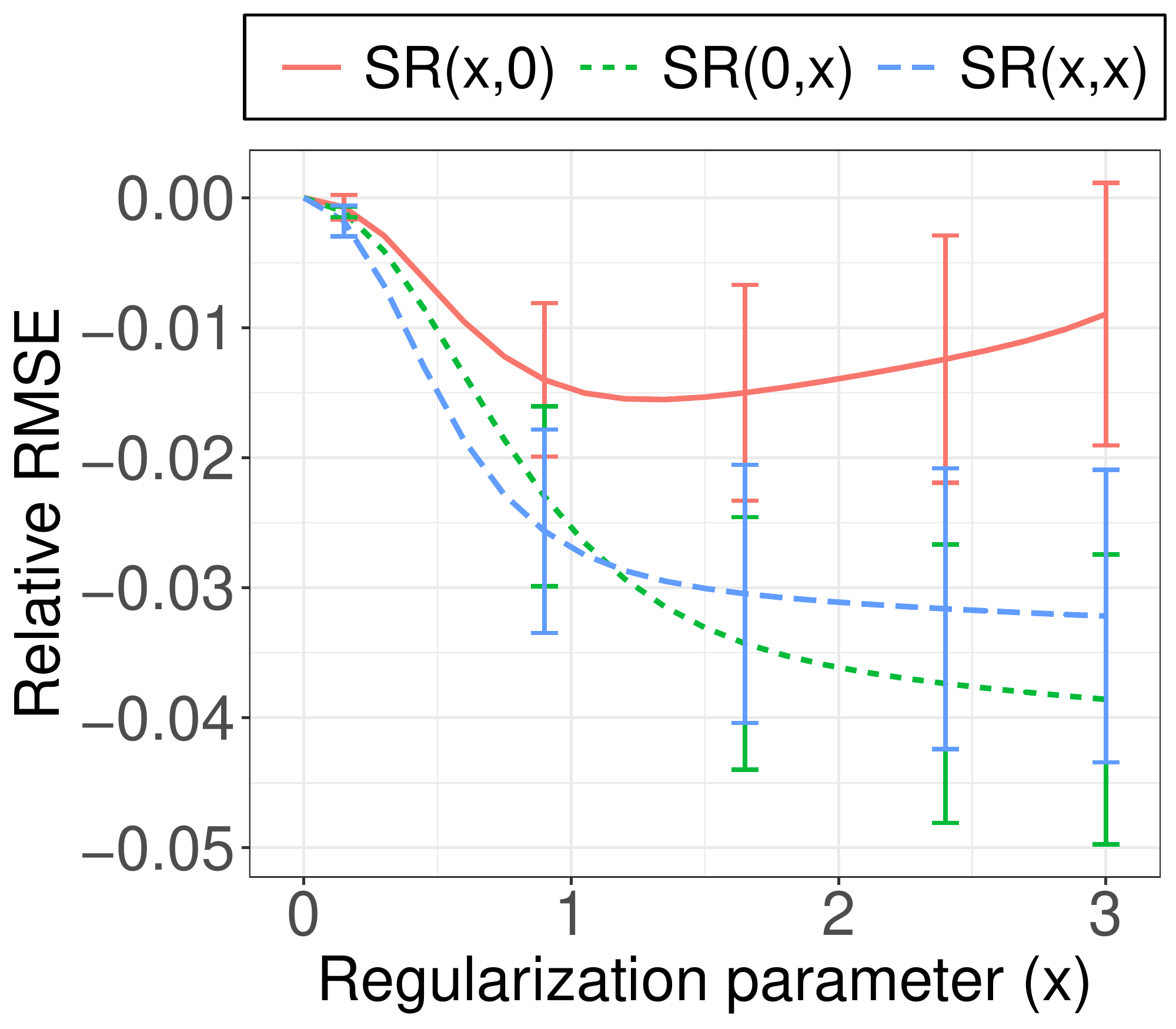} &
\includegraphics[keepaspectratio, scale=0.2]{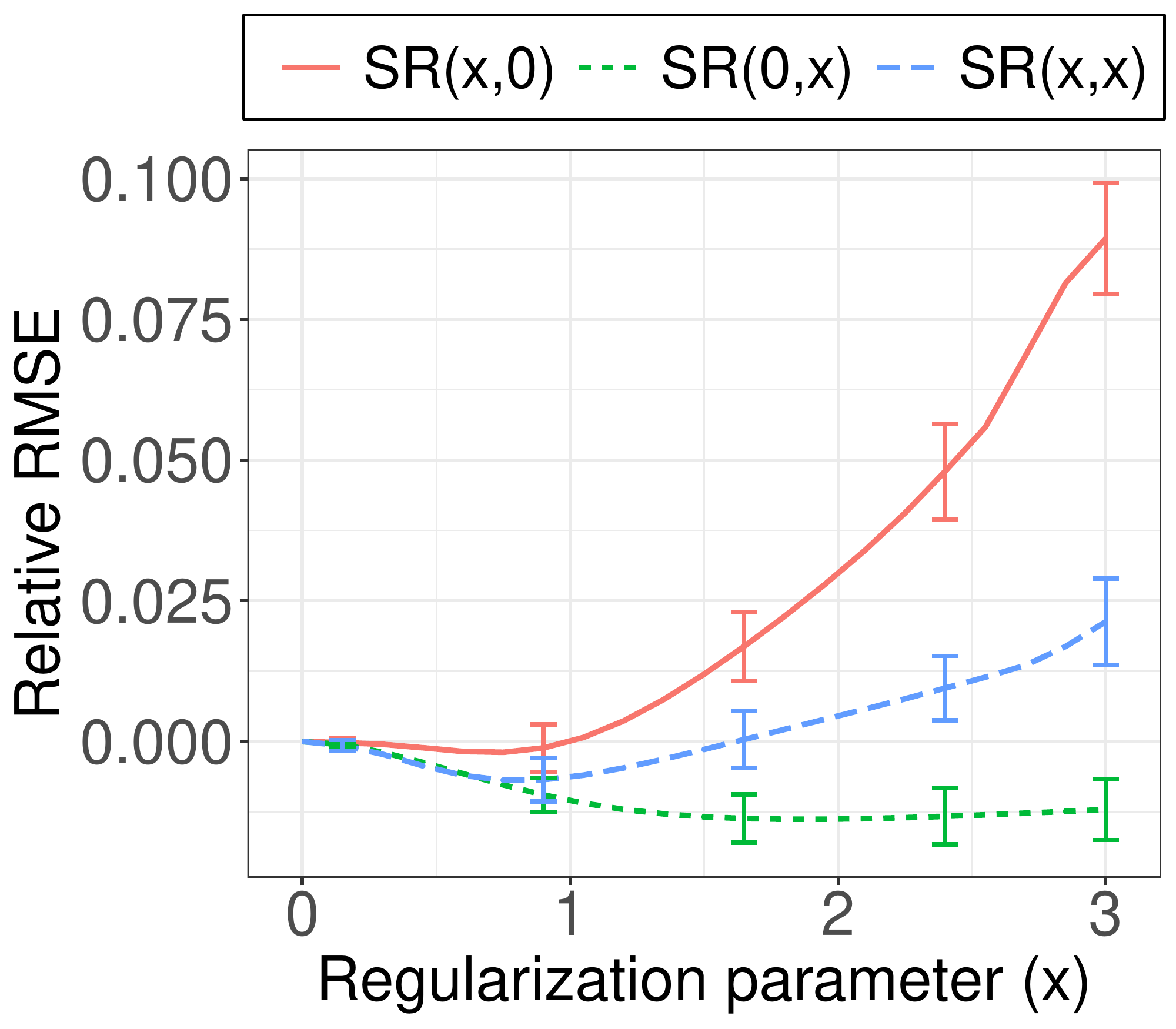} &
\includegraphics[keepaspectratio, scale=0.2]{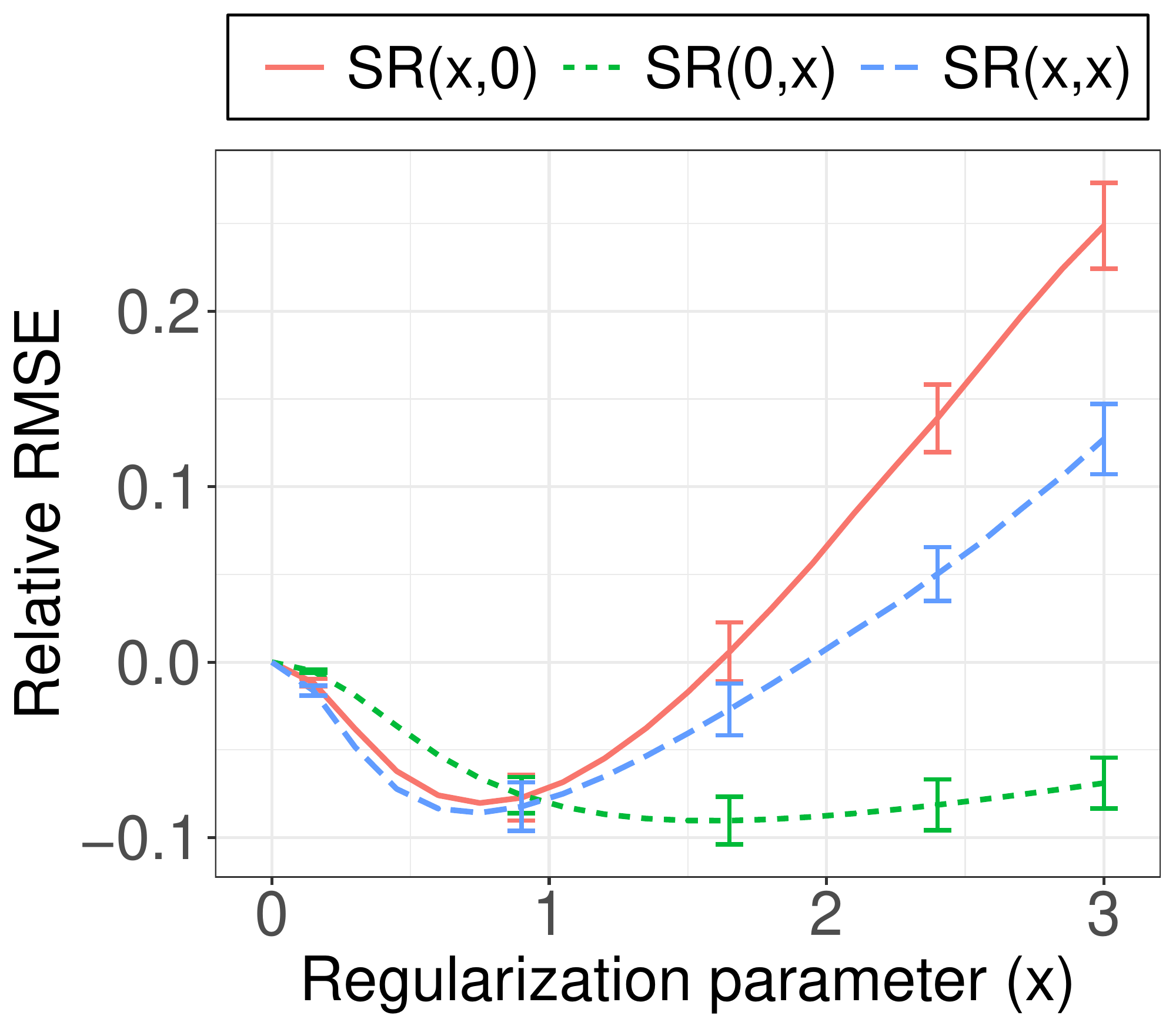} \\
(d) Mid-level (NgtvC) & (e) Mid-level (WeakC) & (f) Mid-level (PstvC) \\[0.3cm]
\includegraphics[keepaspectratio, scale=0.2]{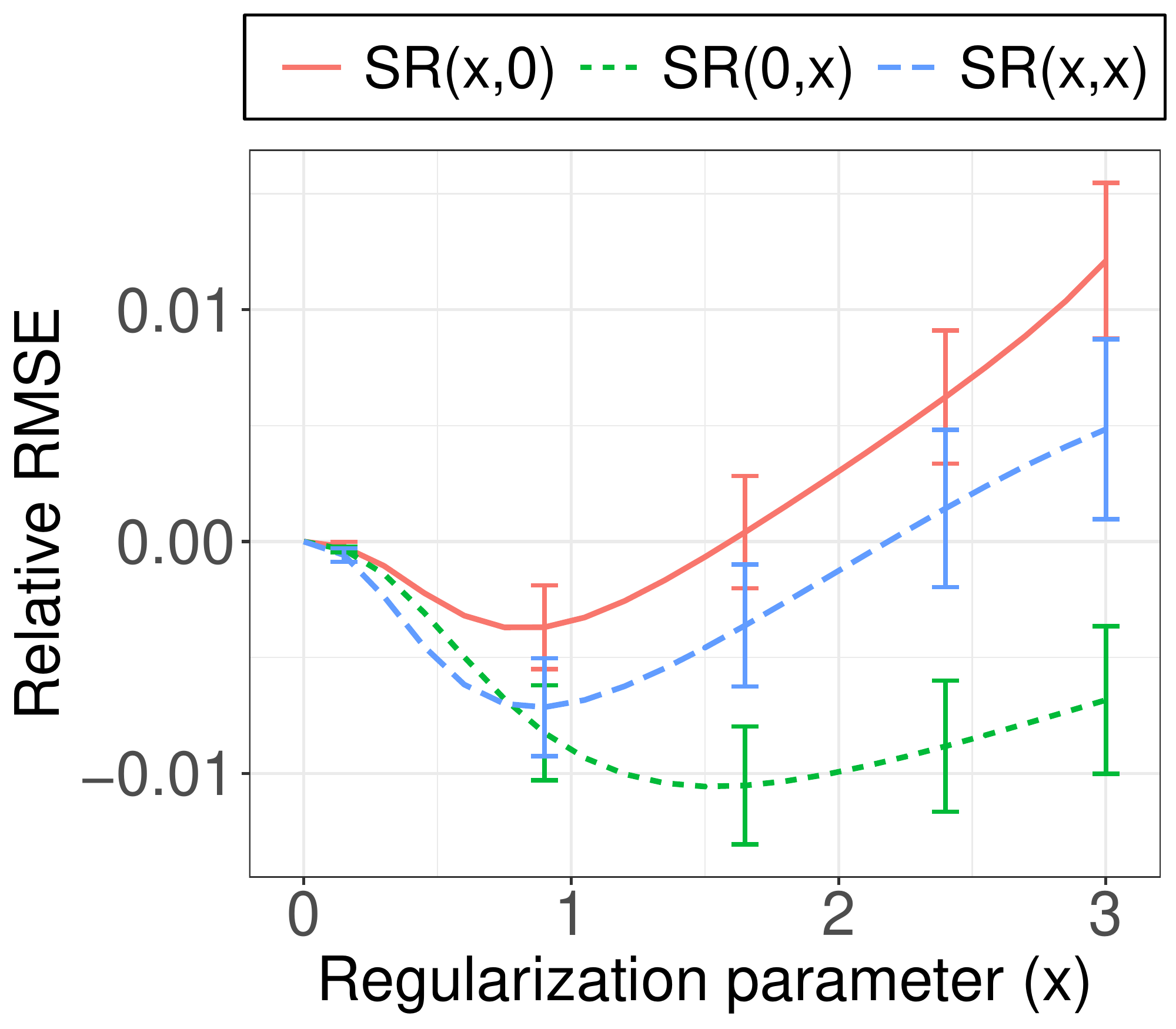} & 
\includegraphics[keepaspectratio, scale=0.2]{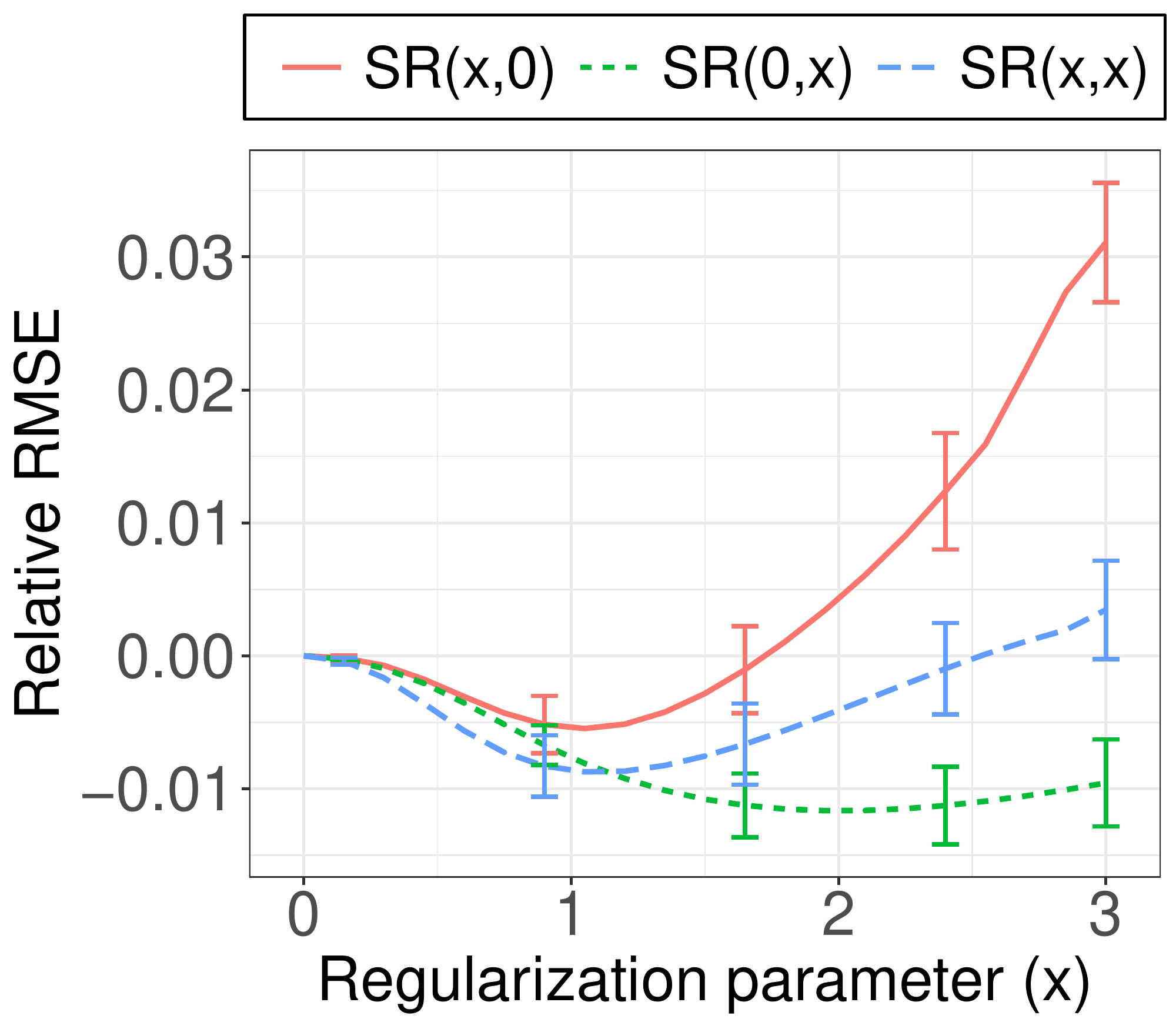} & 
\includegraphics[keepaspectratio, scale=0.2]{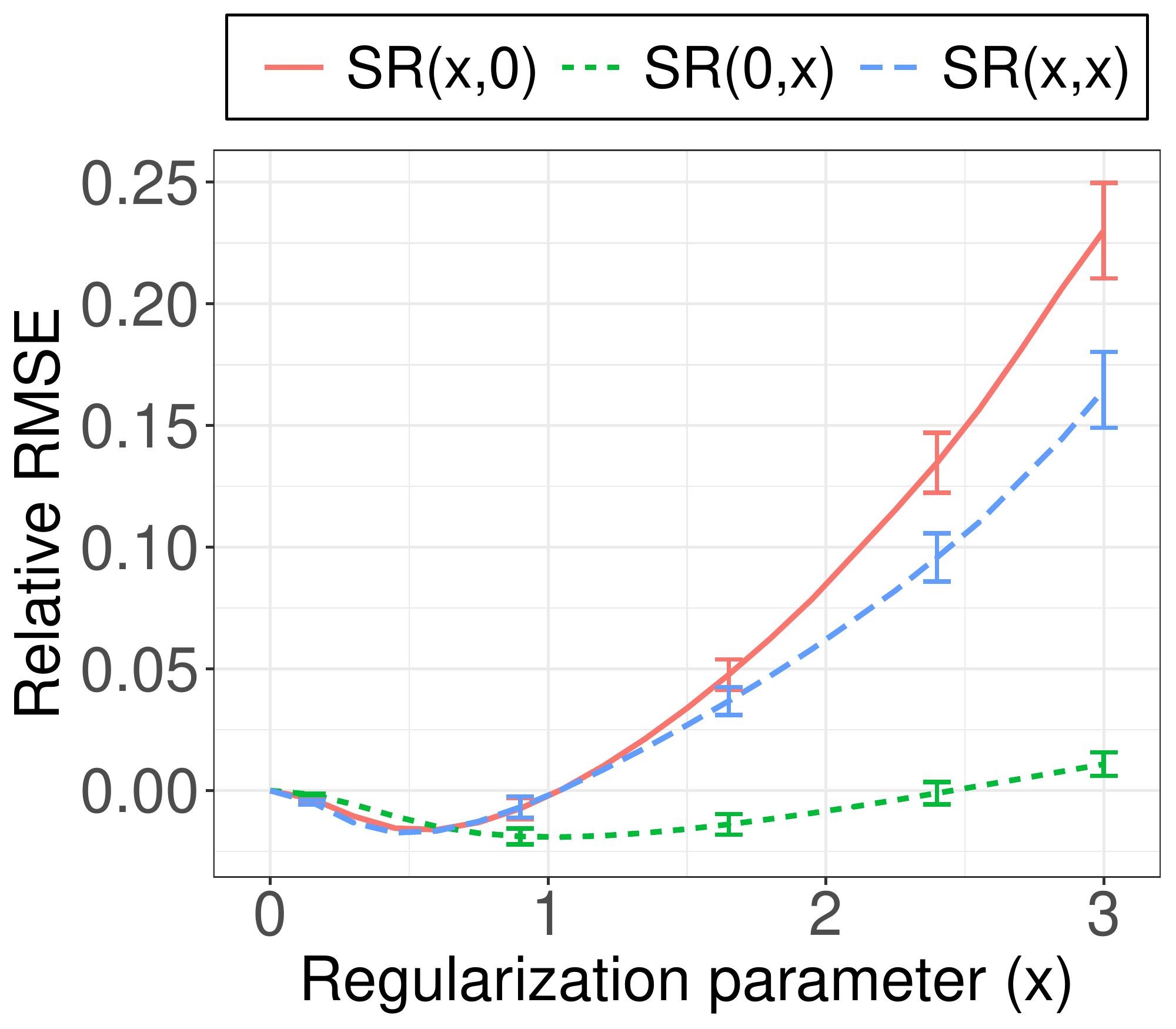} \\
(g) Bottom-level (NgtvC) & (h) Bottom-level (WeakC) & (i) Bottom-level (PstvC) \\[0.3cm]
\includegraphics[keepaspectratio, scale=0.2]{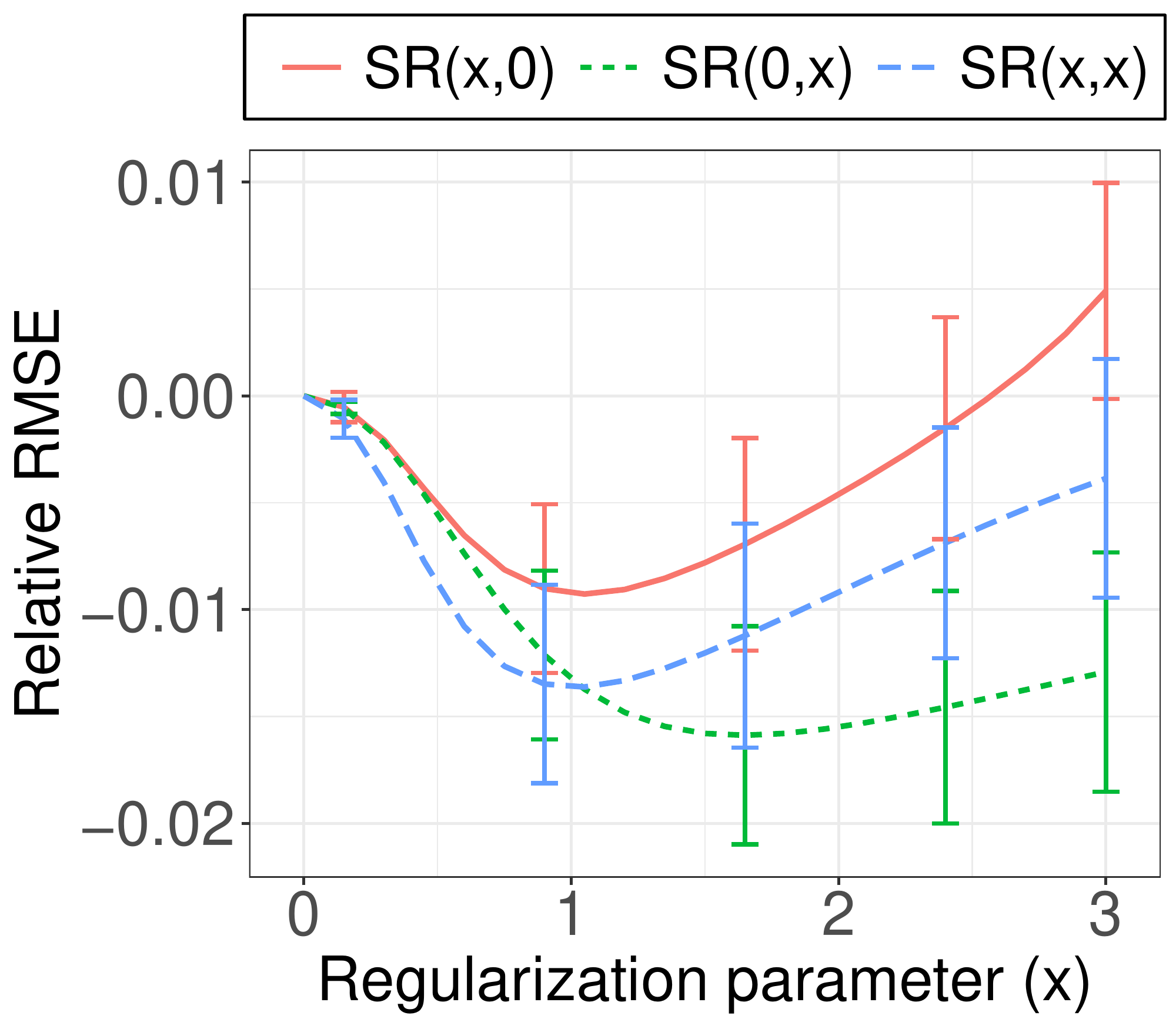} & 
\includegraphics[keepaspectratio, scale=0.2]{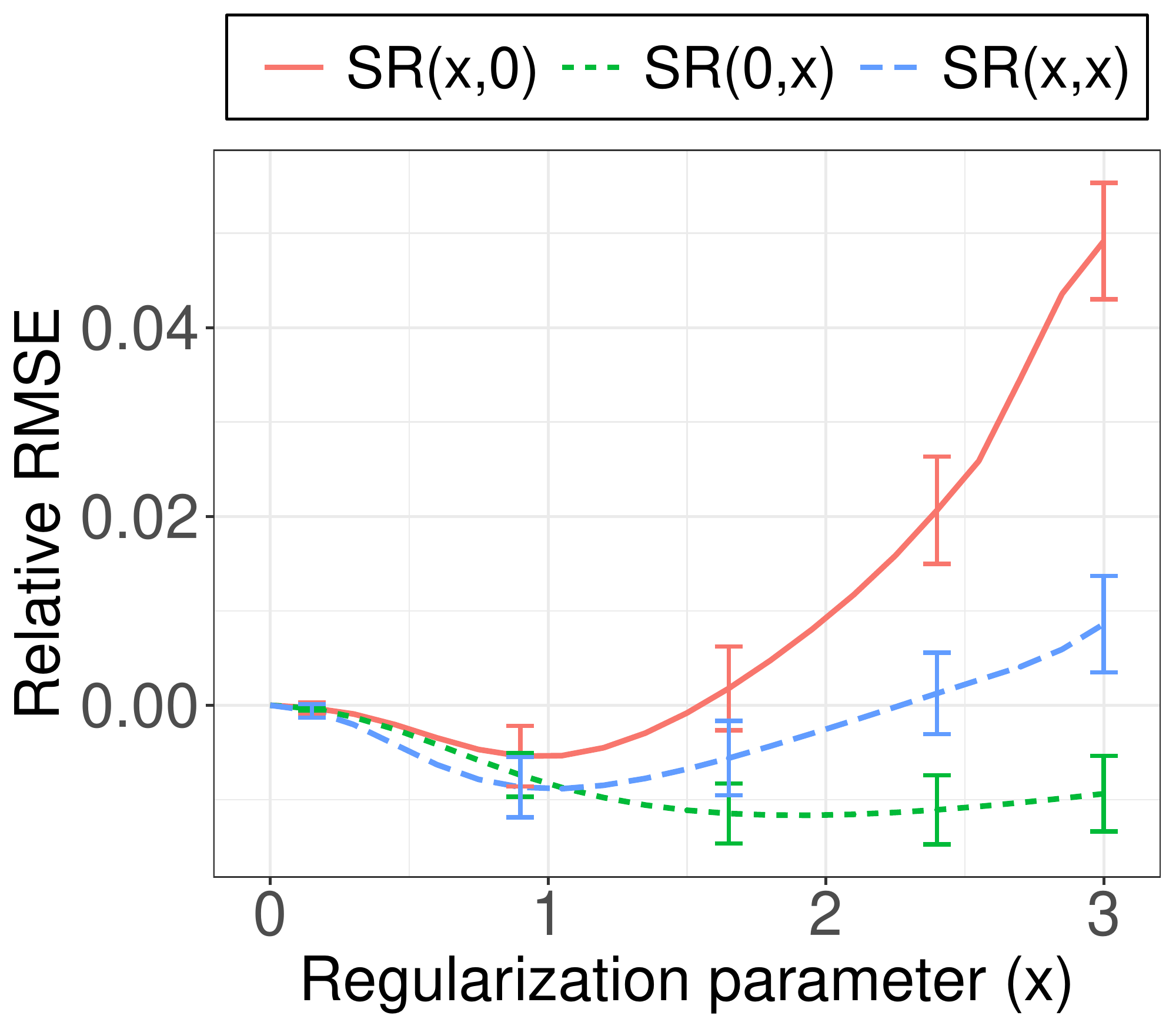} & 
\includegraphics[keepaspectratio, scale=0.2]{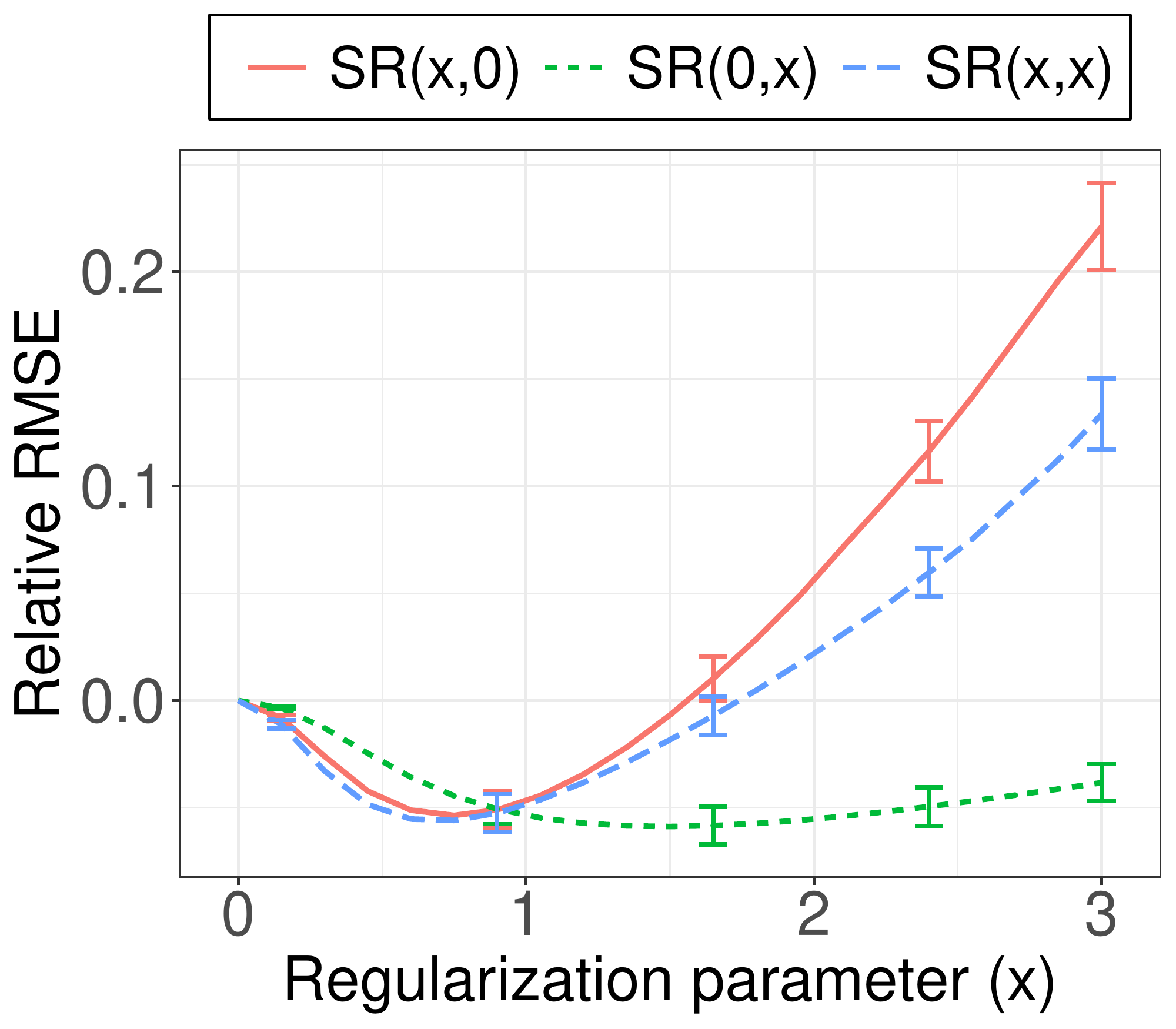} \\
(j) Average (NgtvC) & (k) Average (WeakC) & (l) Average (PstvC) \\[0.3cm]
\end{tabular}
\caption{Effects of structured regularization in the synthetic datasets.}
\label{fig:RegSyn}
\end{figure}

Fig.~\ref{fig:RegSyn} shows the out-of-sample relative RMSE values provided by our structured regularization method NN+SR($\lambda_1,\lambda_M$) for the synthetic datasets. 
This figure shows how regularization for each time series level affects the prediction performance. 
Note that RMSE values were normalized such that the RMSE for $(\lambda_1,\lambda_M) = (0,0)$ was zero in each trial, so the corresponding regularization is effective if this relative RMSE value is negative. 
Relative RMSE values are represented as a function of the regularization parameter value $x$ in Fig.~\ref{fig:RegSyn}. %, where ``NN+SR(\,*\,,\,*\,)'' is abbreviated as ``SR(\,*\,,\,*\,)'' to improve readability.   

NN+SR($x,0$) and NN+SR($0,x$) performed regularization only for the root time series ($\lambda_1 = x$ and $\lambda_M = 0$) and the mid-level time series ($\lambda_1 = 0$ and $\lambda_M = x$), respectively. 
In contrast, NN+SR($x,x$) performed regularization for both using the same regularization parameter values ($\lambda_1 = \lambda_M = x$). 
RMSEs were consistently reduced in the NgtvC dataset.
NN+SR($x,0$) attained relatively small RMSE values only for the root time series, whereas NN+SR($0,x$) delivered the smallest RMSE values for the other time series. 
NN+SR($x,x$) had RMSE values intermediate between NN+SR($x,0$) and NN+SR($0,x$). 
These results suggest that the regularization for the mid-level time series greatly impacted prediction performance. 

\subsection*{Real-world datasets} 
We downloaded historical data describing unemployment rates in Japan from e-Stat, a portal site for official Japanese statistics (\url{https://www.e-stat.go.jp/en}). 
Using these data, we prepared three real-world datasets for Japanese regions: Tohoku, Chubu, and Kansai. 
Table~\ref{tb:pref} lists the prefectures forming the resulting two-level hierarchical structure (Fig.~\ref{fig:hier_sim}). 

We used quarterly statistics (model-based estimates) of unemployment rates during 90 time periods from January 1997 to June 2019, 
taking the first 60 and last 30 time periods as the training and test periods, respectively. 
We used the \texttt{stl} function in the R \texttt{stats} package to remove seasonal and trend components. 
Each time series was standardized according to the mean and variance over the training period. 

\begin{table}[t]
%\centering
\small
\caption{List of prefectures in the real-world datasets.}
\begin{tabular}{cccc} \toprule
         & \multicolumn{3}{c}{Prefectures}  \\ \cmidrule(lr){2-4}
Node $i$ & Tohoku    & Chubu     & Kanasi   \\ \midrule
5        & Aomori    & Niigata   & Mie      \\
6        & Iwate     & Toyama    & Shiga    \\
7        & Miyagi    & Ishikawa  & Kyoto    \\
8        & Akita     & Fukui     & Osaka    \\
9        & Yamagata  & Yamanashi & Hyogo    \\
10       & Fukushima & Nagano    & Nara     \\ 
11       & Ibaraki   & Gifu      & Wakayama \\
12       & Tochigi   & Shizuoka  & Tottori  \\
13       & Gunma     & Mie       & Okayama  \\ \bottomrule
\end{tabular}
\label{tb:pref}
\end{table}

\subsection*{Results for real-world datasets}

Tables~\ref{tb:rmse_tohoku}--\ref{tb:rmse_kansai} list the out-of-sample RMSE values provided by each method for each node in the Tohoku, Chubu, and Kansai datasets. 
For the Tohoku dataset (Table~\ref{tb:rmse_tohoku}), our structured regularization method NN+SR substantially outperformed the other methods. 
For the Chubu dataset (Table~\ref{tb:rmse_chubu}), our method attained average RMSEs that were equally good as those from the exponential smoothing and bottom-up methods, whereas the MinT method showed by far the worst performance. 
For the Kansai dataset (Table~\ref{tb:rmse_kansai}), our method greatly exceeded the prediction performance of the other methods. 
These results demonstrate that our structured regularization method achieved superior performance for the three real-world datasets.

\begin{table}[t]
 \centering
 \small
 \caption{Prediction performance for the Tohoku dataset.}
 \begin{tabular}{crrrrr} \toprule
               &\multicolumn{5}{c}{RMSE} \\ \cmidrule(lr){2-6} 
  Node $i$     & MA(20) & ES(0.12) & NN+BU & NN+MinT & NN+SR(0.4, 1.5) \\ \midrule
  Root         & $6.32$ & $6.45$ & $5.98\pm{0.07}$ & $6.08\pm{0.18}$ & $\bm{5.70}\pm{0.06}$ \\[5pt]
  2            & $2.83$ & $2.91$ & $2.77\pm{0.05}$ & $2.72\pm{0.09}$ & $\bm{2.66}\pm{0.03}$ \\
  3            & $2.06$ & $2.13$ & $2.02\pm{0.05}$ & $2.20\pm{0.06}$ & $\bm{2.01}\pm{0.03}$ \\ 
  4            & $2.86$ & $2.92$ & $2.70\pm{0.04}$ & $2.75\pm{0.07}$ & $\bm{2.63}\pm{0.01}$ \\ \cmidrule(lr){2-6}
  Mid-level    & $2.58$ & $2.65$ & $2.50 \pm{0.01}$ & $2.56 \pm{0.03}$ & $\bm{2.43} \pm{0.01}$ \\[5pt]
  5            & $1.69$ & $1.76$ & $1.68\pm{0.06}$ & $\bm{1.63}\pm{0.06}$ & $1.65\pm{0.05}$  \\
  6            & $0.76$ & $0.77$ & $0.77\pm{0.03}$ & $0.75\pm{0.04}$ & $\bm{0.72}\pm{0.02}$ \\
  7            & $1.15$ & $1.17$ & $1.14\pm{0.04}$ & $1.22\pm{0.07}$ & $\bm{1.11}\pm{0.03}$ \\
  8            & $0.79$ & $0.82$ & $0.79\pm{0.03}$ & $0.83\pm{0.04}$ & $\bm{0.74}\pm{0.02}$ \\
  9            & $0.88$ & $0.91$ & $0.86\pm{0.03}$ & $0.99\pm{0.06}$ & $\bm{0.83}\pm{0.03}$ \\
  10           & $1.01$ & $1.04$ & $1.01\pm{0.03}$ & $1.03\pm{0.04}$ & $\bm{1.00}\pm{0.03}$ \\
  11           & $\bm{1.21}$ & $1.24$ & $\bm{1.21}\pm{0.03}$ & $\bm{1.21}\pm{0.03}$ & $1.25\pm{0.04}$ \\
  12           & $0.90$ & $0.92$ & $\bm{0.88}\pm{0.03}$ & $0.91\pm{0.02}$ & $0.89\pm{0.02}$ \\
  13           & $0.98$ & $1.00$ & $0.94\pm{0.02}$ & $\bm{0.92}\pm{0.03}$ & $0.94\pm{0.02}$ \\ \cmidrule(lr){2-6}
  Bottom-level & $1.04$ & $1.07$ & $1.03 \pm{0.01}$ & $1.05 \pm{0.01}$ & $\bm{1.01} \pm{0.00}$ \\[5pt]
  Average      & $1.80$ & $1.85$ & $1.75 \pm{0.01}$ & $1.79 \pm{0.02}$ & $\bm{1.70} \pm{0.01}$ \\ \bottomrule
 \end{tabular}
 \label{tb:rmse_tohoku}
\end{table}

\begin{table}[t]
 \centering
 \small
 \caption{Prediction performance for the Chubu dataset.}
 \begin{tabular}{crrrrr} \toprule
               &\multicolumn{5}{c}{RMSE} \\ \cmidrule(lr){2-6} 
  Node $i$     & MA(16) & ES(0.03) & NN+BU & NN+MinT & NN+SR(0.4, 0.6) \\ \midrule
  Root         & $4.11$ & $4.09$ & $3.99\pm{0.04}$ & $4.13\pm{0.14}$ & $\bm{3.97}\pm{0.03}$ \\[5pt]
  2            & $1.77$ & $1.75$ & $1.72\pm{0.03}$ & $\bm{1.70}\pm{0.03}$ & $1.72\pm{0.03}$ \\
  3            & $1.38$ & $1.37$ & $1.37\pm{0.03}$ & $1.56\pm{0.06}$ & $\bm{1.36}\pm{0.03}$ \\
  4            & $2.19$ & $\bm{2.17}$ & $2.18\pm{0.03}$ & $2.34\pm{0.20}$  & $2.18\pm{0.03}$ \\ \cmidrule(lr){2-6}
  Mid-level    & $1.78$ & $1.76$ & $1.76 \pm{0.01}$ & $1.87 \pm{0.04}$ & $\bm{1.75} \pm{0.01}$ \\[5pt]
  5            & $0.80$ & $\bm{0.79}$ & $0.81\pm{0.03}$ & $\bm{0.79}\pm{0.03}$ & $0.81\pm{0.03}$ \\
  6            & $0.67$ & $\bm{0.65}$ & $\bm{0.65}\pm{0.03}$ & $0.67\pm{0.03}$ & $0.66\pm{0.03}$ \\
  7            & $0.76$ & $0.76$ & $\bm{0.74}\pm{0.03}$ & $0.75\pm{0.03}$ & $\bm{0.74}\pm{0.03}$ \\
  8            & $0.82$ & $0.81$ & $0.77\pm{0.02}$ & $0.78\pm{0.03}$ & $\bm{0.76}\pm{0.02}$ \\
  9            & $0.71$ & $0.70$ & $0.68\pm{0.02}$ & $0.72\pm{0.03}$ & $\bm{0.67}\pm{0.02}$ \\
  10           & $0.95$ & $\bm{0.97}$ & $0.99\pm{0.02}$ & $1.14\pm{0.04}$ & $0.98\pm{0.02}$ \\
  11           & $0.81$ & $\bm{0.80}$ & $0.88\pm{0.04}$ & $1.09\pm{0.14}$ & $0.89\pm{0.04}$ \\
  12           & $0.99$ & $0.98$ & $0.98\pm{0.03}$ & $1.05\pm{0.07}$ & $\bm{0.97}\pm{0.03}$ \\
  13           & $\bm{0.73}$ & $\bm{0.73}$ & $0.75\pm{0.02}$ & $0.78\pm{0.04}$ & $0.77\pm{0.02}$ \\ \cmidrule(lr){2-6}
  Bottom-level & $\bm{0.80}$ & $\bm{0.80}$ & $0.81 \pm{0.00}$ & $0.86 \pm{0.01}$ & $0.81 \pm{0.00}$ \\[5pt]
  Average      & $1.28$ & $\bm{1.27}$ & $\bm{1.27} \pm{0.00}$ & $1.35 \pm{0.02}$ & $\bm{1.27} \pm{0.00}$ \\ \bottomrule
 \end{tabular}
 \label{tb:rmse_chubu}
\end{table}

\begin{table}[t]
 \centering
 \small
 \caption{Prediction performance for the Kansai dataset.}
 \begin{tabular}{crrrrr} \toprule
               &\multicolumn{5}{c}{RMSE} \\ \cmidrule(lr){2-6} 
  Node $i$     & MA(18) & ES(0.05) & NN+BU & NN+MinT & NN+SR(0.4, 1.2) \\ \midrule
  Root         & $13.88$ & $13.84$ & $13.60\pm{0.68}$ & $14.40\pm{0.54}$ & $\bm{12.20}\pm{0.39}$ \\[5pt]
  2            & $2.57$ & $2.56$ & $2.58\pm{0.09}$ & $2.49\pm{0.08}$ & $\bm{2.37}\pm{0.04}$ \\
  3            & $12.78$ & $12.79$ & $12.56\pm{0.69}$ & $13.31\pm{0.54}$ & $\bm{11.14}\pm{0.41}$ \\
  4            & $1.90$ & $1.90$ & $1.83\pm{0.04}$ & $1.78\pm{0.06}$ & $\bm{1.68}\pm{0.03}$ \\ \cmidrule(lr){2-6}
  Mid-level    & $5.75$ & $5.75$ & $5.66 \pm{0.12}$ & $5.86 \pm{0.09}$ & $\bm{5.06} \pm{0.07}$ \\[5pt]
  5            & $0.73$ & $0.74$ & $0.77\pm{0.03}$ & $\bm{0.69}\pm{0.03}$ & $0.79\pm{0.02}$ \\
  6            & $\bm{1.80}$ & $1.82$ & $1.87\pm{0.07}$ & $1.82\pm{0.04}$ & $1.83\pm{0.04}$ \\
  7            & $1.35$ & $1.36$ & $1.33\pm{0.07}$ & $1.49\pm{0.06}$ & $\bm{1.22}\pm{0.04}$ \\
  8            & $11.31$ & $11.34$ & $11.29\pm{0.66}$ & $12.44\pm{0.57}$ & $\bm{10.02}\pm{0.39}$ \\
  9            & $2.71$ & $2.69$ & $2.62\pm{0.14}$ & $2.50\pm{0.09}$ & $\bm{2.43}\pm{0.10}$ \\
  10           & $1.50$ & $1.49$ & $1.48\pm{0.07}$ & $\bm{1.41}\pm{0.06}$ & $1.43\pm{0.06}$ \\
  11           & $1.16$ & $1.14$ & $1.14\pm{0.04}$ & $1.15\pm{0.07}$ & $\bm{1.03}\pm{0.03}$ \\
  12           & $0.82$ & $0.82$ & $0.79\pm{0.02}$ & $0.86\pm{0.03}$ & $\bm{0.78}\pm{0.01}$ \\
  13           & $0.99$ & $0.99$ & $0.96\pm{0.03}$ & $0.98\pm{0.03}$ & $\bm{0.95}\pm{0.02}$ \\ \cmidrule(lr){2-6}
  Bottom-level & $2.49$ & $2.49$ & $2.47 \pm{0.04}$ & $2.59 \pm{0.03}$ & $\bm{2.28} \pm{0.02}$ \\[5pt]
  Average      & $4.12$ & 4.11 & $4.06 \pm{0.08}$ & $4.26 \pm{0.06}$ & $\bm{3.68} \pm{0.05}$ \\ \bottomrule
 \end{tabular}
 \label{tb:rmse_kansai}
\end{table}

\begin{figure}[t!]
\centering
\footnotesize
\tabcolsep = 2pt
\begin{tabular}{ccc}
\includegraphics[keepaspectratio,scale=0.2]{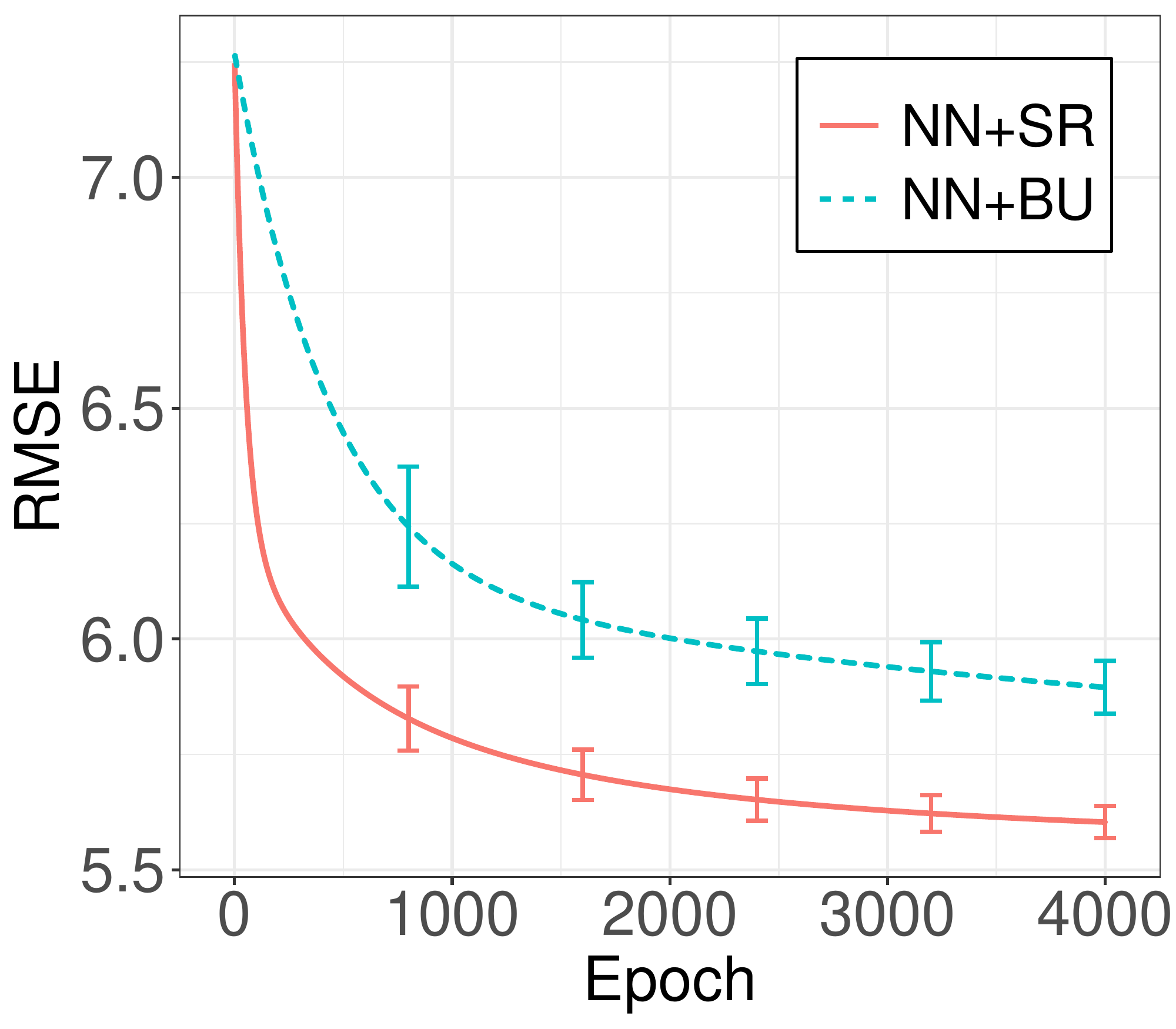} & 
\includegraphics[keepaspectratio, scale=0.2]{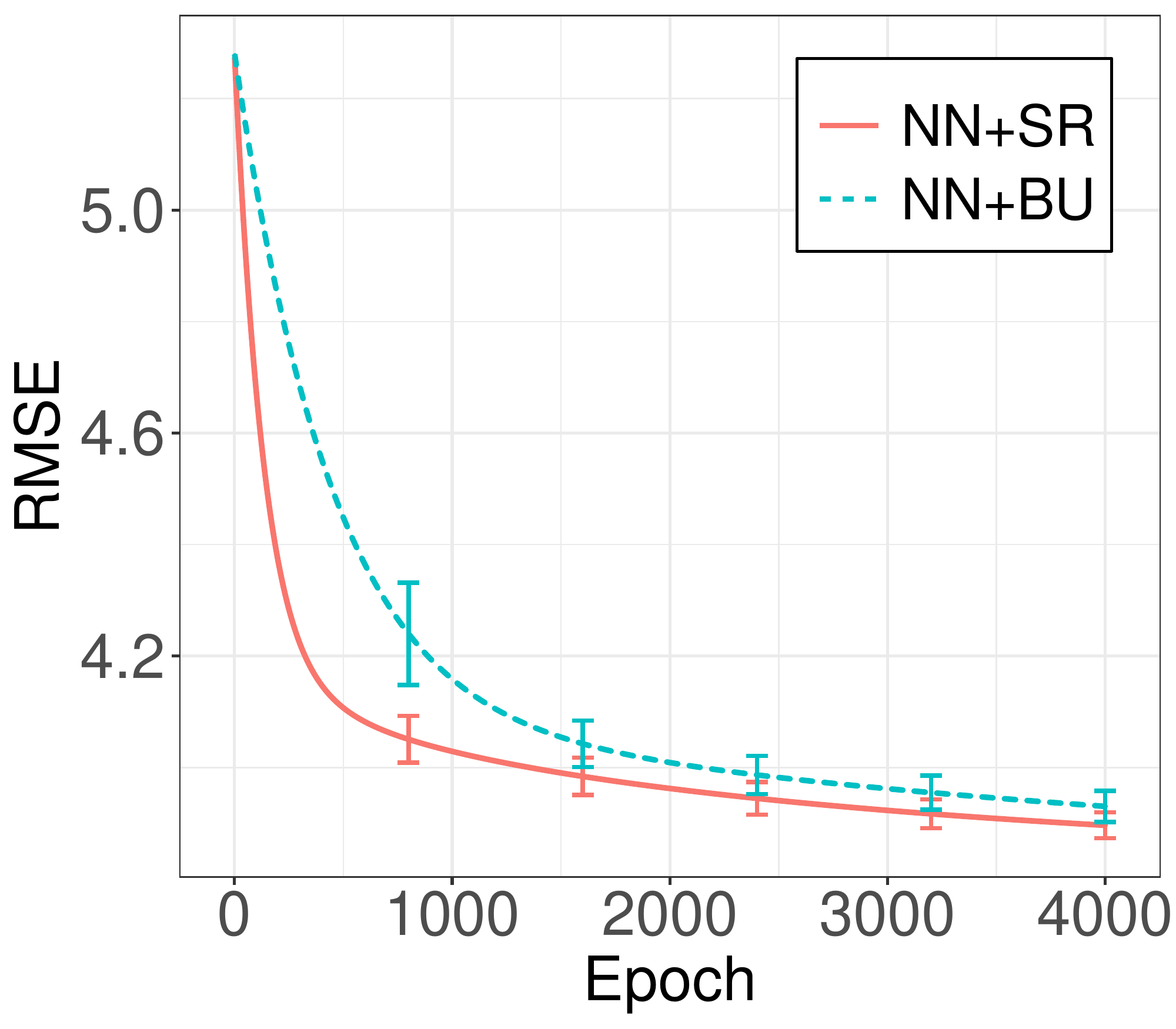} & 
\includegraphics[keepaspectratio, scale=0.2]{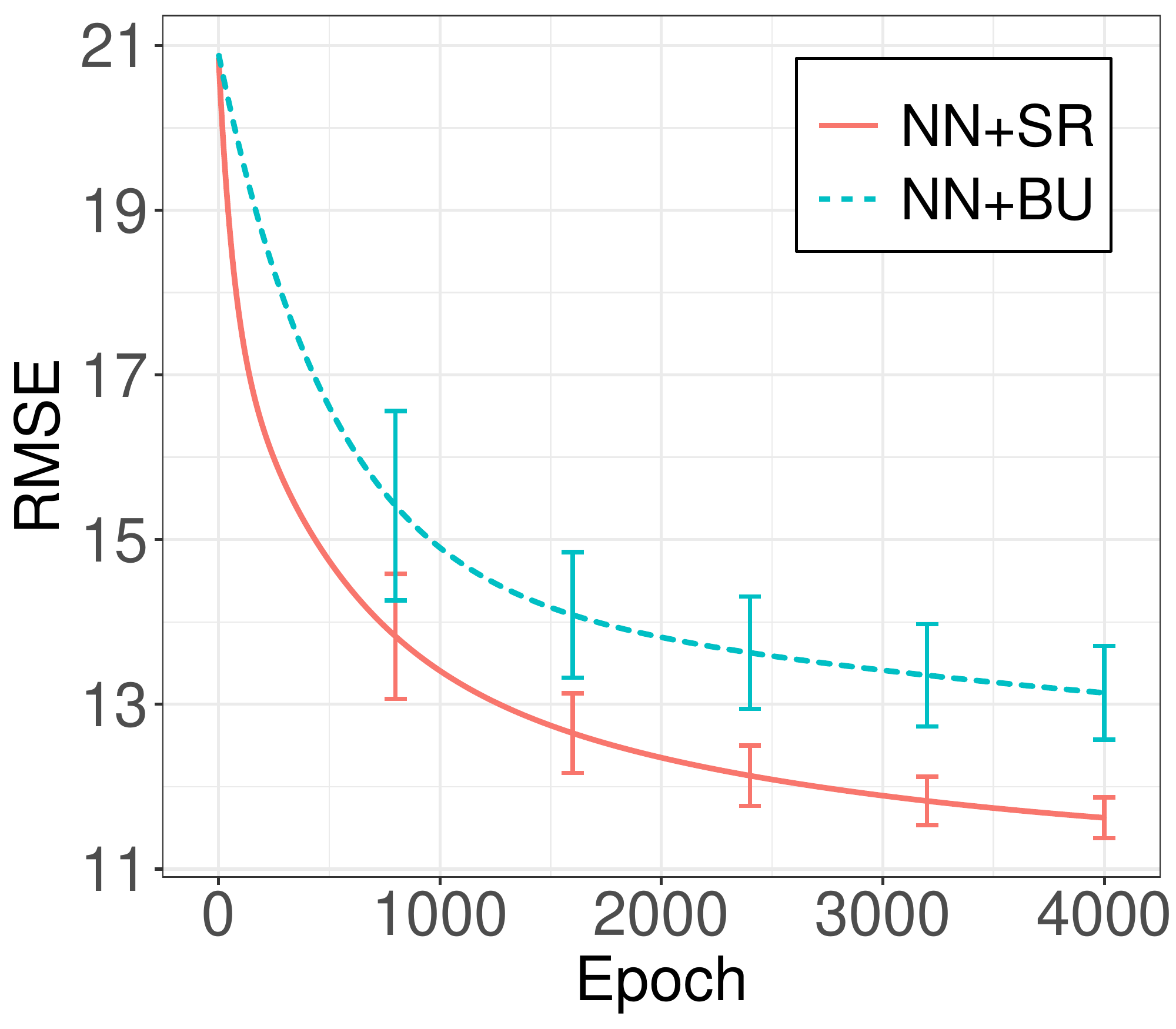} \\
(a) Root (Tohoku) & (b) Root (Chubu) & (c) Root (Kansai) \\[0.3cm]
\includegraphics[keepaspectratio, scale=0.2]{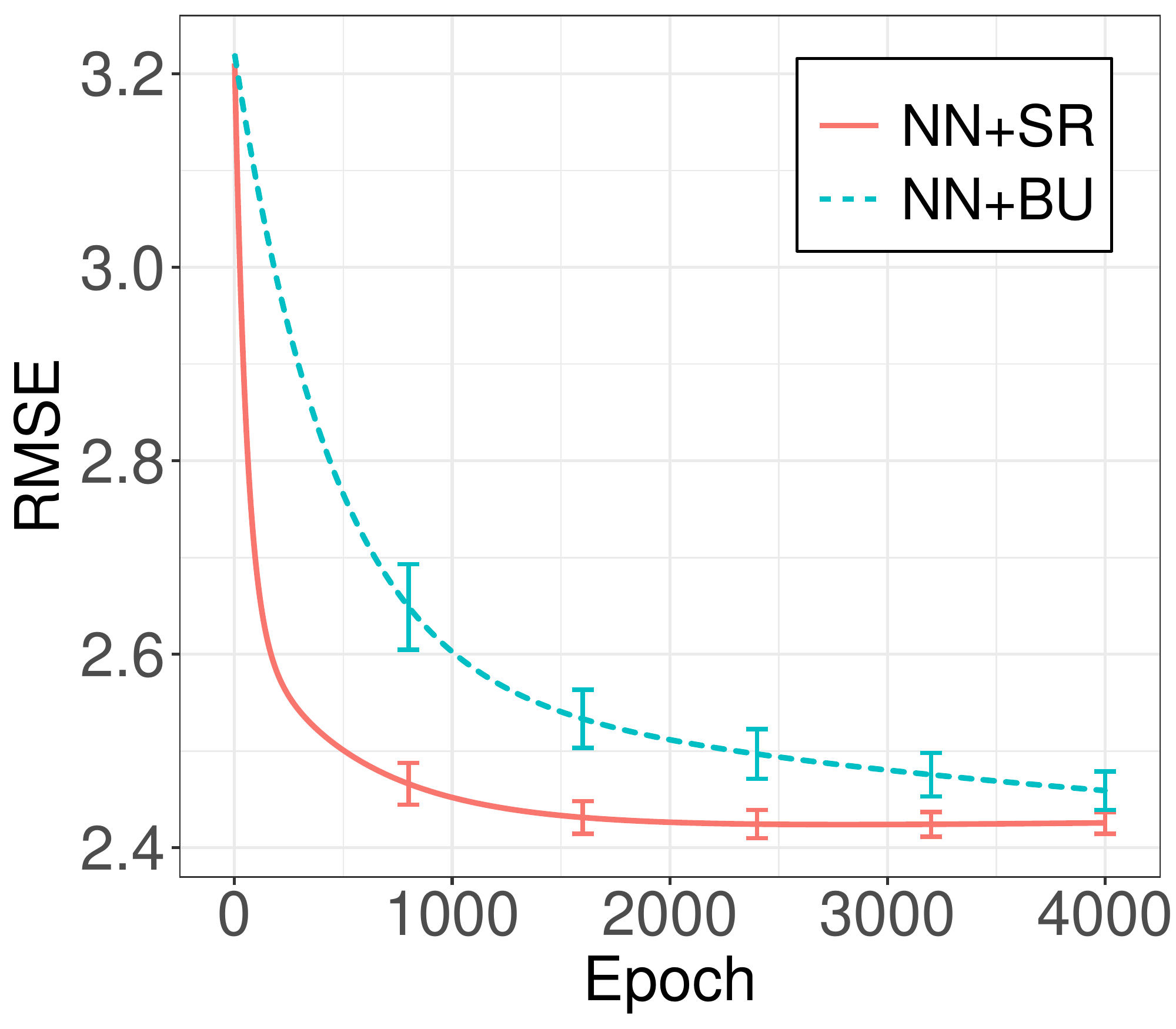} &
\includegraphics[keepaspectratio, scale=0.2]{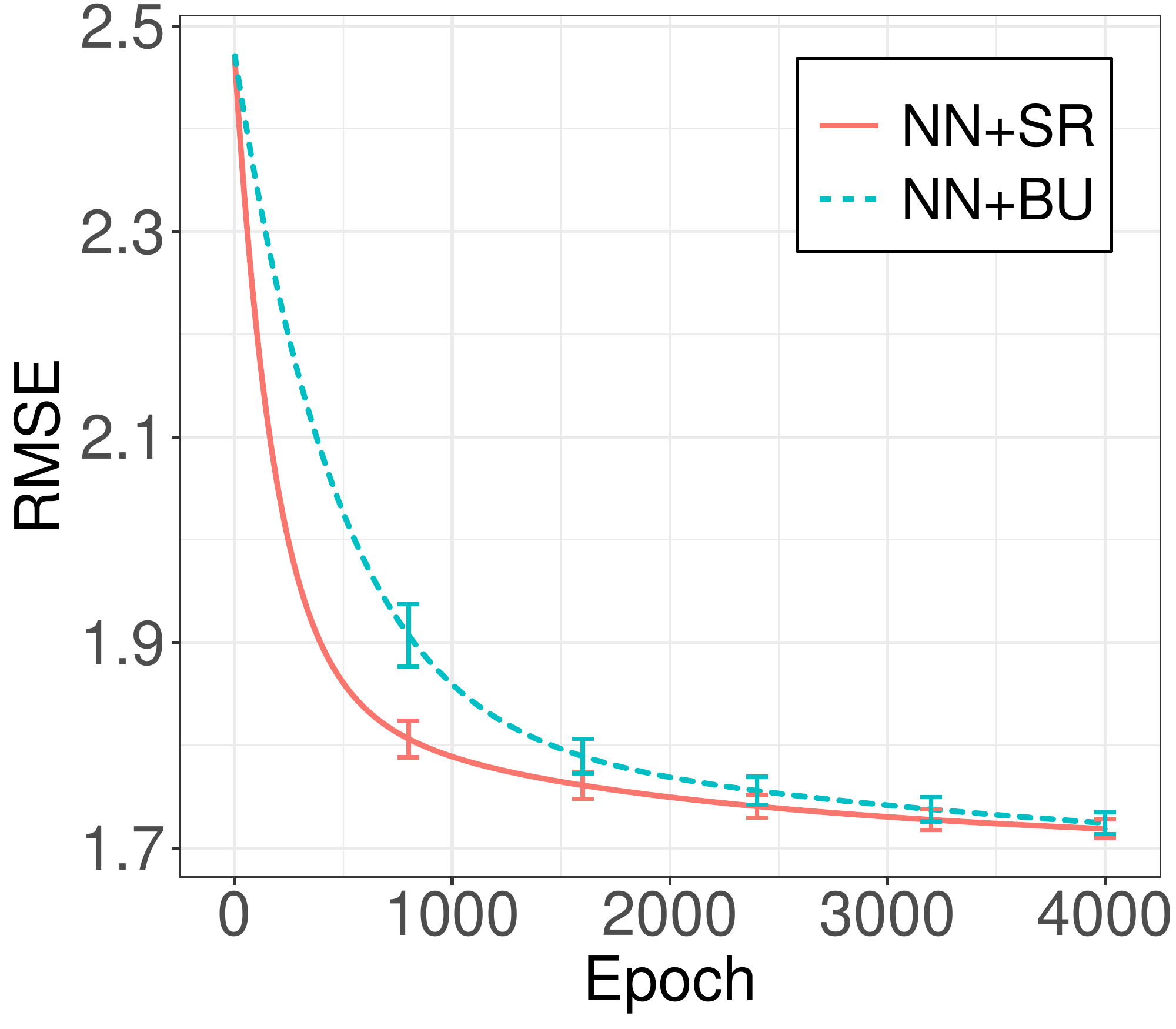} &
\includegraphics[keepaspectratio, scale=0.2]{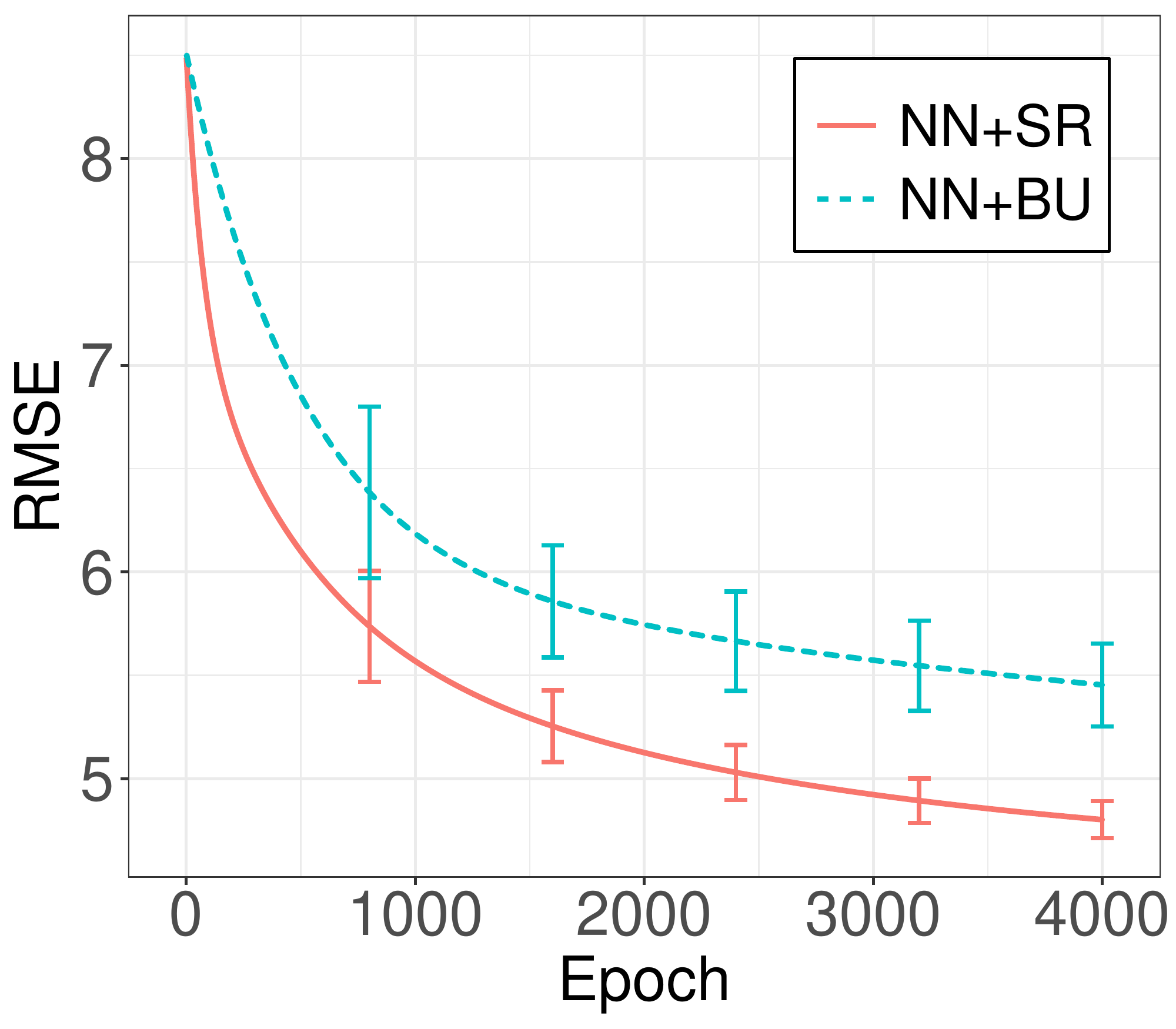} \\
(d) Mid-level (Tohoku) & (e) Mid-level (Chubu) & (f) Mid-level (Kansai) \\[0.3cm]
\includegraphics[keepaspectratio, scale=0.2]{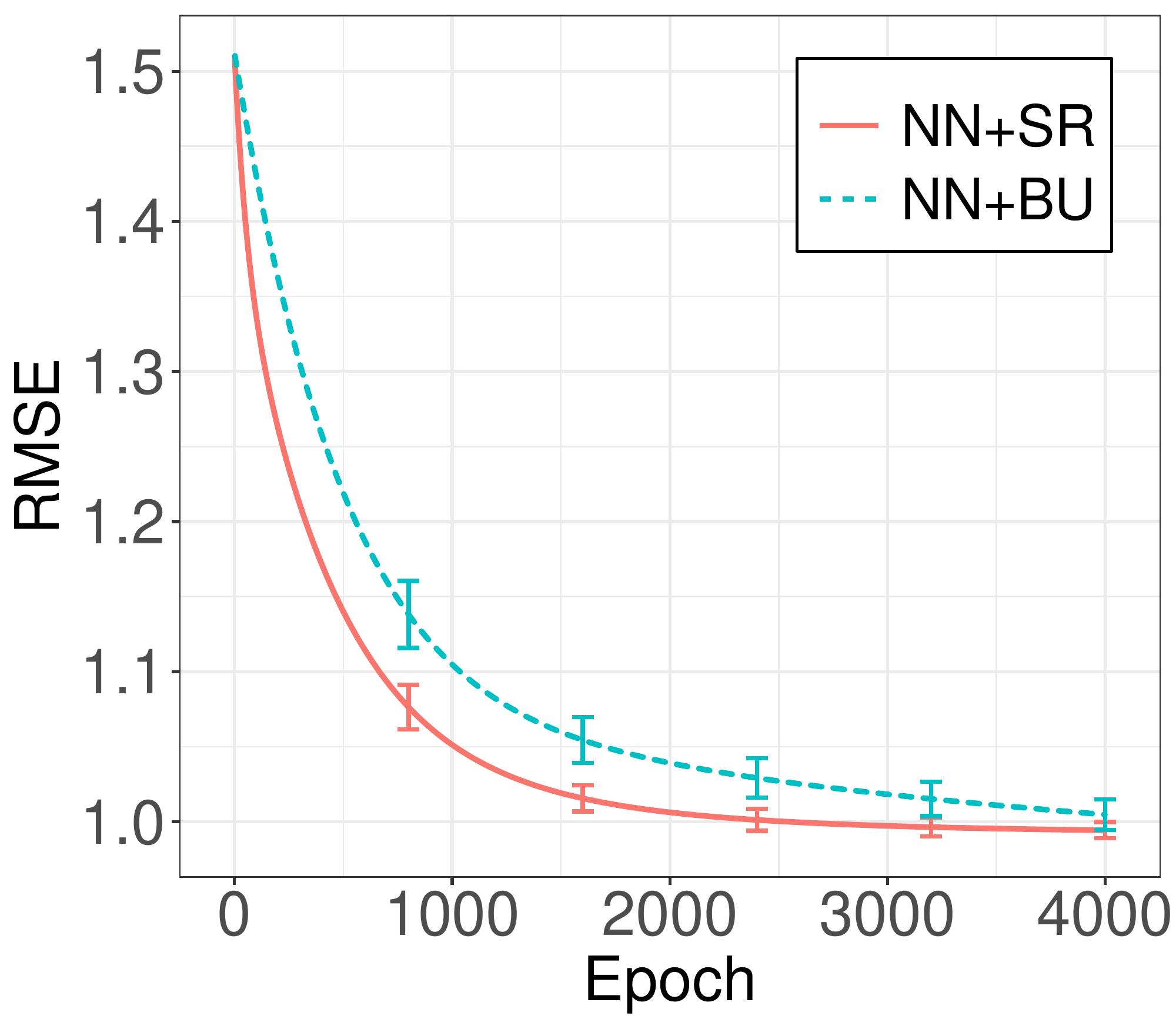} & 
\includegraphics[keepaspectratio, scale=0.2]{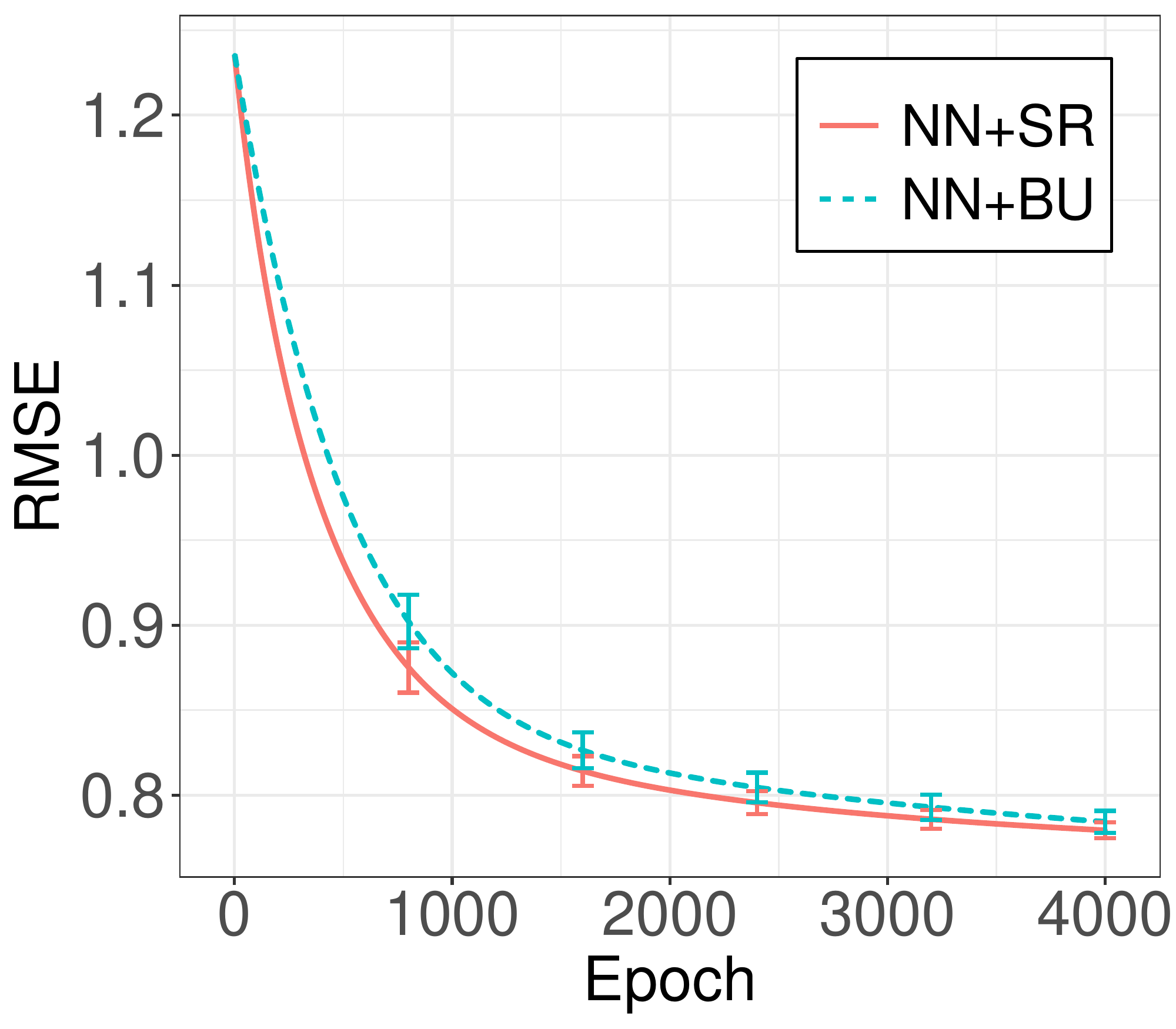} & 
\includegraphics[keepaspectratio, scale=0.2]{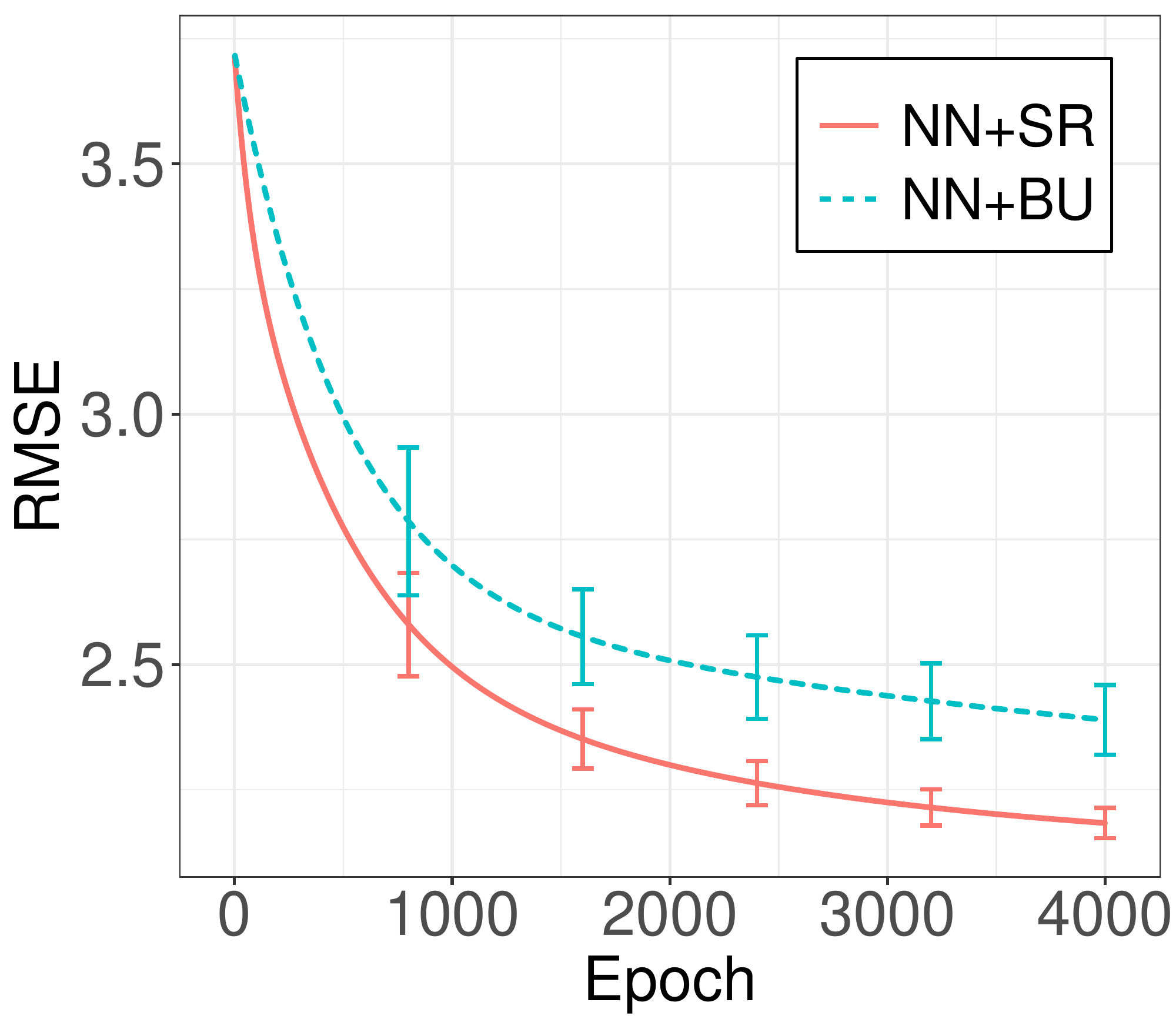} \\
(g) Bottom-level (Tohoku) & (h) Bottom-level (Chubu) & (i) Bottom-level (Kansa) \\[0.3cm]
\includegraphics[keepaspectratio, scale=0.2]{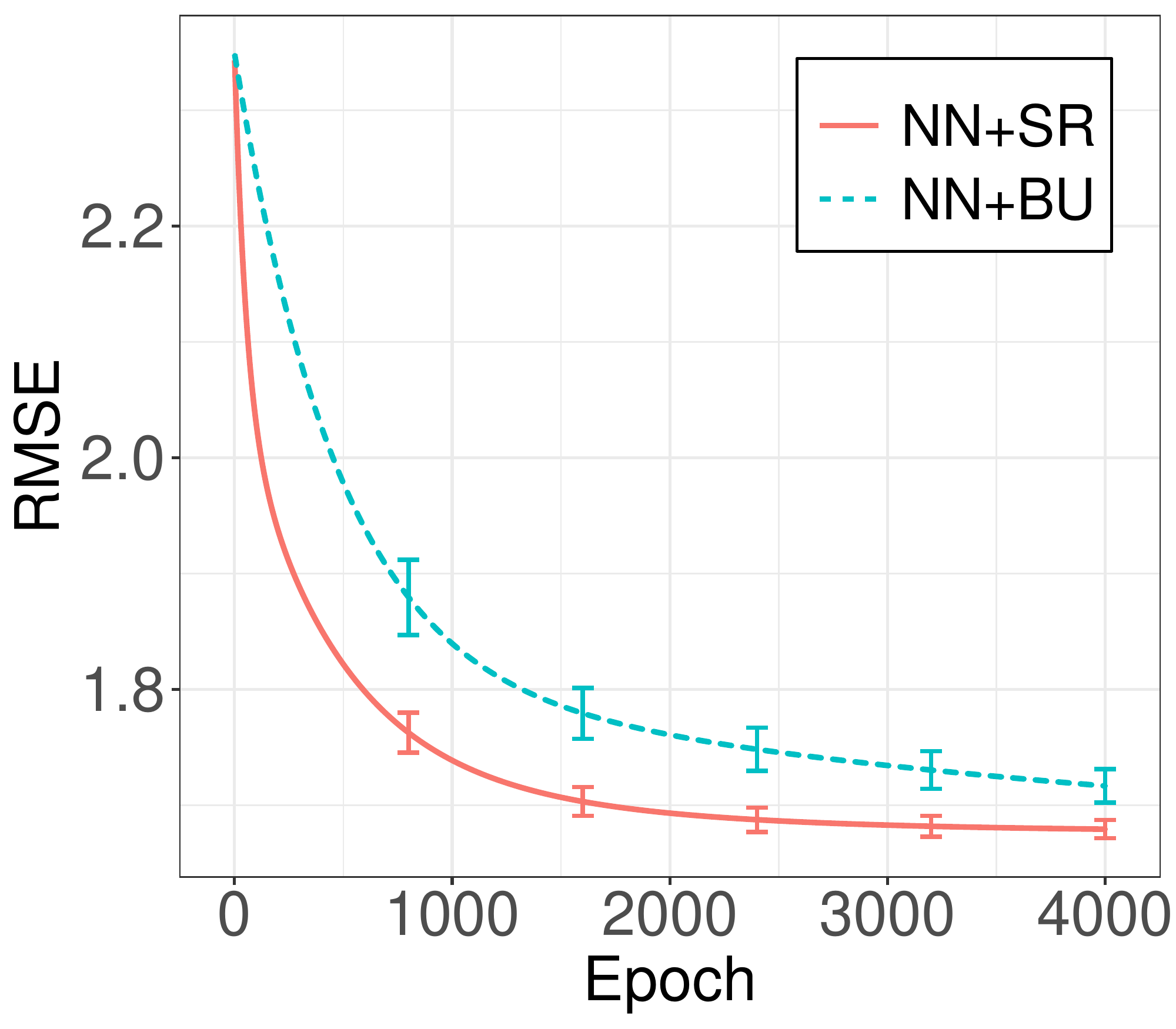} & 
\includegraphics[keepaspectratio, scale=0.2]{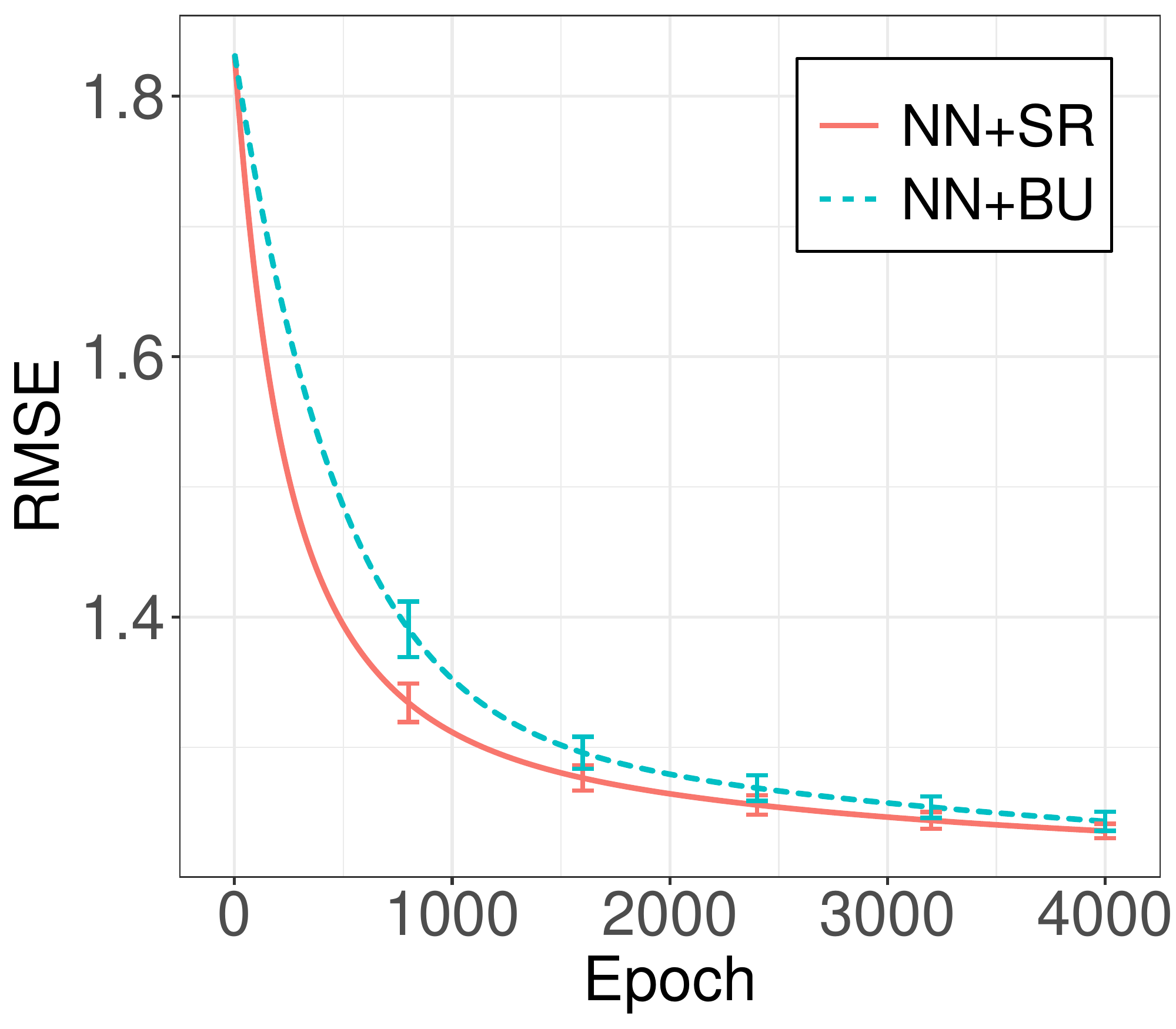} & 
\includegraphics[keepaspectratio, scale=0.2]{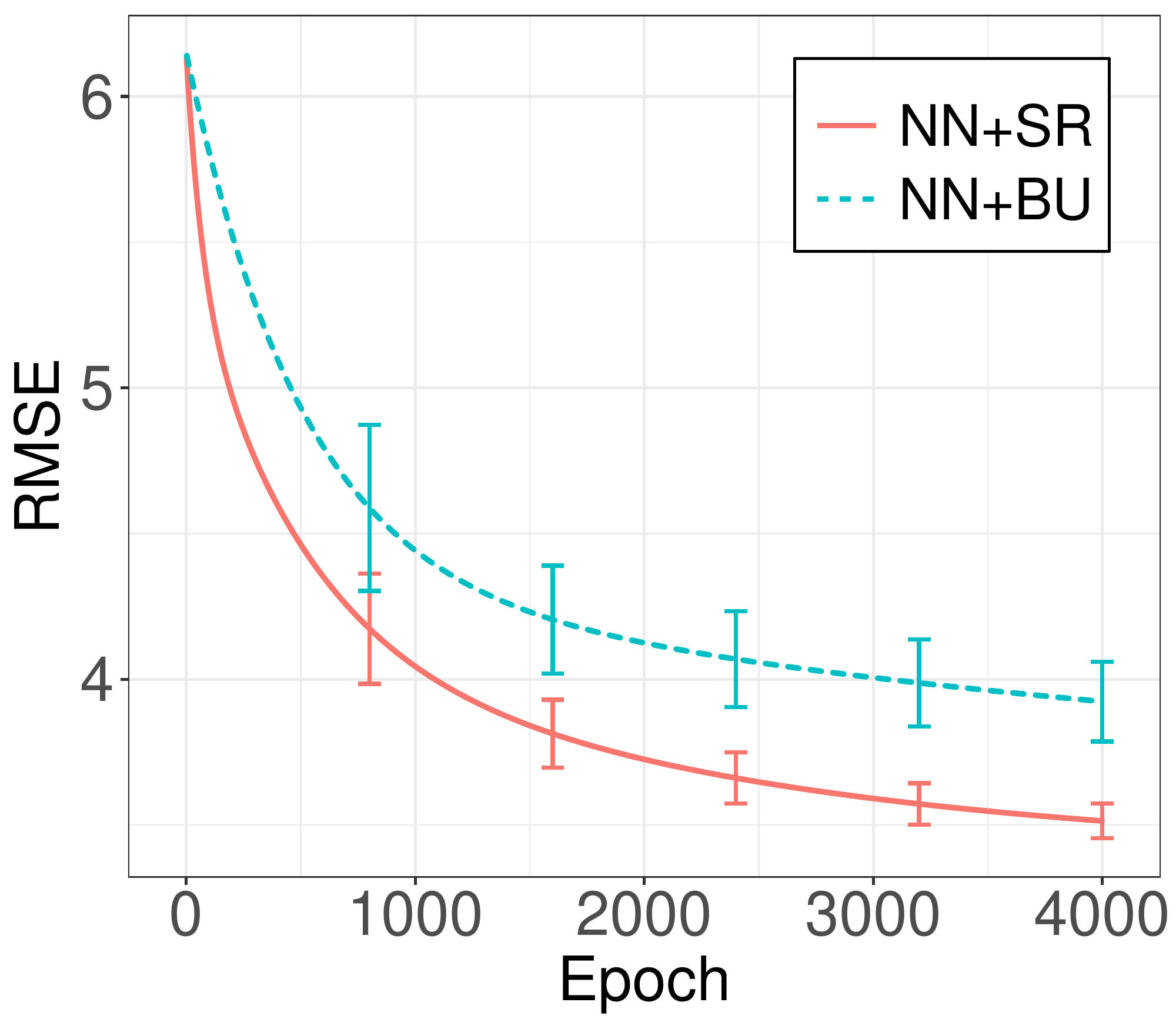} \\
(j) Average (Tohoku) & (k) Average (Chubu) & (l) Average (Kansai) \\[0.3cm]
\end{tabular}
\caption{Convergence performance of the backpropagation algorithm for the real-world datasets.}
\label{fig:EpoReal}
\end{figure}

Fig.~\ref{fig:EpoReal} shows the out-of-sample relative RMSE values as a function of epoch in the backpropagation algorithm for the real-world datasets. 
The convergence of RMSE values was consistently faster for our structured regularization method NN+SR than for the bottom-up method NN+BU. 
For the Tohoku and Chubu datasets, our method greatly accelerated convergence for upper-level time series. 
For the Kansai dataset, our method was superior to the bottom-up method in terms of both prediction accuracy and convergence speed. 
These results suggest that our structured regularization method improves the convergence performance of the backpropagation algorithm. 

\begin{figure}[t!]
\centering
\footnotesize
\tabcolsep = 2pt
\begin{tabular}{ccc}
\includegraphics[keepaspectratio, scale=0.2]{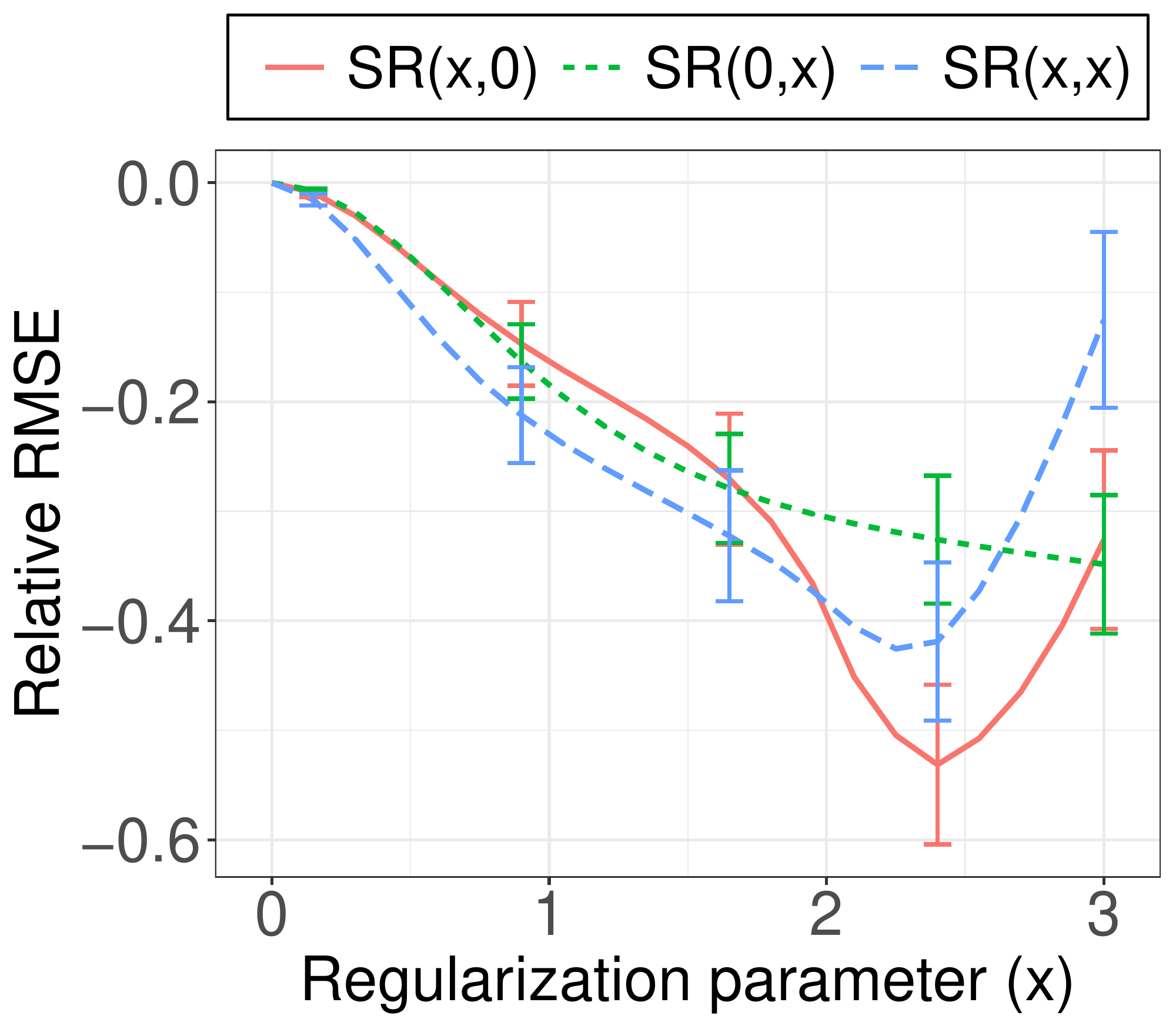} & 
\includegraphics[keepaspectratio, scale=0.2]{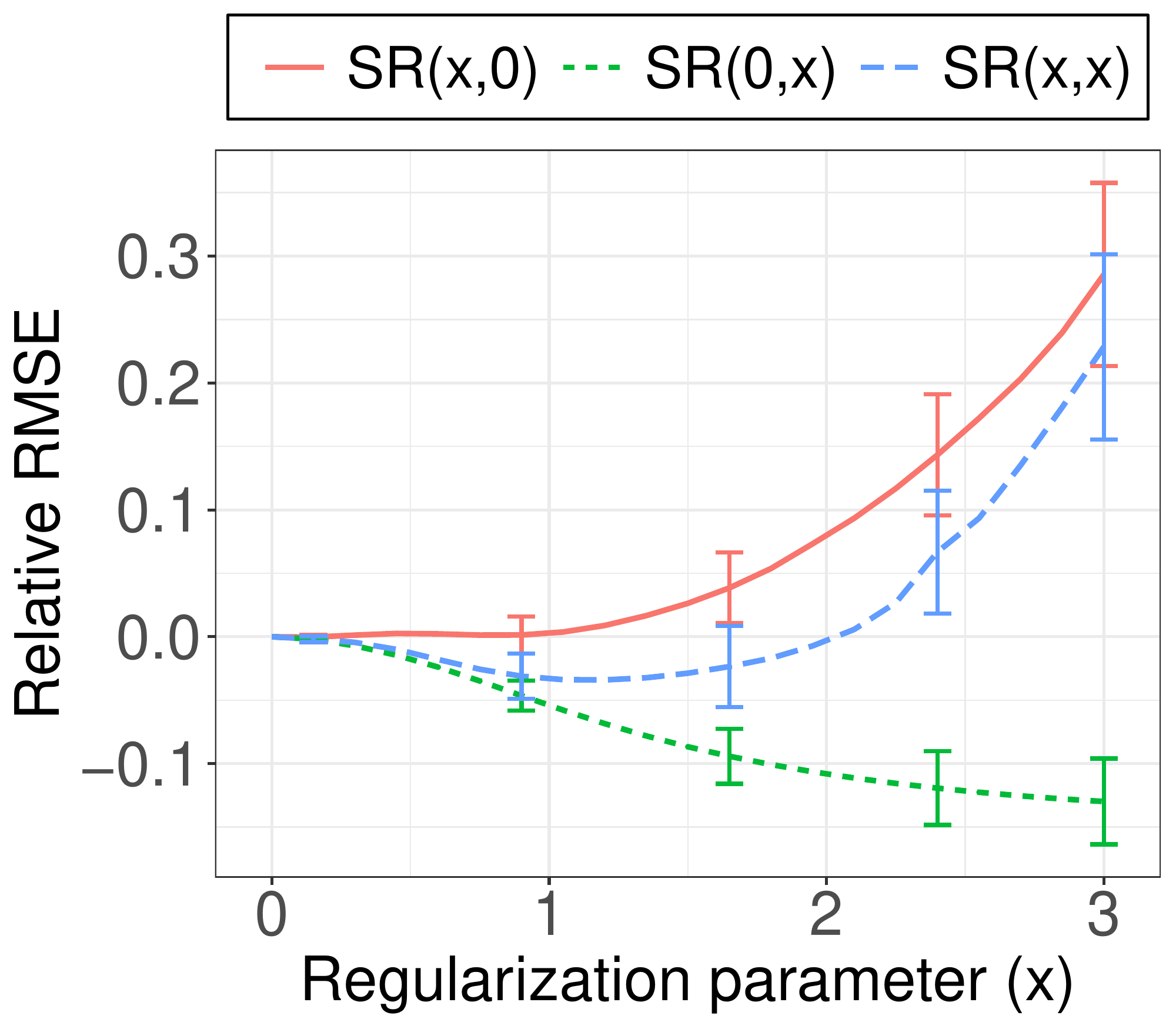} & 
\includegraphics[keepaspectratio, scale=0.2]{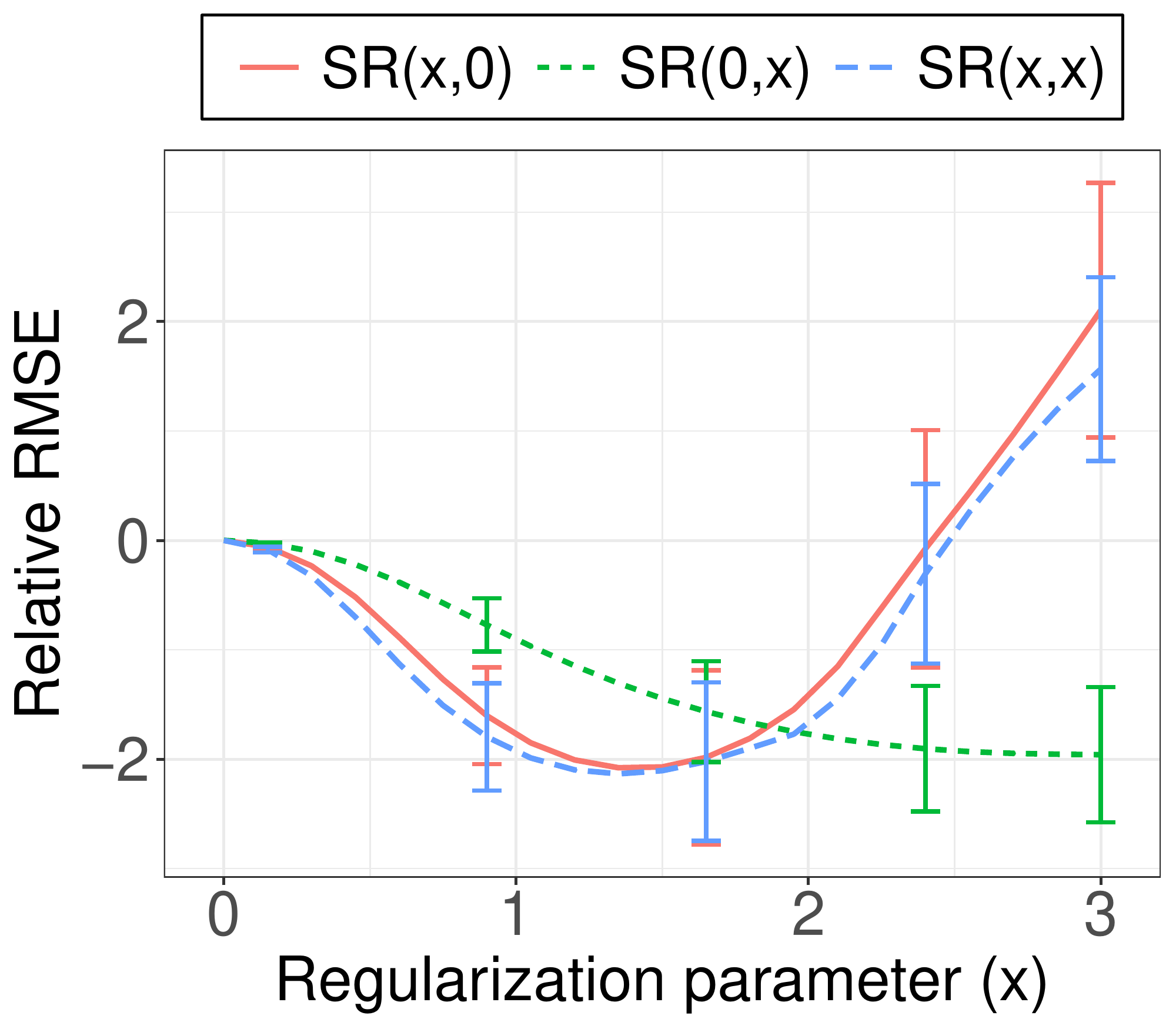} \\
(a) Root (Tohoku) & (b) Root (Chubu) & (c) Root (Kansai) \\[0.3cm]
\includegraphics[keepaspectratio, scale=0.2]{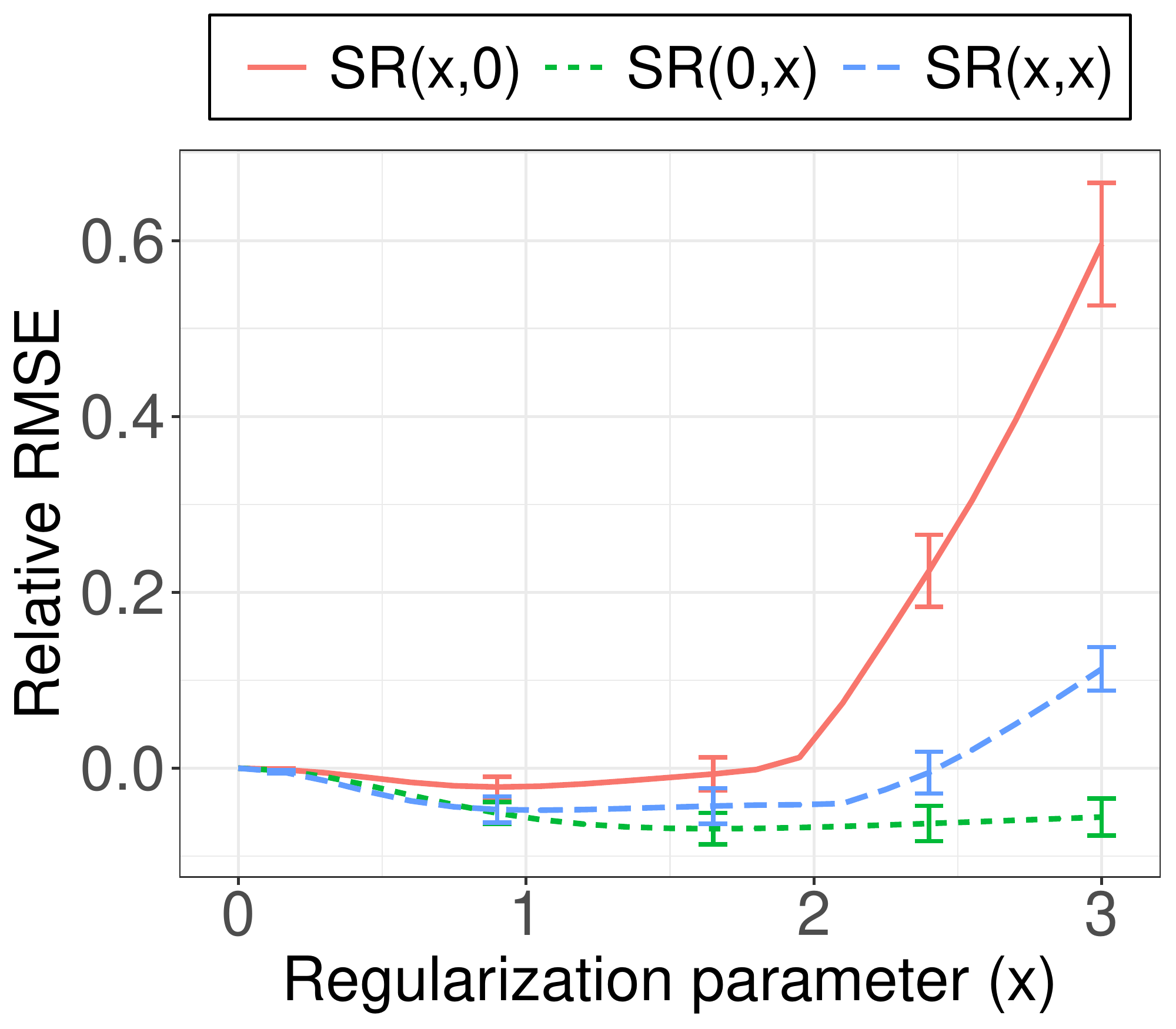} &
\includegraphics[keepaspectratio, scale=0.2]{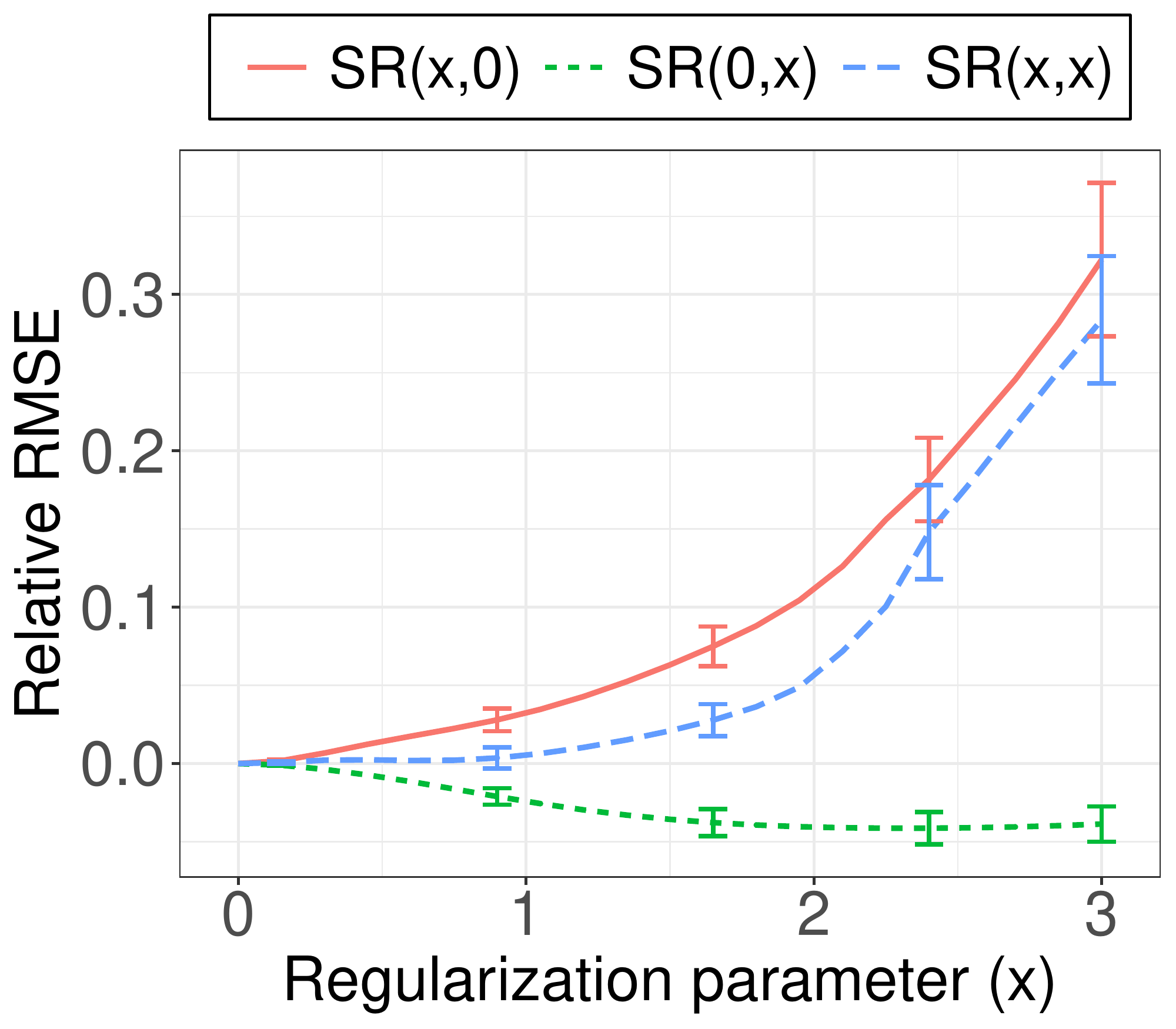} &
\includegraphics[keepaspectratio, scale=0.2]{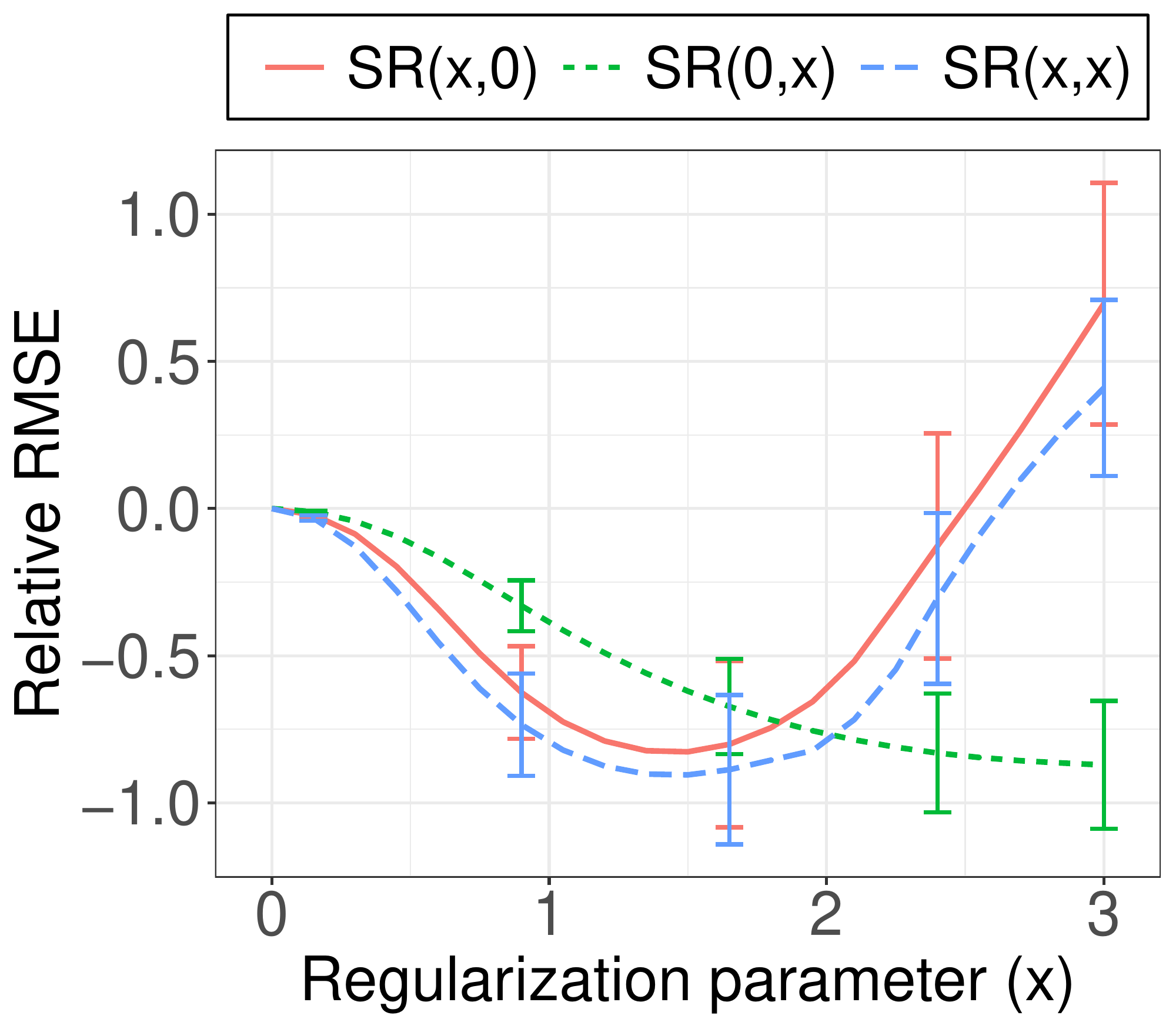} \\
(d) Mid-level (Tohoku) & (e) Mid-level (Chubu) & (f) Mid-level (Kansai) \\[0.3cm]
\includegraphics[keepaspectratio, scale=0.2]{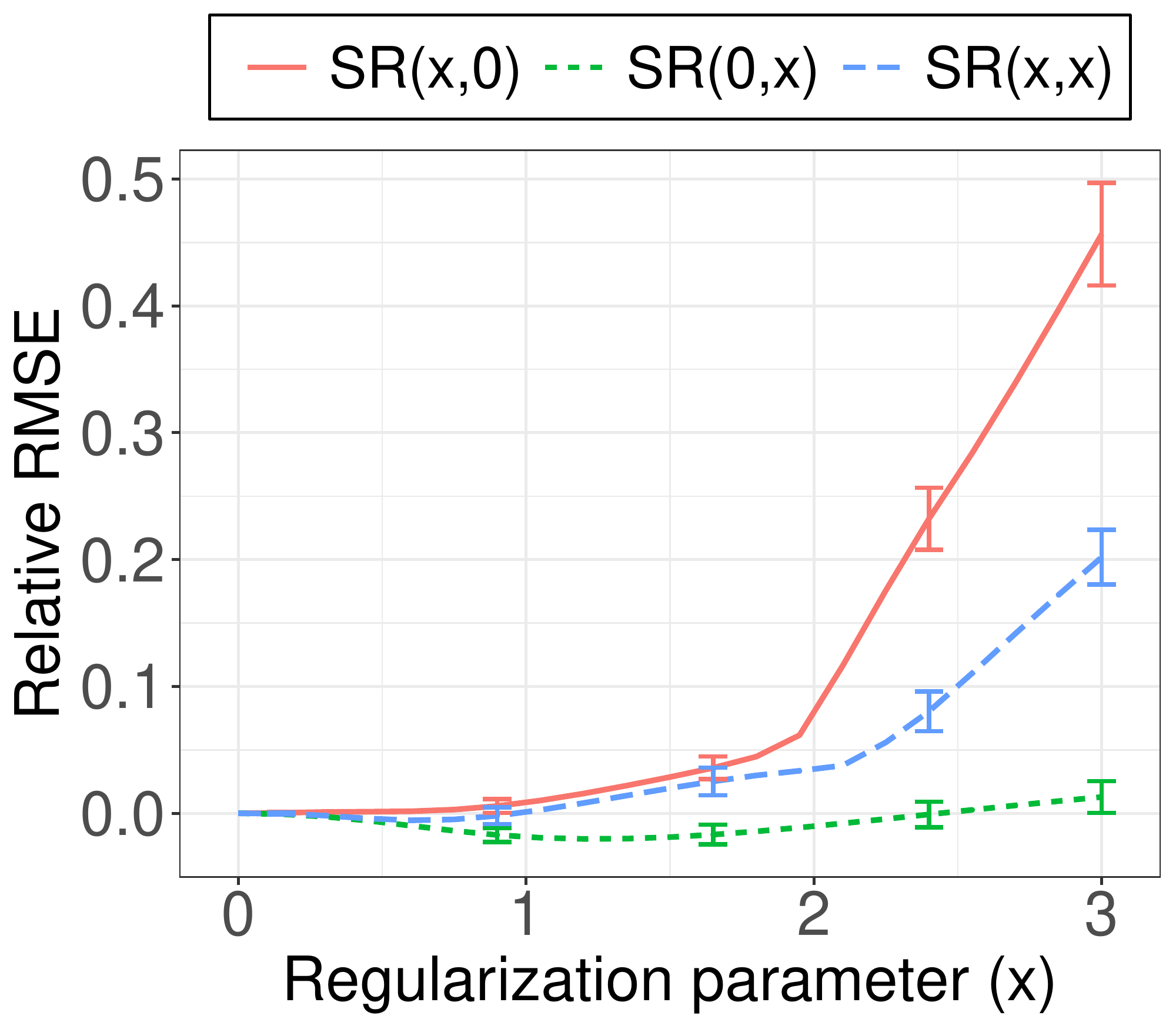} & 
\includegraphics[keepaspectratio, scale=0.2]{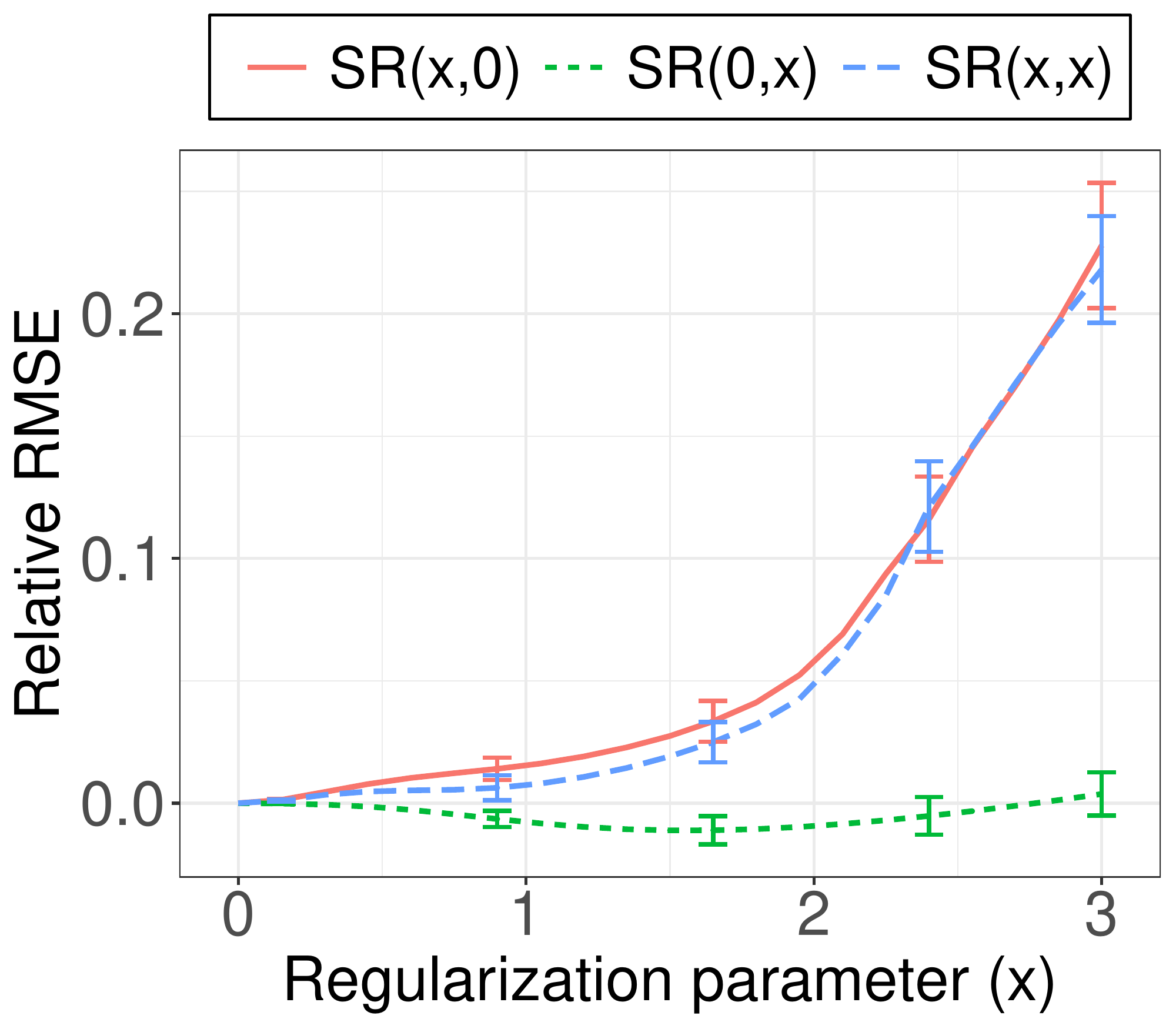} & 
\includegraphics[keepaspectratio, scale=0.2]{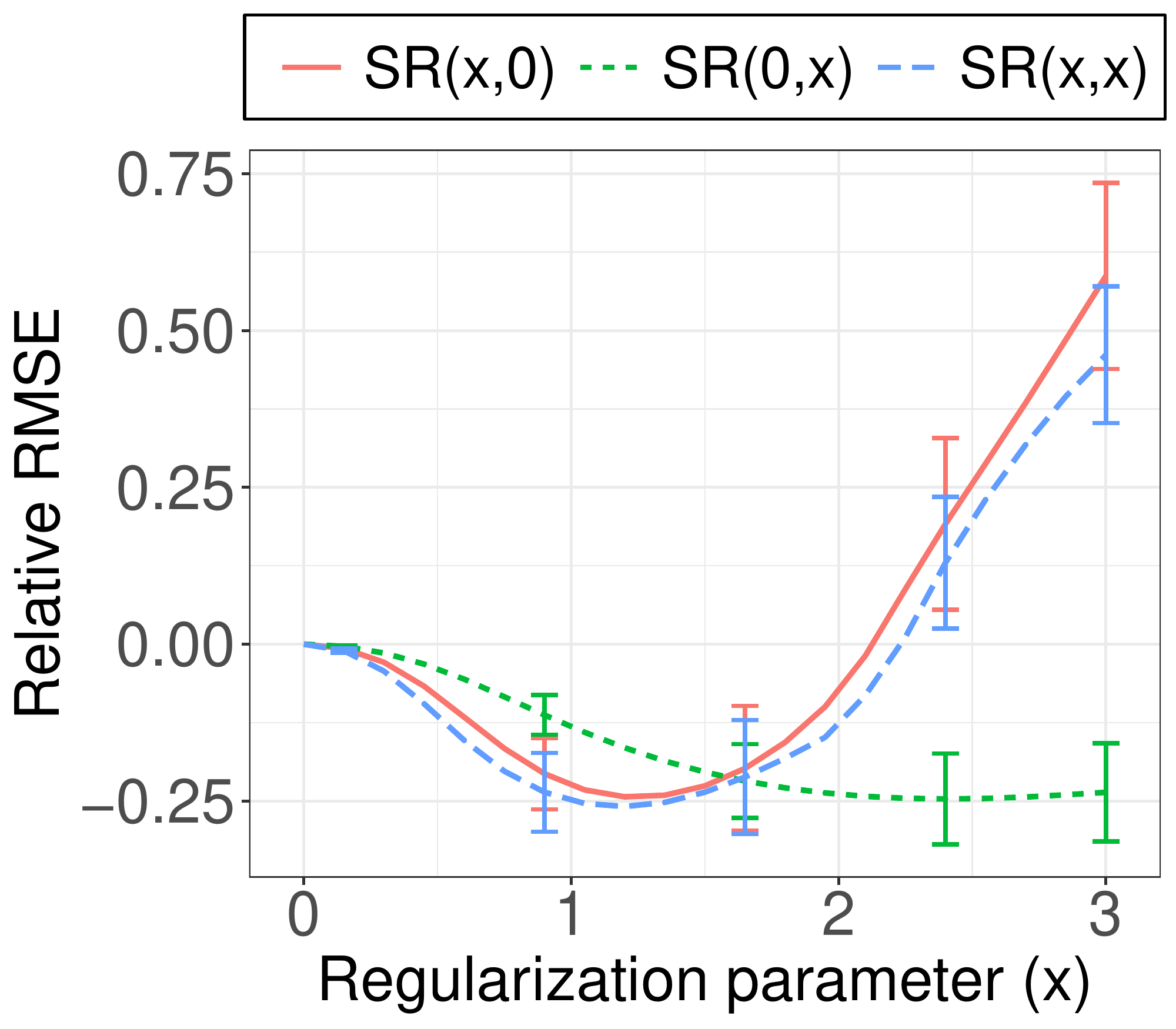} \\
(g) Bottom-level (Tohoku) & (h) Bottom-level (Chubu) & (i) Bottom-level (Kansai) \\[0.3cm]
\includegraphics[keepaspectratio, scale=0.2]{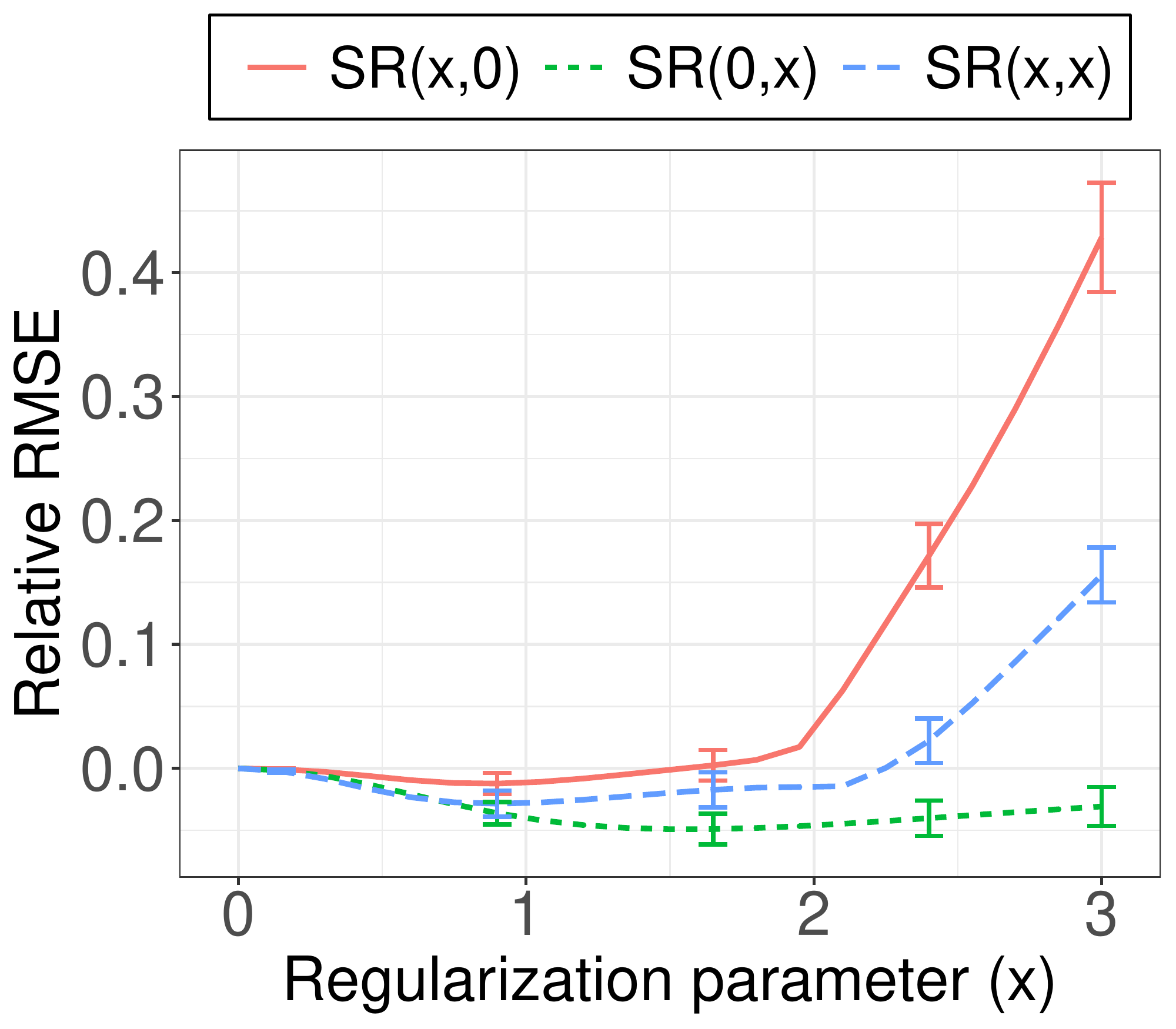} & 
\includegraphics[keepaspectratio, scale=0.2]{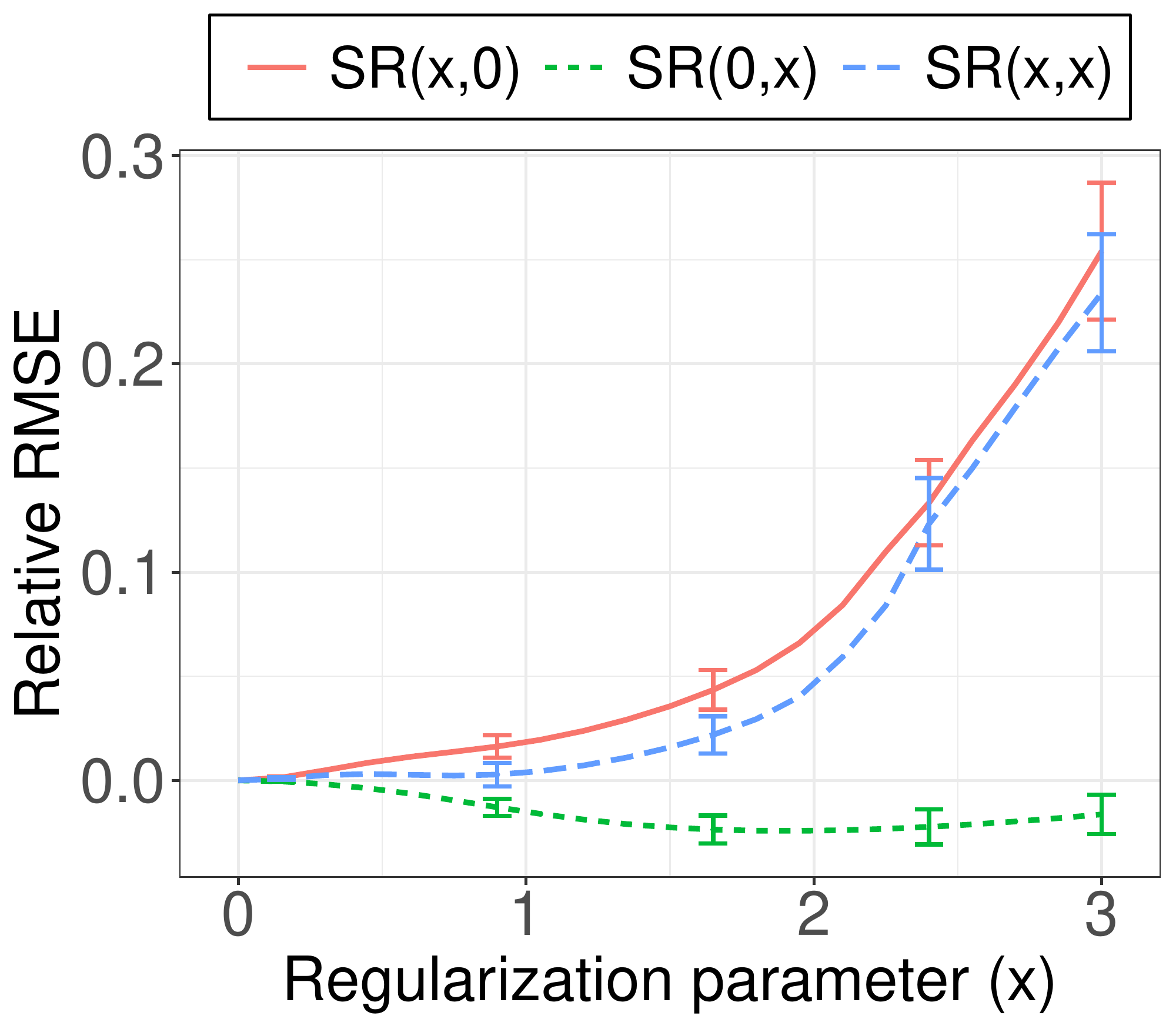} & 
\includegraphics[keepaspectratio, scale=0.2]{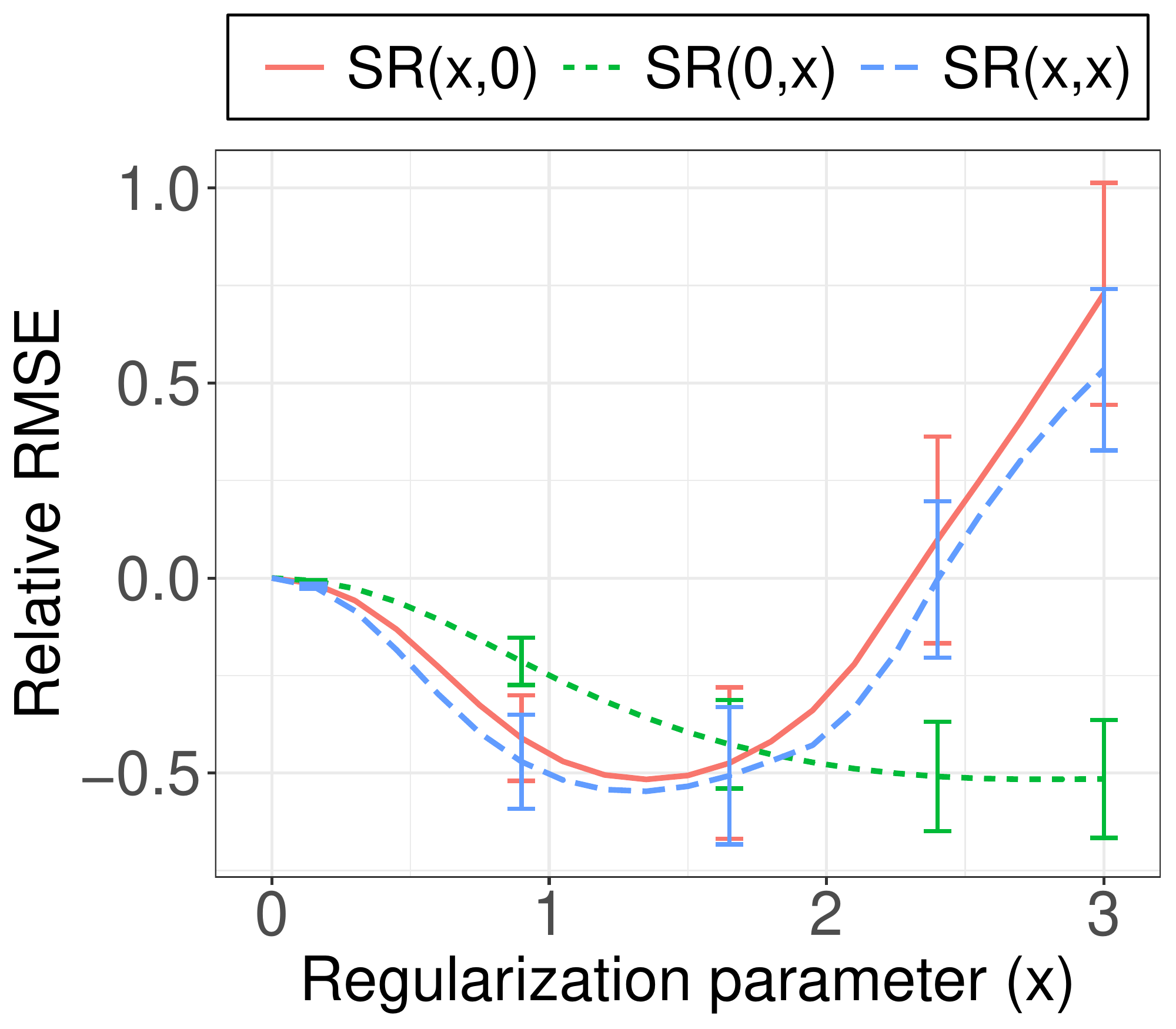} \\
(j) Average (Tohoku) & (k) Average (Chubu) & (l) Average (Kansai) \\[0.3cm]
\end{tabular}
\caption{Effects of structured regularization in the real-world datasets.}
\label{fig:RegReal}
\end{figure}

Fig.~\ref{fig:RegReal} shows the out-of-sample relative RMSE values provided by our structured regularization method NN+SR($\lambda_1,\lambda_M$) for the real-world datasets. 
For the Tohoku dataset, NN+SR($0,x$) reduced the RMSE values at all levels, and the reduction was particularly large for the root time series. 
For the Chubu dataset, NN+SR($0,x$) outperformed the other methods at all levels, meaning that the regularization for mid-level time series was the most effective. 
For the Kansai dataset, all the methods can greatly reduce the RMSE values if the regularization parameters are properly tuned. 

\section*{Conclusion}
We proposed a structured regularization model for predicting hierarchical time series. 
Our model uses the regularization term for improving upper-level forecasts to correct bottom-level forecasts. 
We demonstrated application of our model to artificial neural networks for time series prediction. 
We also developed a backpropagation algorithm specialized for training our model based on artificial neural networks. 

We investigated the efficacy of our method through experiments using synthetic and real-world datasets. 
The experimental results demonstrated that our method, which can adjust regularization parameters to fit data characteristics, achieved better prediction performance than did other methods that develop coherent forecasts for hierarchical time series. 
Our regularization term accelerated the backpropagation algorithm, and 
regularization for mid-level time series was particularly useful for achieving better prediction performance. 

This study made three main contributions. 
First, we devised a structured regularization method that effectively provides good predictions of hierarchical time series. 
Next, we established a new computational framework of artificial neural networks for time series predictions.  
Finally, our experiments using synthetic and real-world datasets demonstrated the potential of specialized prediction methods for hierarchical time series. 

In future studies, we will extend our structured regularization model to other time series prediction methods, such as the autoregressive integrated moving average model~\cite{BrDa91,HyAt18} and support vector regression~\cite{KaMa19}. 
Another direction of future research will be to develop a high-performance estimation algorithm for our method based on various mathematical optimization techniques~\cite{BeKi16,BePa19,KuTa20,TaMi20,TaKo17,TaKo19}. 

\nolinenumbers

% Either type in your references using
% \begin{thebibliography}{}
% Text
% \end{thebibliography}
%
% or
%
% Compile your BiBTeX database using our plos2015.bst
% style file and paste the contents of your .bbl file
% here. See http://journals.plos.org/plosone/s/latex for 


\begin{thebibliography}{99}

\bibitem{AtAh09}
Athanasopoulos, G., Ahmed, R. A., \& Hyndman, R. J. (2009). 
Hierarchical forecasts for Australian domestic tourism. 
International Journal of Forecasting, 25(1), 146--166.

%\bibitem{BaJe12}
%Bach, F., Jenatton, R., Mairal, J., \& Obozinski, G. (2012). 
%Optimization with sparsity-inducing penalties. 
%Foundations and Trends{\textregistered} in Machine Learning, 4(1), 1--106.

\bibitem{BeKo19}
Ben Taieb, S., \& Koo, B. (2019, July). Regularized Regression for Hierarchical Forecasting Without Unbiasedness Conditions. In Proceedings of the 25th ACM SIGKDD International Conference on Knowledge Discovery \& Data Mining (pp. 1337--1347).

\bibitem{BeTa17}
Ben Taieb, S., Taylor, J. W., \& Hyndman, R. J. (2017, August). 
Coherent probabilistic forecasts for hierarchical time series. 
In Proceedings of the 34th International Conference on Machine Learning-Volume 70 (pp. 3348--3357). JMLR. org.

\bibitem{BeYu17}
Ben Taieb, S., Yu, J., Barreto, M. N., \& Rajagopal, R. (2017, February). 
Regularization in hierarchical time series forecasting with application to electricity smart meter data. 
In Thirty-First AAAI Conference on Artificial Intelligence.

\bibitem{BeHy18}
Bergmeir, C., Hyndman, R. J., \& Koo, B. (2018). 
A note on the validity of cross-validation for evaluating autoregressive time series prediction. 
Computational Statistics \& Data Analysis, 120, 70--83.

\bibitem{BeKi16}
Bertsimas, D., King, A., \& Mazumder, R. (2016). 
Best subset selection via a modern optimization lens. 
The annals of statistics, 813--852.

\bibitem{BePa19}
Bertsimas, D., Pauphilet, J., \& Van Parys, B. (2019). 
Sparse regression: Scalable algorithms and empirical performance. 
arXiv preprint arXiv:1902.06547.

\bibitem{BiTa13}
Bien, J., Taylor, J., \& Tibshirani, R. (2013). 
A lasso for hierarchical interactions. 
Annals of statistics, 41(3), 1111.

\bibitem{Bi06}
Bishop, C. M. (2006). 
Pattern recognition and machine learning. 
Springer.

\bibitem{BrDa91}
Brockwell, P. J., Davis, R. A., \& Fienberg, S. E. (1991). 
Time series: Theory and methods. Springer Science \& Business Media.

\bibitem{CaCo10}
Capistr{\'a}n, C., Constandse, C., \& Ramos-Francia, M. (2010). 
Multi-horizon inflation forecasts using disaggregated data. 
Economic Modelling, 27(3), 666--677.

\bibitem{Ca97}
Caruana, R. (1997). 
Multitask learning. 
Machine learning, 28(1), 41--75.

%\bibitem{GoHy06}
%De Gooijer, J. G., \& Hyndman, R. J. (2006). 
%25 years of time series forecasting. 
%International Journal of Forecasting, 22(3), 443--473.

\bibitem{EvPo04}
Evgeniou, T., \& Pontil, M. (2004, August). 
Regularized multi-task learning. 
In Proceedings of the tenth ACM SIGKDD international conference on Knowledge discovery and data mining (pp. 109--117).

\bibitem{Fl99}
Fliedner, G. (1999). 
An investigation of aggregate variable time series forecast strategies with specific subaggregate time series statistical correlation. 
Computers \& Operations Research, 26(10--11), 1133--1149.

\bibitem{GaMu18}
Gao, J., Murphey, Y. L., \& Zhu, H. (2018). 
Multivariate time series prediction of lane changing behavior using deep neural network. 
Applied Intelligence, 48(10), 3523--3537.

\bibitem{GoBe16}
Goodfellow, I., Bengio, Y., \& Courville, A. (2016). 
Deep learning. 
MIT press.

\bibitem{HaTi15}
Hastie, T., Tibshirani, R., \& Wainwright, M. (2015). 
Statistical learning with sparsity: the lasso and generalizations. 
CRC press.

\bibitem{Hs04}
Hsieh, W. W. (2004). 
Nonlinear multivariate and time series analysis by neural network methods. 
Reviews of Geophysics, 42(1).

\bibitem{HyAh11}
Hyndman, R. J., Ahmed, R. A., Athanasopoulos, G., \& Shang, H. L. (2011). 
Optimal combination forecasts for hierarchical time series. 
Computational Statistics \& Data Analysis, 55(9), 2579--2589.

\bibitem{HyAt18}
Hyndman, R. J., \& Athanasopoulos, G. (2018). 
Forecasting: principles and practice. 
OTexts.

\bibitem{HyLe16}
Hyndman, R. J., Lee, A. J., \& Wang, E. (2016). 
Fast computation of reconciled forecasts for hierarchical and grouped time series. 
Computational Statistics \& Data Analysis, 97, 16--32.

\bibitem{JaVe09}
Jacob, L., Vert, J. P., \& Bach, F. R. (2009). 
Clustered multi-task learning: A convex formulation. 
In Advances in neural information processing systems (pp. 745--752).

\bibitem{JeAu11}
Jenatton, R., Audibert, J. Y., \& Bach, F. (2011). 
Structured variable selection with sparsity-inducing norms. 
Journal of Machine Learning Research, 12(Oct), 2777--2824.

\bibitem{KaMa19}
Karmy, J. P., \& Maldonado, S. (2019). 
Hierarchical time series forecasting via support vector regression in the European travel retail industry. 
Expert Systems with Applications, 137, 59--73.

\bibitem{Ka12}
Karsoliya, S. (2012). 
Approximating number of hidden layer neurons in multiple hidden layer BPNN architecture. 
International Journal of Engineering Trends and Technology, 3(6), 714--717.

\bibitem{KhBi10}
Khashei, M., \& Bijari, M. (2010). 
An artificial neural network $(p, d, q)$ model for timeseries forecasting. 
Expert Systems with Applications, 37(1), 479--489.

\bibitem{KiXi12}
Kim, S., \& Xing, E. P. (2012). 
Tree-guided group lasso for multi-response regression with structured sparsity, with an application to eQTL mapping. 
The Annals of Applied Statistics, 6(3), 1095--1117.

\bibitem{KrSi16}
Kremer, M., Siemsen, E., \& Thomas, D. J. (2016). 
The sum and its parts: Judgmental hierarchical forecasting. 
Management Science, 62(9), 2745--2764.

\bibitem{KuTa20}
Kudo, K., Takano, Y., \& Nomura, R. (2020). 
Stochastic discrete first-order algorithm for feature subset selection. 
IEICE Transactions on Information and Systems, 103(7), 1693--1702.

\bibitem{LaCa18}
Lai, G., Chang, W. C., Yang, Y., \& Liu, H. (2018, June). 
Modeling long-and short-term temporal patterns with deep neural networks. 
In The 41st International ACM SIGIR Conference on Research \& Development in Information Retrieval (pp. 95--104).

\bibitem{LiHa15}
Lim, M., \& Hastie, T. (2015). 
Learning interactions via hierarchical group-lasso regularization. 
Journal of Computational and Graphical Statistics, 24(3), 627--654.

\bibitem{Li87}
Lippmann, R. (1987). 
An introduction to computing with neural nets. 
IEEE ASSP Magazine, 4(2), 4--22.

\bibitem{Lu11}
L{\"u}tkepohl, H. (2011). 
Forecasting aggregated time series variables. 
OECD Journal: Journal of Business Cycle Measurement and Analysis, 2010(2), 1--26.

\bibitem{NiMa17}
Nicholson, W. B., Matteson, D. S., \& Bien, J. (2017). 
VARX-L: Structured regularization for large vector autoregressions with exogenous variables. 
International Journal of Forecasting, 33(3), 627--651.

\bibitem{PaNa14}
Park, M., \& Nassar, M. (2014). 
Variational Bayesian inference for forecasting hierarchical time series. 
In International conference on machine learning (ICML), workshop on divergence methods for probabilistic inference, Beijing.

\bibitem{Ru17}
Ruder, S. (2017). 
An overview of multi-task learning in deep neural networks. 
arXiv preprint arXiv:1706.05098.

\bibitem{SaTa19}
Sato, T., Takano, Y., \& Nakahara, T. (2019). 
Investigating consumers' store-choice behavior via hierarchical variable selection. 
Advances in Data Analysis and Classification, 13(3), 621--639.

\bibitem{ScMo17}
Schimbinschi, F., Moreira-Matias, L., Nguyen, V. X., \& Bailey, J. (2017). 
Topology-regularized universal vector autoregression for traffic forecasting in large urban areas. 
Expert Systems with Applications, 82, 301--316.

\bibitem{TaMi20}
Takano, Y., \& Miyashiro, R. (2020). 
Best subset selection via cross-validation criterion. 
TOP, 1--14.

\bibitem{TaKo17}
Tamura, R., Kobayashi, K., Takano, Y., Miyashiro, R., Nakata, K., \& Matsui, T. (2017). 
Best subset selection for eliminating multicollinearity. 
Journal of the Operations Research Society of Japan, 60(3), 321--336.

\bibitem{TaKo19}
Tamura, R., Kobayashi, K., Takano, Y., Miyashiro, R., Nakata, K., \& Matsui, T. (2019). 
Mixed integer quadratic optimization formulations for eliminating multicollinearity based on variance inflation factor. 
Journal of Global Optimization, 73(2), 431--446.

\bibitem{VaCu15}
van Erven, T., \& Cugliari, J. (2015). 
Game-theoretically optimal reconciliation of contemporaneous hierarchical time series forecasts. 
In Modeling and stochastic learning for forecasting in high dimensions (pp. 297--317). Springer, Cham.

\bibitem{WeWu16}
Wen, W., Wu, C., Wang, Y., Chen, Y., \& Li, H. (2016). 
Learning structured sparsity in deep neural networks. 
In Advances in neural information processing systems (pp. 2074--2082).

\bibitem{WiAt19}
Wickramasuriya, S. L., Athanasopoulos, G., \& Hyndman, R. J. (2019). 
Optimal forecast reconciliation for hierarchical and grouped time series through trace minimization. 
Journal of the American Statistical Association, 114(526), 804--819.

\bibitem{WiVi09}
Widiarta, H., Viswanathan, S., \& Piplani, R. (2009). 
Forecasting aggregate demand: An analytical evaluation of top-down versus bottom-up forecasting in a production planning framework. 
International Journal of Production Economics, 118(1), 87--94.

\bibitem{Zh03}
Zhang, G. P. (2003). 
Time series forecasting using a hybrid ARIMA and neural network model. 
Neurocomputing, 50, 159--175.

\bibitem{ZhQi05}
Zhang, G. P., \& Qi, M. (2005). 
Neural network forecasting for seasonal and trend time series. 
European Journal of Operational Research, 160(2), 501--514.

\bibitem{ZhYa17}
Zhang, Y., \& Yang, Q. (2017). 
A survey on multi-task learning. 
arXiv preprint arXiv:1707.08114.

\bibitem{ZhRo09}
Zhao, P., Rocha, G., \& Yu, B. (2009). 
The composite absolute penalties family for grouped and hierarchical variable selection. 
The Annals of Statistics, 37(6A), 3468--3497.

\end{thebibliography}
\end{document}